\documentclass[11pt]{article}
\PassOptionsToPackage{T2A,T1}{fontenc}
\usepackage{yandex-research}
\pdfmapfile{=XCharter-report-cyrillic.map}
\usepackage{amsmath,amssymb,array,longtable,booktabs,calc,needspace}
\usepackage[normalem]{ulem}
\usepackage{xurl}
\DeclareUnicodeCharacter{2192}{\ensuremath{\rightarrow}}

\providecommand{\ul}[1]{\uline{#1}}
\title{AliceAI-Foundation-80B-A3B-Base}
\author{Alice AI Team}
\affiliations{Yandex}
\hypersetup{pdftitle={AliceAI-Foundation-80B-A3B-Base},pdfauthor={Alice AI Team, Yandex}}

\begin{document}
\maketitle

The full author list and contributions are provided in Appendix A.

A Technical Report

\begin{abstract}\label{abstract}
We present AliceAI-Foundation-80B-A3B-Base, a mixture-of-experts
language model with 80 billion total parameters and approximately 3
billion active parameters per token, trained from scratch at Yandex and
released under the Apache 2.0 license. The model combines Kimi Delta
Attention with full attention, uses Attention Residuals, and is trained
with Muon. Training comprises general pretraining on 17.5 trillion
tokens, two context-extension stages, and a dedicated stage for
reasoning and agentic interaction. We describe the data collection and
augmentation pipelines, evaluation protocols, architectural ablations,
and scaling-law experiments underlying the final system. We identify the
limitations of held-out training loss as a proxy for downstream
performance and introduce an auxiliary loss on mixture-of-experts
outputs to stabilize training at the predicted learning rate.
Domain-specific data selection and augmentation improve factual
knowledge, mathematics, and programming skills, while agent trajectories
introduced during pretraining improve performance on agentic benchmarks
after supervised fine-tuning. In zero-shot reasoning evaluation, the
released model achieves leading results among open base models of comparable
size, with 88.7\% pass@8 on IMO-AnswerBench and 82.9\% pass@8 on
LiveCodeBench v5--v6. To evaluate Russian-language factual knowledge, we
introduce and release two benchmarks with their evaluation protocols:
WikiWebFacts, which tests short factual questions drawn from
encyclopedic sources and search queries, and HardMultiQA, which tests
more demanding factual questions inspired by user interactions. Our
model scores 86.5\% on WikiWebFacts and 67.9\% on HardMultiQA, compared
with 83.2\% and 65.4\% for DeepSeek-V4-Flash-Base (284B total and 13B
active parameters) and 72.8\% and 54.5\% for
Nemotron-3-Super-120B-A12B-Base (120B total and 12B active parameters),
outperforming both larger models on these benchmarks.
\end{abstract}

\textbf{Model:} AliceAI-Foundation-80B-A3B-Base\footnote{\url{https://huggingface.co/yandex/AliceAI-Foundation-80B-A3B-Base/blob/main/README_en.md}}

\textbf{Benchmarks:} WikiWebFacts\footnote{\url{https://huggingface.co/datasets/yandex/WikiWebFacts/blob/main/README_en.md}};
HardMultiQA\footnote{\url{https://huggingface.co/datasets/yandex/HardMultiQA/blob/main/README_en.md}}.

\clearpage
\section*{Introduction}\label{introduction}

We released the weights of AliceAI-Foundation-80B-A3B-Base under the
Apache 2.0 license. We trained the model from random initialization at
Yandex as part of our work toward a unified reasoning model for the
Alice AI assistant.

In base-model evaluations, the model outperforms the larger
DeepSeek-V4-Flash-Base and Nemotron-3-Super models on many factual,
expert, mathematical, and programming tasks. It also improves on our
previous proprietary Alice AI LLM 235B Base in factual knowledge,
mathematics, programming, and long-context processing, despite having
approximately 3× fewer total parameters and 7× fewer active parameters.

This release is based on our earlier work on Alice AI LLM, a
proprietary model introduced in October 2025 and described in a
Russian-language technical blog post\footnote{\url{https://habr.com/ru/companies/yandex/articles/974594/}}
{[}\hyperref[_heading=h.78rggyhkzwah]{1}{]}. Whereas that model was
initialized from Qwen3-235B-A22B, the present model was trained from
scratch. The entire development process, including rebuilding the
training corpus, evaluating architectural alternatives, tuning
hyperparameters, and preparing reasoning and agentic training data, took
approximately six months.

Specifically we quantify the contributions of the corpus, architecture,
and hyperparameters through sequential ablation experiments, training
each configuration from scratch on 2 trillion tokens. We also report
experiments that yielded no performance gains. Our evaluation analysis
examines prompt sensitivity, automated grading, and how well base-model
evaluations predict performance after post-training.

Several observations guided the final design. Held-out loss on the
training distribution was a poor predictor of downstream benchmark
rankings. Hyperparameters predicted by scaling laws exposed unstable
growth in mixture-of-experts (MoE) activations, requiring a separate
stabilization mechanism. Data augmentation methods effective in short
continued-pretraining runs required substantial revision before they
yielded gains in full-scale pretraining. Finally, reasoning traces alone
were insufficient for strong agentic performance: adding tool
interactions during pretraining improved subsequent supervised
fine-tuning (SFT) and reinforcement learning (RL).

\section{Main Results}\label{main-results}

We compare AliceAI-Foundation-80B-A3B-Base with Qwen3.5-35B-A3B-Base,
GLM-4.5-Air-Base (106B-A12B), Nemotron-3-Super-120B-A12B-Base,
DeepSeek-V4-Flash-Base (284B-A13B). Except for Qwen, these models have
both more total and more active parameters than our model. Qwen has
fewer total parameters and the same number of active parameters.

To better assess progress in factual knowledge, particularly knowledge
relevant to Russian-speaking users, we also include our previous
proprietary base model, Alice AI LLM 235B Base. Introduced in October
2025 and described in a Russian-language technical blog post
{[}1{]}\footnote{\url{https://habr.com/ru/companies/yandex/articles/974594/}},
this model was developed through continued pretraining of
Qwen3-235B-A22B\footnote{\url{https://huggingface.co/Qwen/Qwen3-235B-A22B}}
on several trillion tokens, with an emphasis on Russian-language
capabilities. It provides a reference for evaluating whether our smaller
model improves on a substantially larger predecessor developed for the
same linguistic context.

All results for our model represent the final checkpoint after the
reasoning stage. Agentic evaluations instead use a common SFT procedure,
as described in Section~\hyperref[why-a-separate-reasoning-stage]{7}.

Our evaluations span factual knowledge, educational benchmarks,
mathematics, programming, and long-context processing. For challenging
mathematics and programming problems , we report pass@k, defined as the
fraction of problems solved in at least one of \(k\) attempts.
Evaluation protocols are detailed in
Section~\hyperref[general-protocol]{2}.

\par\addvspace{8pt}\noindent\begin{minipage}{\linewidth}\centering

\includegraphics[width=\linewidth,height=.46\textheight,keepaspectratio]{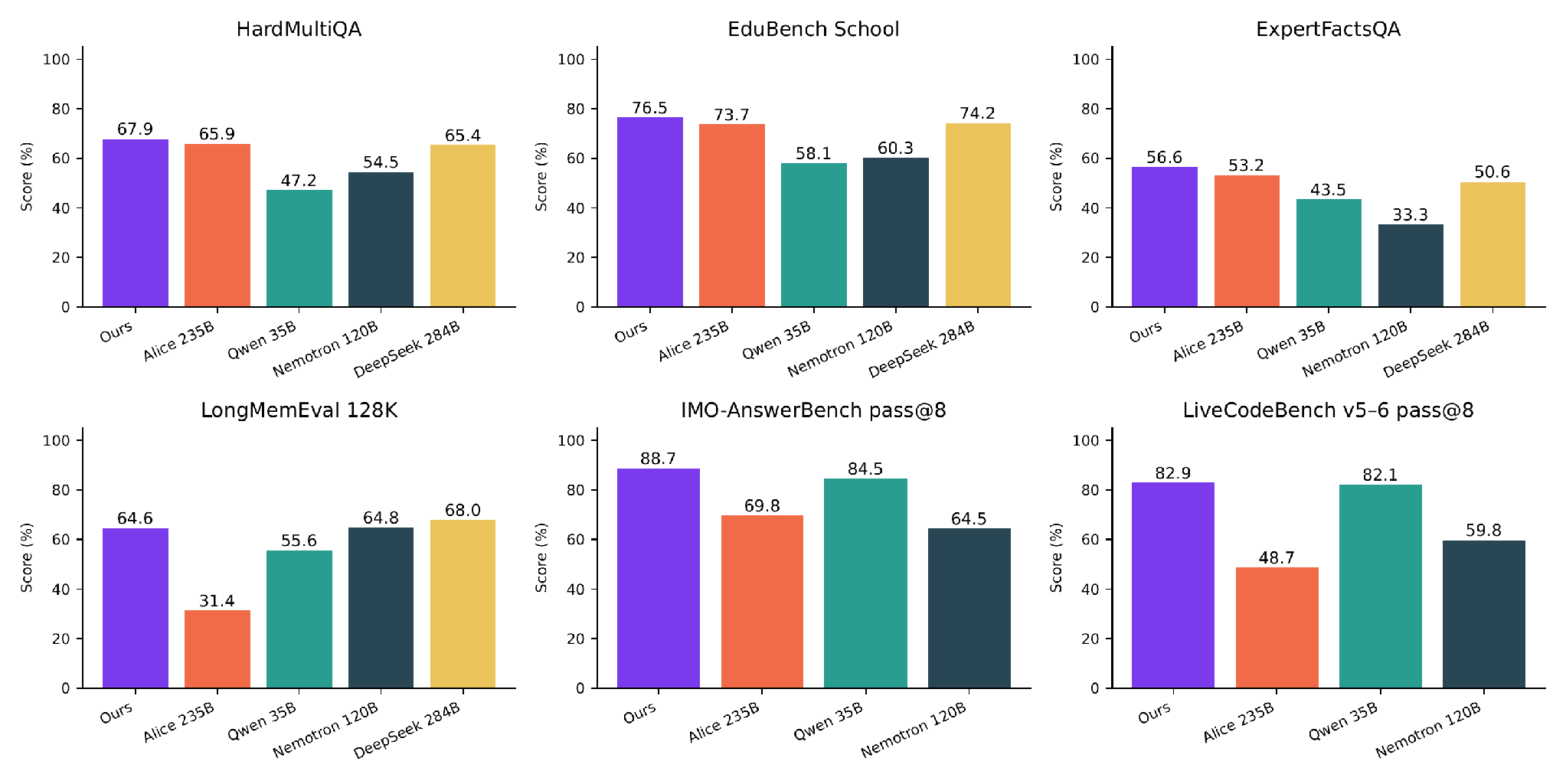}

\par

\captionof*{figure}{Selected benchmark results (\%). EduBench School is the unweighted
mean of the Russian, literature, history, and English-subject scores;
ExpertFactsQA is the mean of medicine and law. Individual benchmark
results are given in Tables~\hyperref[_heading=h.8qo1u2zptgl]{1}
and~\hyperref[_heading=h.6y8hn7xy6ljb]{2}. The Alice 235B reasoning
scores shown here are additional reference results not listed in
Table~\hyperref[_heading=h.6y8hn7xy6ljb]{2}.}

\end{minipage}\par\addvspace{7pt}

\subsection{Knowledge, Reasoning, and Long-Context
Performance}\label{knowledge-reasoning-and-long-context-performance}

All measurements in this subsection use our internal evaluation
infrastructure and vLLM inference with temperature~0 for every model. In
Table~\hyperref[_heading=h.8qo1u2zptgl]{1}, the best result in each row
is bold and the second-best distinct result is underlined.

\Needspace{9\baselineskip}\begingroup\small\setlength{\LTpre}{4pt}\setlength{\LTpost}{5pt}

\protect\phantomsection\label{_heading=h.8qo1u2zptgl}{}\captionof{table}{Knowledge and core-capability scores (\%). Unless stated otherwise,
evaluations use five demonstrations; CultCat uses four and code tasks
use one. TriviaQA and MATH-500 use an LLM judge. MATH-500 uses extended
reasoning demonstrations; BigCodeBench uses our revised tests. Best
scores are shown in bold and second-best distinct scores are
underlined.}

\nopagebreak[4]

\footnotesize\setlength{\tabcolsep}{2pt}

{\def\LTcaptype{reporttablebody} 
\begin{longtable}[]{@{}
  >{\raggedright\arraybackslash}p{(\linewidth - 14\tabcolsep) * \real{0.2420}}
  >{\raggedright\arraybackslash}p{(\linewidth - 14\tabcolsep) * \real{0.0600}}
  >{\raggedright\arraybackslash}p{(\linewidth - 14\tabcolsep) * \real{0.1280}}
  >{\raggedright\arraybackslash}p{(\linewidth - 14\tabcolsep) * \real{0.1050}}
  >{\raggedright\arraybackslash}p{(\linewidth - 14\tabcolsep) * \real{0.1150}}
  >{\raggedright\arraybackslash}p{(\linewidth - 14\tabcolsep) * \real{0.1100}}
  >{\raggedright\arraybackslash}p{(\linewidth - 14\tabcolsep) * \real{0.1200}}
  >{\raggedright\arraybackslash}p{(\linewidth - 14\tabcolsep) * \real{0.1200}}@{}}
\toprule\noalign{}
\begin{minipage}[b]{\linewidth}\raggedright
\textbf{Benchmark}
\end{minipage} & \begin{minipage}[b]{\linewidth}\raggedright
\textbf{Lang.}
\end{minipage} & \begin{minipage}[b]{\linewidth}\raggedright
\textbf{AliceAI-Foundation-80B-A3B-Base}
\end{minipage} & \begin{minipage}[b]{\linewidth}\raggedright
\textbf{Alice AI LLM 235B Base}
\end{minipage} & \begin{minipage}[b]{\linewidth}\raggedright
\textbf{Qwen3.5-35B-A3B-Base}
\end{minipage} & \begin{minipage}[b]{\linewidth}\raggedright
\textbf{GLM-4.5-Air-Base (106B-A12B)}
\end{minipage} & \begin{minipage}[b]{\linewidth}\raggedright
\textbf{Nemotron-3-Super-120B-A12B-Base}
\end{minipage} & \begin{minipage}[b]{\linewidth}\raggedright
\textbf{DeepSeek-V4-Flash-Base (284B-A13B)}
\end{minipage} \\
\midrule\noalign{}
\endhead
\bottomrule\noalign{}
\endlastfoot
\multicolumn{8}{@{}>{\raggedright\arraybackslash}p{(\linewidth - 14\tabcolsep) * \real{1.0000} + 14\tabcolsep}@{}}{%
\emph{Factual knowledge}} \\
WikiWebFacts & RU & \textbf{86.5} & \ul{86.2} & 62.4 & 70.2 & 72.8 &
83.2 \\
HardMultiQA & RU & \textbf{67.9} & \ul{65.9} & 47.2 & 48.6 & 54.5 &
65.4 \\
CultCat & RU & \textbf{86.5} & \ul{81.2} & 59.2 & 59.1 & 66.3 & 80.7 \\
TriviaQA & EN & 79.0 & 84.0 & 71.4 & 83.5 & \textbf{89.8} & \ul{89.4} \\
\multicolumn{8}{@{}>{\raggedright\arraybackslash}p{(\linewidth - 14\tabcolsep) * \real{1.0000} + 14\tabcolsep}@{}}{%
\emph{Education}} \\
EduBench Russian & RU & \textbf{74.2} & \ul{71.1} & 42.9 & 39.0 & 44.0 &
67.7 \\
EduBench Literature & RU & \textbf{73.8} & \ul{72.0} & 51.8 & 51.4 &
55.8 & 69.1 \\
EduBench History & RU & \textbf{82.0} & 76.7 & 65.9 & 62.8 & 70.2 &
\ul{76.9} \\
EduBench English & RU & \ul{76.1} & 74.9 & 71.7 & 67.2 & 71.3 &
\textbf{82.9} \\
\multicolumn{8}{@{}>{\raggedright\arraybackslash}p{(\linewidth - 14\tabcolsep) * \real{1.0000} + 14\tabcolsep}@{}}{%
\emph{Expert knowledge}} \\
ExpertFactsQA Medicine & RU & \textbf{63.6} & 59.9 & 59.0 & 50.6 & 42.3
& \ul{60.7} \\
ExpertFactsQA Law & RU & \textbf{49.6} & \ul{46.5} & 27.9 & 22.5 & 24.3
& 40.5 \\
\multicolumn{8}{@{}>{\raggedright\arraybackslash}p{(\linewidth - 14\tabcolsep) * \real{1.0000} + 14\tabcolsep}@{}}{%
\emph{Examinations}} \\
EGE CoT & RU & \ul{90.5} & \textbf{93.3} & 84.7 & 77.8 & 84.3 & 90.3 \\
MMLU-Pro CoT & EN & 66.8 & \ul{68.2} & 63.2 & 58.4 & \textbf{69.9} &
66.5 \\
SuperGPQA CoT & EN & 44.3 & 43.5 & 43.6 & 35.4 & \textbf{46.6} &
\ul{46.1} \\
\multicolumn{8}{@{}>{\raggedright\arraybackslash}p{(\linewidth - 14\tabcolsep) * \real{1.0000} + 14\tabcolsep}@{}}{%
\emph{Mathematics}} \\
MATH-500 & EN & \textbf{91.1} & 72.6 & 81.9 & 60.2 & \ul{84.8} & 80.7 \\
EduBench Math & RU & \ul{79.3} & 75.3 & \textbf{80.0} & 56.9 & 69.7 &
76.3 \\
EduBench Math University & RU & \textbf{70.1} & 63.0 & \ul{69.9} & 51.4
& 67.4 & 68.6 \\
\multicolumn{8}{@{}>{\raggedright\arraybackslash}p{(\linewidth - 14\tabcolsep) * \real{1.0000} + 14\tabcolsep}@{}}{%
\emph{Programming}} \\
BigCodeBench 1-shot pass@1 & EN & 48.3 & \textbf{49.8} & 43.5 & 44.5 &
48.8 & \ul{49.1} \\
LiveCodeBench v5-6 CoT 1-shot pass@1 & EN & \textbf{50.5} & 15.4 &
\ul{50.4} & 22.6 & \ul{50.4} & 38.1 \\
\multicolumn{8}{@{}>{\raggedright\arraybackslash}p{(\linewidth - 14\tabcolsep) * \real{1.0000} + 14\tabcolsep}@{}}{%
\emph{Long context}} \\
FinQA 128k & EN & \textbf{74.1} & 44.6 & \ul{73.5} & 35.5 & 71.7 &
\textbf{74.1} \\
LongMemEval 128k & EN & 64.6 & 31.4 & 55.6 & 50.6 & \ul{64.8} &
\textbf{68.0} \\
\end{longtable}
}

\endgroup

Our model ranks first on eight of the ten factual, educational, and
expert-knowledge benchmarks. It is particularly strong on Russian
culture, language, literature, history, medicine, and law. On
examination benchmarks, its performance is comparable to that of strong
open models. It achieves the highest scores in this comparison on
MATH-500, university mathematics, and the few-shot LiveCodeBench
evaluation. It ties DeepSeek on FinQA 128K and is 0.2 percentage points
behind second-ranked Nemotron on LongMemEval.

\subsection{Zero-Shot Reasoning}\label{zero-shot-reasoning}

We also evaluate performance on tasks requiring extended reasoning.
DeepSeek-V4-Flash-Base is omitted from
Table~\hyperref[_heading=h.6y8hn7xy6ljb]{2}: its scores remained close
to zero across the evaluation settings we tested. All models are
evaluated using vLLM with temperature=1.0, repetition\_penalty=1, and
presence\_penalty=1.5. The highest score in each row is shown in bold.

\Needspace{9\baselineskip}\begingroup\small\setlength{\LTpre}{4pt}\setlength{\LTpost}{5pt}

\protect\phantomsection\label{_heading=h.6y8hn7xy6ljb}{}\captionof{table}{Zero-shot reasoning evaluations (\%). Best scores are bold.}

\nopagebreak[4]

\small\setlength{\tabcolsep}{3pt}

{\def\LTcaptype{reporttablebody} 
\begin{longtable}[]{@{}
  >{\raggedright\arraybackslash}p{(\linewidth - 8\tabcolsep) * \real{0.4201}}
  >{\raggedright\arraybackslash}p{(\linewidth - 8\tabcolsep) * \real{0.0700}}
  >{\raggedright\arraybackslash}p{(\linewidth - 8\tabcolsep) * \real{0.1700}}
  >{\raggedright\arraybackslash}p{(\linewidth - 8\tabcolsep) * \real{0.1700}}
  >{\raggedright\arraybackslash}p{(\linewidth - 8\tabcolsep) * \real{0.1700}}@{}}
\toprule\noalign{}
\begin{minipage}[b]{\linewidth}\raggedright
\textbf{Benchmark}
\end{minipage} & \begin{minipage}[b]{\linewidth}\raggedright
\textbf{Lang.}
\end{minipage} & \begin{minipage}[b]{\linewidth}\raggedright
\textbf{AliceAI-Foundation-80B-A3B-Base}
\end{minipage} & \begin{minipage}[b]{\linewidth}\raggedright
\textbf{Qwen3.5-35B-A3B-Base}
\end{minipage} & \begin{minipage}[b]{\linewidth}\raggedright
\textbf{Nemotron-3-Super-120B-A12B-Base}
\end{minipage} \\
\midrule\noalign{}
\endhead
\bottomrule\noalign{}
\endlastfoot
\multicolumn{5}{@{}>{\raggedright\arraybackslash}p{(\linewidth - 8\tabcolsep) * \real{1.0000} + 8\tabcolsep}@{}}{%
\emph{Advanced reasoning}} \\
AIME 2026 pass@32 & EN & \textbf{96.7} & \textbf{96.7} & 90.0 \\
HMMT Feb 2026 pass@32 & EN & \textbf{96.9} & 87.9 & 66.7 \\
IMO-AnswerBench pass@8 & EN & \textbf{88.7} & 84.5 & 64.5 \\
Codeforces CPP pass@8 & EN & 68.9 & \textbf{73.7} & 56.6 \\
LiveCodeBench v5-6 pass@1 & EN & \textbf{60.4} & 51.9 & 34.7 \\
LiveCodeBench v5-6 pass@8 & EN & \textbf{82.9} & 82.1 & 59.8 \\
\end{longtable}
}

\endgroup

Our model leads this comparison on IMO-AnswerBench, HMMT, and
LiveCodeBench, and ties for first on AIME.

\subsection{Availability}\label{availability}

The final pretrained checkpoint is available on Hugging Face\footnote{\url{https://huggingface.co/yandex/AliceAI-Foundation-80B-A3B-Base/blob/main/README_en.md}}.
We release the checkpoint after the reasoning stage; the intermediate
checkpoint is used only for the comparisons in this report.

\section{Evaluation Methodology}\label{evaluation-methodology}

\subsection{General Protocol}\label{general-protocol}

We evaluate each pretraining experiment across a broad range of
capabilities: factual knowledge, mathematics, reasoning, demanding
academic examinations, and long-context processing. Public benchmarks
support comparisons with other models, but do not fully cover the
capabilities relevant to our applications. Their scores can also
saturate over time, while repeated use increases the risk of overfitting
to these benchmarks. We therefore develop a substantial part of our
evaluation suite internally, guided by user needs and observed model
weaknesses.

We release WikiWebFacts and HardMultiQA together with the model to
support independent evaluation. Scores on generative benchmarks can be
sensitive to prompt wording, demonstrations, generation settings, and
grading procedures. We therefore validate new evaluation protocols
against human judgments and examine their sensitivity to small changes
in settings. For public benchmarks, we first reproduce evaluations on
models with published results. We investigate discrepancies in either
direction and select a protocol that reasonably reproduces the published
results.

\subsection{Factual Knowledge}\label{factual-knowledge}

A base model should both reason over supplied context and retain broad
knowledge of people, events, culture, science, and everyday subjects.
Examples of English-language factual knowledge benchmarks are TriviaQA,
Natural Questions, WebQuestions, and SimpleQA
{[}\hyperref[_heading=h.22q7kzru1l9p]{2},
\hyperref[_heading=h.nbdmi4tu2oiu]{3},
\hyperref[_heading=h.3rz67r67kuuq]{4},
\hyperref[_heading=h.2p67eldlh7ec]{5}{]}. Far fewer publicly available
benchmarks exist for Russian, and simply translating English-language
datasets into Russian does not address this gap. The knowledge most
relevant to users varies considerably across countries and languages,
particularly in areas such as history, geography, and culture.
Translated benchmarks fail to capture these differences.

To evaluate factual knowledge, we developed several benchmarks of our
own, two of which we are releasing publicly. The first, WikiWebFacts,
focuses on relatively short factual questions with clear, unambiguous
answers. The second, HardMultiQA, was developed later and deliberately
designed to be more challenging, with a distribution of questions that
more closely reflects real-world user interactions with language models.

\subsubsection{WikiWebFacts}\label{wikiwebfacts}

WikiWebFacts was designed to cover a broad range of topics without
overrepresenting obscure facts of limited interest to users.

The \textbf{Wiki subset} is based on Russian-language online
encyclopedias. Rather than randomly sampling articles, we selected those
that frequently appeared in Yandex Search results, prioritizing topics
relevant to users. We used a language model to extract facts from the
selected articles and generate question--answer pairs. Human annotators
then checked that the questions were valid, the answers unambiguous, and
the generated pairs consistent with the source articles.

The \textbf{Web subset} draws on aggregated, anonymized user search
queries. We selected queries likely to have short factual answers,
converted them into benchmark items, and had them reviewed by human
annotators. This complements encyclopedic knowledge with questions
reflecting everyday information needs.

The benchmark covers topics including history, literature, law, and the
exact sciences (Figure 1). Questions in domains requiring specialized
expertise, such as medicine and law, were also reviewed by
subject-matter experts. Figure 2 illustrates the benchmark construction
pipeline.

\par\addvspace{8pt}\noindent\begin{minipage}{\linewidth}\centering

\includegraphics[width=\linewidth,height=.46\textheight,keepaspectratio]{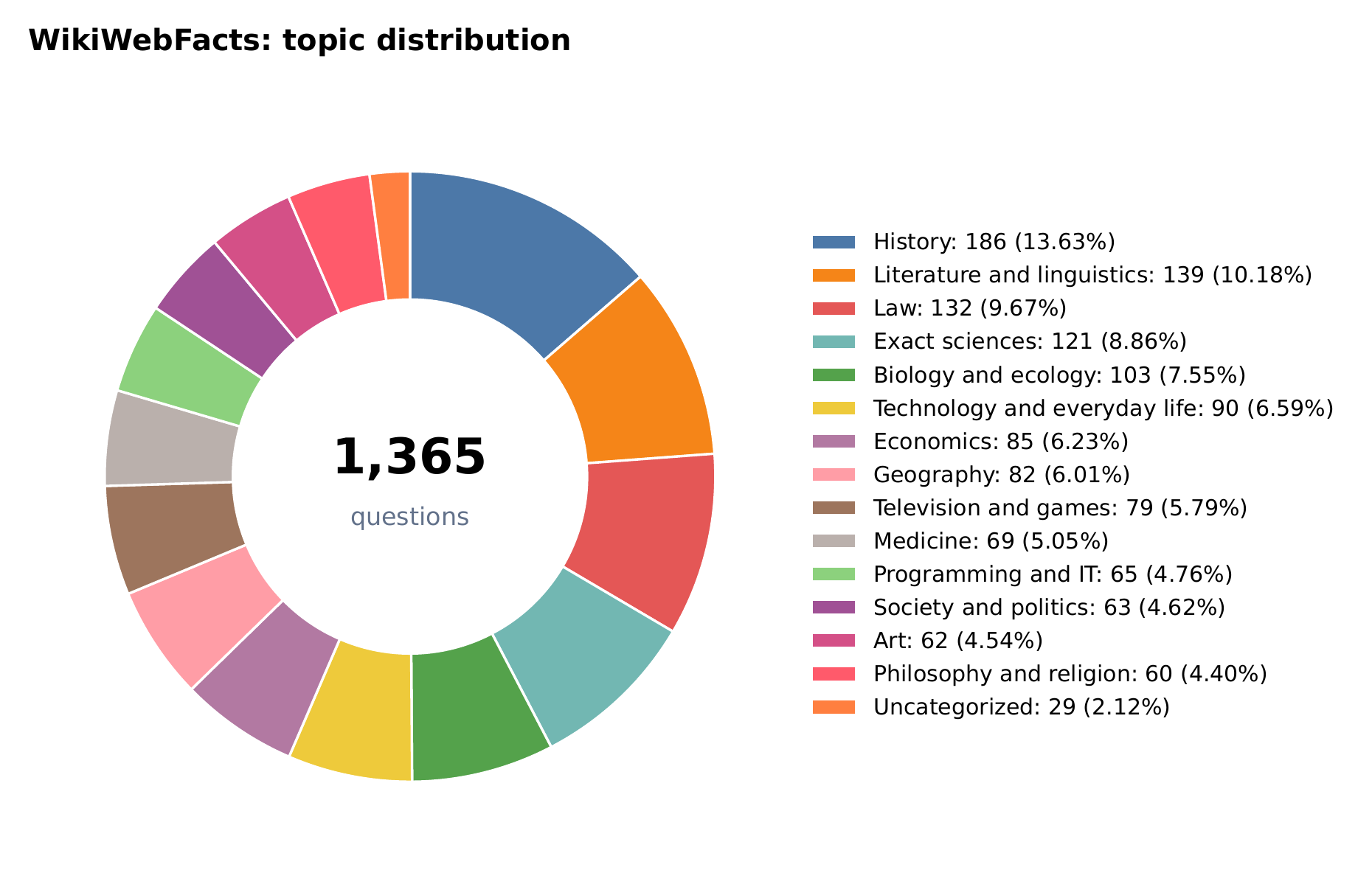}

\par

\captionof{figure}{Topic distribution in WikiWebFacts.}

\end{minipage}\par\addvspace{7pt}

Most questions have short, unambiguous answers, making automated
evaluation with an LLM-as-a-judge (LLMaJ) practical. We evaluate base
models using five in-context examples and greedy decoding. Table 3
presents examples translated into English for readability, while
evaluation uses the original Russian questions.

\Needspace{9\baselineskip}\begingroup\small\setlength{\LTpre}{4pt}\setlength{\LTpost}{5pt}

\captionof{table}{Translated examples from WikiWebFacts. Evaluation uses
the original Russian questions.}

\nopagebreak[4]

\small\setlength{\tabcolsep}{3pt}

{\def\LTcaptype{reporttablebody} 
\begin{longtable}[]{@{}
  >{\raggedright\arraybackslash}p{(\linewidth - 4\tabcolsep) * \real{0.2401}}
  >{\raggedright\arraybackslash}p{(\linewidth - 4\tabcolsep) * \real{0.5400}}
  >{\raggedright\arraybackslash}p{(\linewidth - 4\tabcolsep) * \real{0.2200}}@{}}
\toprule\noalign{}
\begin{minipage}[b]{\linewidth}\raggedright
\textbf{Question source}
\end{minipage} & \begin{minipage}[b]{\linewidth}\raggedright
\textbf{Question}
\end{minipage} & \begin{minipage}[b]{\linewidth}\raggedright
\textbf{Answer}
\end{minipage} \\
\midrule\noalign{}
\endhead
\bottomrule\noalign{}
\endlastfoot
Encyclopedia article: Jalapeño & Which city gave the jalapeño pepper its
name? & Xalapa \\
Encyclopedia article: Deathly Hallows & Which magical object did the
eldest Peverell brother receive? & The Elder Wand \\
Search query: Fathers and Sons cannot live without principles who &
Which character in ``Fathers and Sons'' believes that one cannot live
without principles? & Pavel Petrovich Kirsanov \\
Search query: how many genders in Latin & How many grammatical genders
do Latin nouns have? & 3 \\
\end{longtable}
}

\endgroup

\par\addvspace{8pt}\noindent\begin{minipage}{\linewidth}\centering

\includegraphics[width=\linewidth,height=.46\textheight,keepaspectratio]{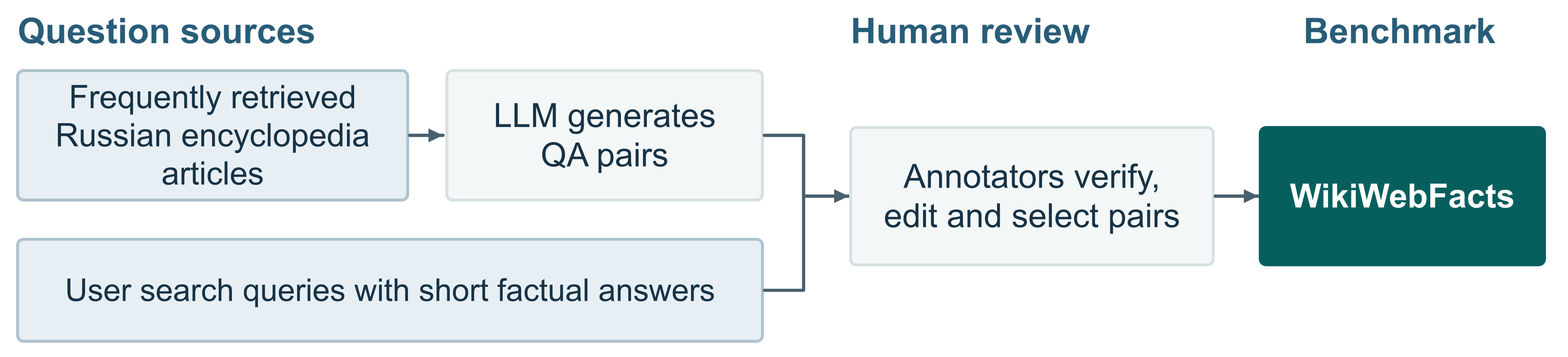}

\par

\captionof{figure}{WikiWebFacts construction pipeline.}

\end{minipage}\par\addvspace{7pt}

WikiWebFacts is not designed to be particularly challenging for
state-of-the-art models. Instead, it provides a stable, interpretable
measure for evaluating from-scratch training runs, intermediate
checkpoints, and smaller models. Scores tend to saturate for the
strongest models, while access to web search makes many questions
substantially easier. We release the dataset, along with separate
evaluation protocols for base and instruction-tuned models, on Hugging
Face\footnote{\url{https://huggingface.co/datasets/yandex/WikiWebFacts}}.

\subsubsection{HardMultiQA}\label{hardmultiqa}

HardMultiQA addresses two limitations of WikiWebFacts. Encyclopedic
knowledge does not fully reflect the topics users discuss with language
models. Search queries also tend to be shorter than LLM requests and
often require different types of answers. Users may expect detailed
responses that involve recalling, comparing, analyzing, or synthesizing
multiple facts.

We therefore used anonymized queries submitted to Alice AI as a source
of topics. Rather than using these queries directly, many of which
lacked clear correctness criteria, editors identified the underlying
knowledge and wrote new questions with explicit grading criteria. To
cover different types of responses, we included several question
formats:

\begin{itemize}
\item
  Short factual questions with unambiguous answers;
\item
  Multiple-choice questions with one or more correct options;
\item
  Questions requiring lists of entities or facts;
\item
  Questions requiring the identification of factual errors in short
  passages.
\end{itemize}

Multiple-choice questions become harder to solve by elimination when the
number of correct options is unspecified. For list questions, we
distinguish between exhaustive sets, where missing or extra items count
as errors, and core sets, where additional correct items are allowed.

Editors chose the most appropriate format for each topic and tested
question difficulty against internal model checkpoints and external
models. Questions consistently solved by all models could be made more
challenging or refocused on another aspect of the topic. Other editors
independently answered the questions, with disagreements prompting
further review. Domain-specific questions also underwent expert review.

The variety of question formats required more detailed grading
protocols. As with WikiWebFacts, we used a larger model to audit each
task type and iteratively refined the grading prompts. Figures 3 and 4
show the construction pipeline and question distributions, respectively.
Table 4 provides examples translated into English.

\par\addvspace{8pt}\noindent\begin{minipage}{\linewidth}\centering

\includegraphics[width=\linewidth,height=.46\textheight,keepaspectratio]{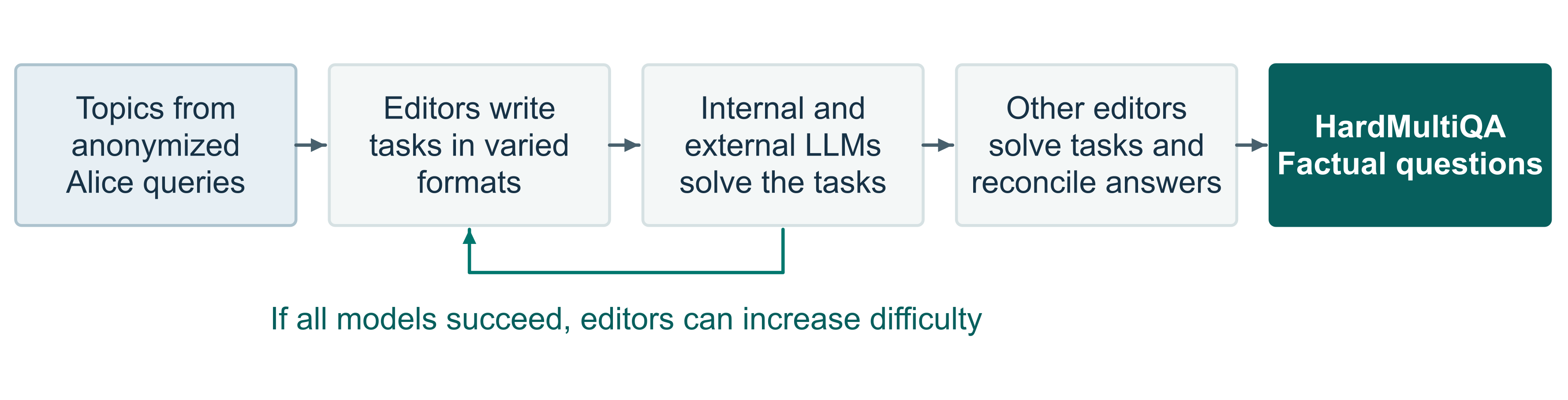}

\par

\captionof{figure}{HardMultiQA construction pipeline.}

\end{minipage}\par\addvspace{7pt}

\par\addvspace{8pt}\noindent\begin{minipage}{\linewidth}\centering

\includegraphics[width=\linewidth,height=.46\textheight,keepaspectratio]{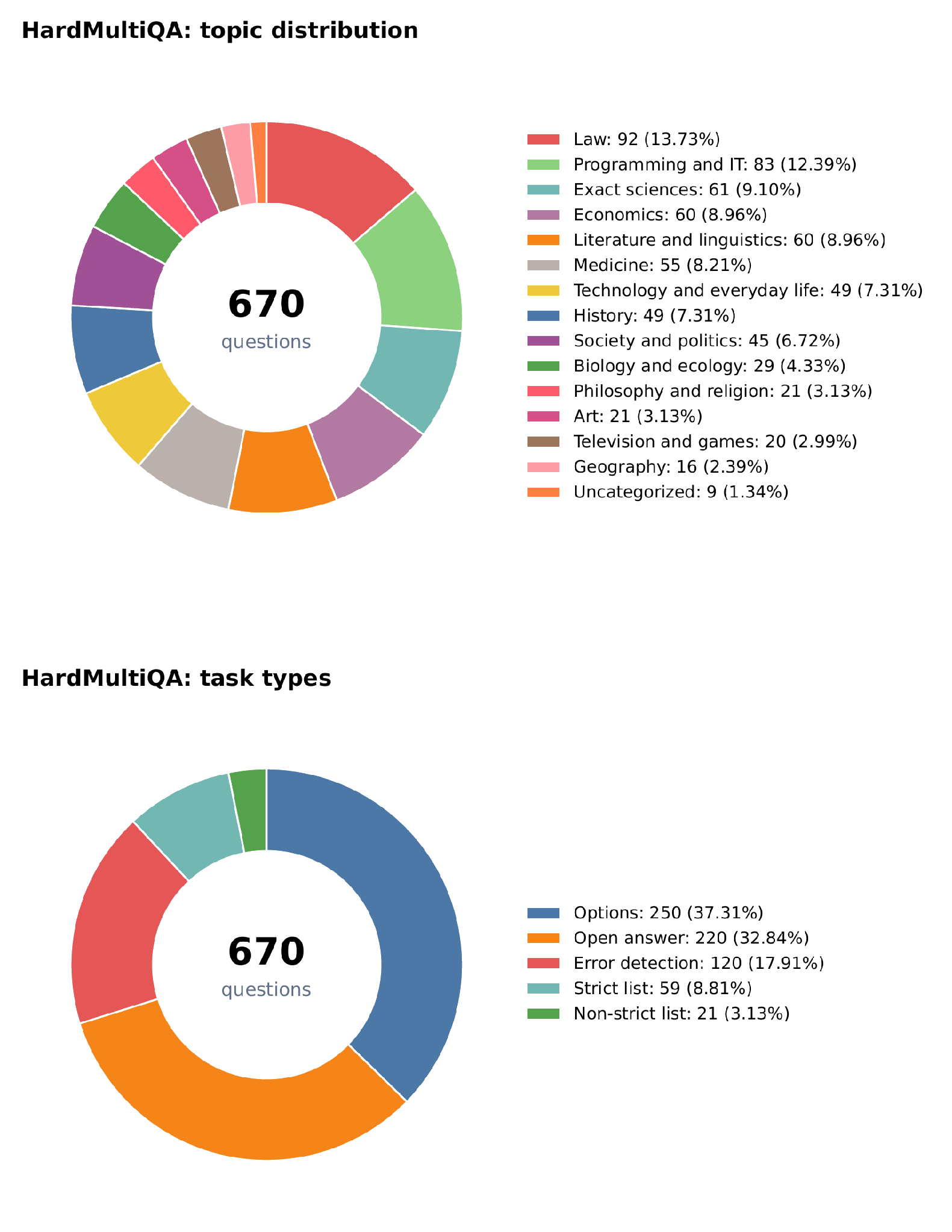}

\par

\captionof{figure}{HardMultiQA distributions by topic (top) and task type
(bottom).}

\end{minipage}\par\addvspace{7pt}

\Needspace{9\baselineskip}\begingroup\small\setlength{\LTpre}{4pt}\setlength{\LTpost}{5pt}

\captionof{table}{Translated examples from HardMultiQA. Evaluation uses the
original Russian questions.}

\nopagebreak[4]

\small\setlength{\tabcolsep}{3pt}

{\def\LTcaptype{reporttablebody} 
\begin{longtable}[]{@{}
  >{\raggedright\arraybackslash}p{(\linewidth - 6\tabcolsep) * \real{0.2401}}
  >{\raggedright\arraybackslash}p{(\linewidth - 6\tabcolsep) * \real{0.4399}}
  >{\raggedright\arraybackslash}p{(\linewidth - 6\tabcolsep) * \real{0.1801}}
  >{\raggedright\arraybackslash}p{(\linewidth - 6\tabcolsep) * \real{0.1400}}@{}}
\toprule\noalign{}
\begin{minipage}[b]{\linewidth}\raggedright
\textbf{Source LLM query}
\end{minipage} & \begin{minipage}[b]{\linewidth}\raggedright
\textbf{Question}
\end{minipage} & \begin{minipage}[b]{\linewidth}\raggedright
\textbf{Answer}
\end{minipage} & \begin{minipage}[b]{\linewidth}\raggedright
\textbf{Type}
\end{minipage} \\
\midrule\noalign{}
\endhead
\bottomrule\noalign{}
\endlastfoot
Why are capybaras so large? & Order these cavies by decreasing size:
capybara, guinea pig, mara, rock cavy. & Capybara, mara, rock cavy,
guinea pig & Short question \\
Why has the Barsova Gora area been continuously inhabited for 7.000
years? & Three names can refer to the same people. Identify the others:
1. Ostyaks; 2. Evenks; 3. Khanty; 4. Ugrians; 5. Tofalars. & 2, 5 &
Options \\
Write four pages of recommendations for developing new technical
specifications for nut paste. & Correct any errors; if there are none,
say so. Urbech is a natural nut paste made from shells, seeds, kernels,
and the oil released in processing. & Remove the word shells. & Error
detection \\
Gamification in teaching English. & What does PBL mean in educational
gamification? & Points, Badges, Leaderboards & List \\
\end{longtable}
}

\endgroup

HardMultiQA is substantially more challenging than WikiWebFacts and
better distinguishes strong models. Even with web search, many questions
require comparing facts, distinguishing between similar options, or
verifying claims. We also evaluate model robustness to distracting
context (Section 5.3). The benchmark can also be used to measure
appropriate abstention separately from incorrect answers.

For base models, we use five in-context examples and greedy decoding.
HardMultiQA serves as our primary factual-knowledge benchmark for larger
models, while WikiWebFacts remains useful for smaller models and early
checkpoints, where HardMultiQA may be too challenging. We release the
dataset and evaluation protocols for both base and instruction-tuned
models on Hugging Face\footnote{\url{https://huggingface.co/datasets/yandex/HardMultiQA}}..

\subsection{Educational and Expert
Benchmarks}\label{educational-and-expert-benchmarks}

Beyond the factual-knowledge benchmarks described above, we use separate
datasets to evaluate model performance on educational and expert-level
tasks. Both were designed around real-world user scenarios: we started
with queries submitted to Alice AI, identified the most relevant task
types, and involved subject-matter experts in preparing and validating
answers.

\subsubsection{Educational
Benchmarks}\label{educational-benchmarks}

Educational queries are an important use case for Alice AI, so we
evaluate model performance on school- and university-level tasks
separately. We built these benchmarks using aggregated, anonymized user
queries, selecting those related to school and university assignments.
Annotators with subject-specific expertise reviewed the problem
statements and prepared reference answers. Unlike conventional exam
benchmarks, the task distribution reflects what users actually ask the
model rather than the structure of a particular exam. We focus on
Russian language, literature, history, English, school mathematics, and
university mathematics.

Some queries may originate from textbooks, exams, or other public
sources. To avoid inflated evaluation scores due to data contamination,
we searched the training corpus for benchmark questions and answers
using n-gram overlap and removed matching training data.

\subsubsection{Expert Benchmarks}\label{expert-benchmarks}

Error analysis on HardMultiQA highlighted specialized professional
domains as an important area for improvement. Although HardMultiQA
includes such questions, its broad scope makes it difficult to assess
model performance within individual domains. We therefore developed
separate benchmarks for medicine and law, prioritizing topics that
frequently appear in user queries and on which the current model makes
significant errors.

To build each benchmark, we start with aggregated, anonymized queries to
Alice AI and generate answers using several in-house models,
experimental checkpoints, and external models. This helps avoid focusing
on the weaknesses of a single model. Domain experts identify factual
errors and inaccuracies in the generated responses and use them to
construct questions with unambiguous answers that can be reliably graded
by an LLM judge. Question formats include direct factual questions,
theoretical questions, multiple-choice questions, and case-based
problems.

Each question is then answered independently by a second expert without
access to the author\textquotesingle s answer. A third expert reviews
disagreements and revises the question wording or reference answer when
needed. This process corrected the initial annotations in approximately
20\% of cases, often resolving ambiguities or missing conditions. Figure
5 illustrates the benchmark construction and review process.

\par\addvspace{8pt}\noindent\begin{minipage}{\linewidth}\centering

\includegraphics[width=\linewidth,height=.46\textheight,keepaspectratio]{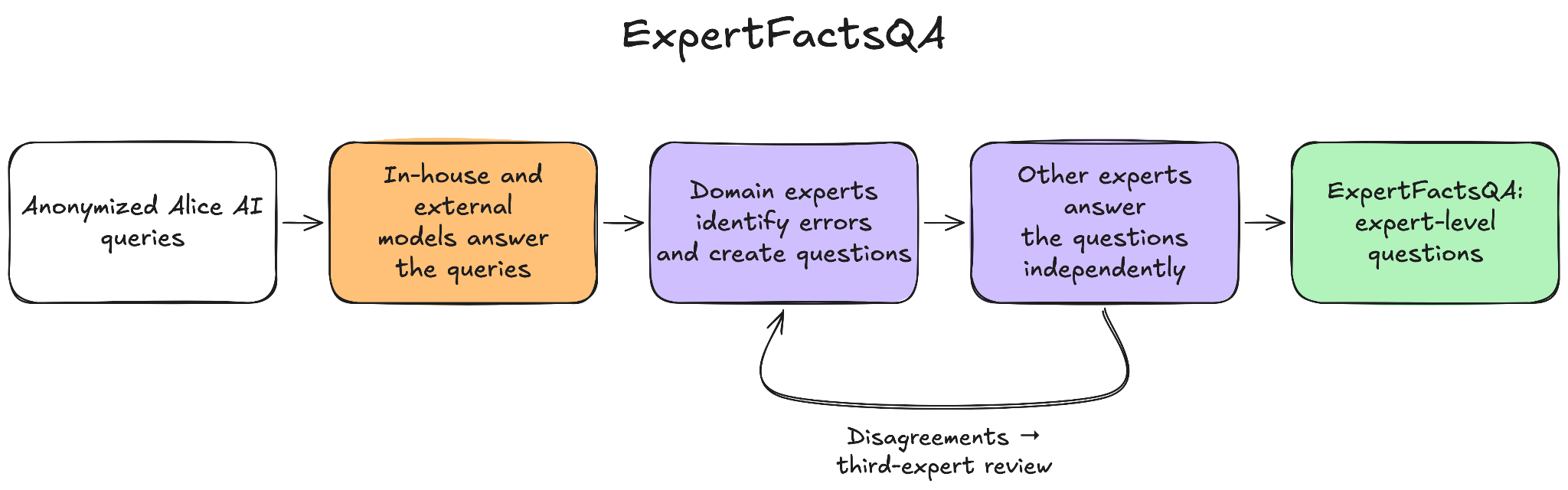}

\par

\captionof{figure}{ExpertFactsQA construction and independent specialist
review.}

\end{minipage}\par\addvspace{7pt}

\subsection{Reasoning Benchmarks}\label{reasoning-benchmarks}

\subsubsection{Academic, Mathematical, and Coding
Benchmarks}\label{academic-mathematical-and-coding-benchmarks}

We use MMLU-Pro~{[}\hyperref[_heading=h.gturprr44st]{6}{]} and
SuperGPQA~{[}\hyperref[_heading=h.xhnjvb9w2dn]{7}{]} for broad academic
knowledge and reasoning; SuperGPQA is especially useful for its subject
coverage. The evaluation prompts include demonstrations with reasoning
traces. For Russian, we additionally use an internal benchmark based on
Part A of the Unified State Examination (EGE) across several subjects.

MATH-500, drawn from the MATH
benchmark~{[}\hyperref[_heading=h.5r81yw8g9oeh]{8}{]}, provides a
standard mathematical comparison, complemented by the Russian school and
university benchmarks described above. For programming, we use
BigCodeBench~{[}\hyperref[_heading=h.bklrgfrfrlut]{9}{]} and
LiveCodeBench~{[}\hyperref[_heading=h.hih2dyk9dnms]{10}{]}.
HumanEval~{[}\hyperref[_heading=h.8g5tcyk6mw3j]{11}{]} and
MBPP~{[}\hyperref[_heading=h.l5sclznlh0k3]{12}{]} are comparatively easy
for models of this scale, and their scores are sensitive to prompt
format and demonstration count.

BigCodeBench tests practical Python programming with libraries. We found
that some supplied tests did not adequately check the intended behavior.
Programming specialists investigated failures and corrected or extended
tests for a subset of tasks. Our results therefore use this modified
test suite. LiveCodeBench requires finding an algorithm and implementing
it. We use reasoning demonstrations because a protocol that excludes
reasoning is less natural for these tasks and less informative about
algorithmic problem solving.

\subsubsection{Prompting and Sampling for Base-Model
Reasoning}\label{prompting-and-sampling-for-base-model-reasoning}

Reasoning evaluations proved particularly sensitive to demonstrations.
Both their quality and their length can substantially change scores for
internal and open base models.
Figure~\hyperref[_heading=h.6c3azz3w9p1t]{6} illustrates this effect on
MATH-500: individual models respond differently, but longer
demonstration traces generally improve performance. For standard
reasoning benchmarks, we selected demonstrations that produced
consistently strong results across several models.

\par\addvspace{8pt}\noindent\begin{minipage}{\linewidth}\centering

\protect\phantomsection\label{_heading=h.6c3azz3w9p1t}{}

\includegraphics[width=\linewidth,height=.46\textheight,keepaspectratio]{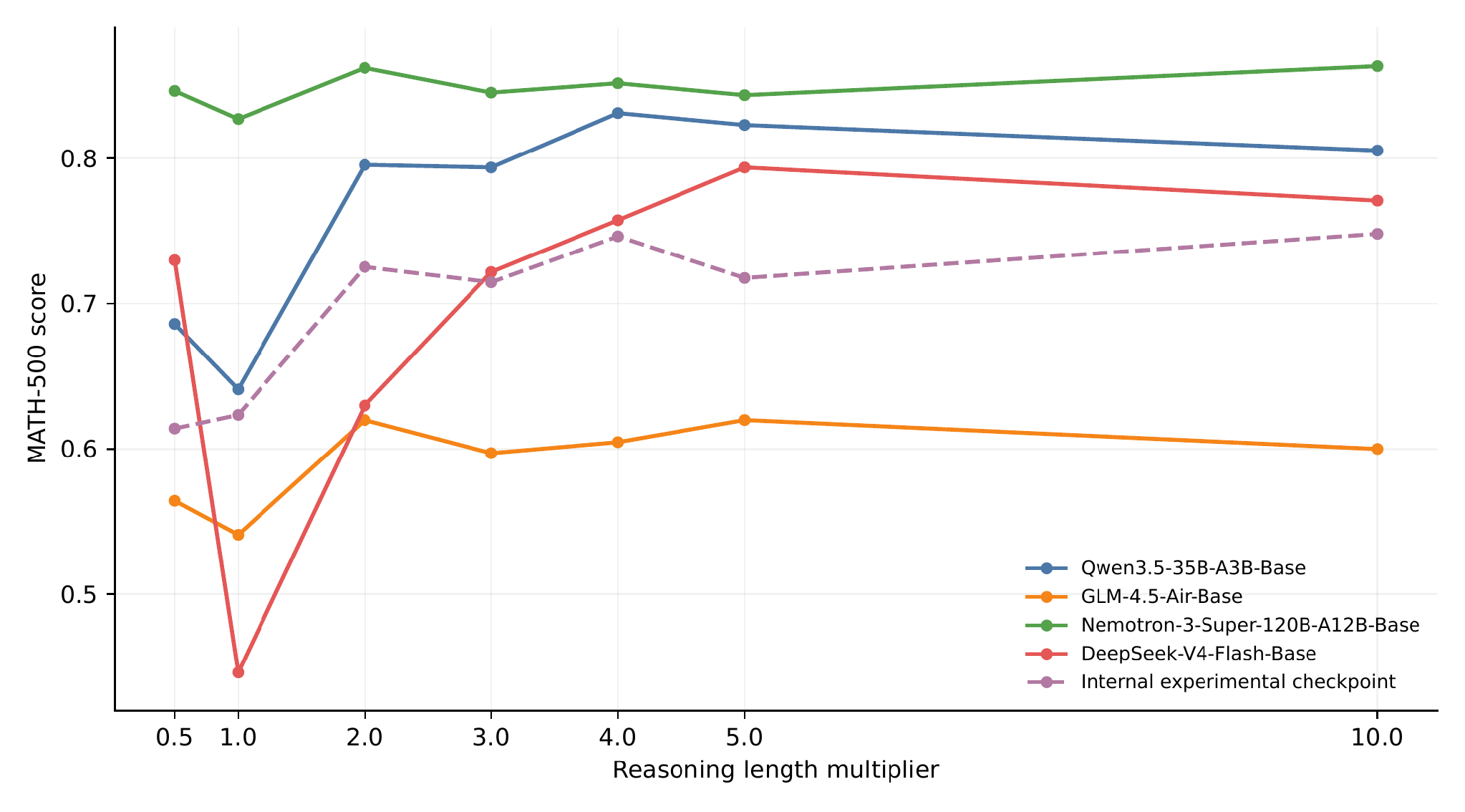}

\par

\captionof{figure}{MATH-500 performance as a function of reasoning length
in few-shot demonstrations. The horizontal axis gives the length
multiplier relative to the reference demonstrations. Curves are
digitized from the published figure and are approximate.}

\end{minipage}\par\addvspace{7pt}

On harder problems, demonstrations can constrain the model's reasoning
length. We therefore use zero-shot prompts and allow the model to choose
how extensively to reason. Because base models often produce formatting
errors and results vary substantially between samples, we use pass@\(k\)
with many independent generations. Related work also studies base-model
reasoning using large-\(k\)
pass@\(k\)~{[}\hyperref[_heading=h.xpemf2syv2y2]{13}{]}; NVIDIA reports
AIME 2024 pass@32 for
Nemotron-3-Super-120B-A12B-Base~{[}\hyperref[_heading=h.hzsdvcm30v1u]{14}{]}.

We apply this protocol to Codeforces competitive programming and the
AIME, HMMT, and IMO-AnswerBench mathematical benchmarks. It is
especially informative after the reasoning stage. It also works well for
Qwen3.5-35B-A3B-Base and Nemotron-3-Super-120B-A12B-Base, which can
solve difficult tasks without demonstrations even when successful
samples are infrequent. GLM-4.5-Air and DeepSeek-V4-Flash perform
substantially worse in this zero-shot setting, making it less
informative for those models.

\subsection{Long-Context Benchmarks}\label{long-context-benchmarks}

The long-context evaluations reported here use up to 128K tokens.
Training includes an intermediate 256K context-extension stage, followed
by a final stage at 128K (Table~\hyperref[_heading=h.hfd7a7nfvjmt]{5});
the reported scores do not establish performance at 256K. We emphasize
practical use of long conversations and multiple documents rather than
isolated fact retrieval.
LongMemEval~{[}\hyperref[_heading=h.tagj5qdjldqz]{15}{]} tests locating,
combining, and updating information in a long dialogue history. Our
base-model prompt reuses one long dialogue for four questions, retaining
demonstrations within the available context budget.

We also extend FinQA~{[}\hyperref[_heading=h.m9vokux8vld1]{16}{]} to
128K tokens. The original benchmark requires extracting information from
financial reports and performing calculations. Our version combines many
documents in one context, requiring the model to identify relevant
information before answering.

\subsection{Agentic Benchmarks}\label{agentic-benchmarks}

Base models are not directly adapted to agentic interaction. Before
evaluation, all compared base models therefore receive SFT on the same
agentic dataset. Full RL improves absolute scores further, but important
differences between pretraining experiments are already visible after
SFT.

Our suite includes
\(\tau^{2}\)-bench~{[}\hyperref[_heading=h.sdrn8c2szi6m]{17}{]} and
VitaBench~{[}\hyperref[_heading=h.82csxetqjvhn]{18}{]} for interaction
with users and services,
DeepPlanning~{[}\hyperref[_heading=h.hhxjg5l63n7d]{19}{]} for planning,
and SWE-bench Verified~{[}\hyperref[_heading=h.ox5e9jqvoq4r]{20}{]} with
the OpenHands framework~{[}\hyperref[_heading=h.r0prcmdlmmzh]{21}{]} for
software engineering. We also developed proxy benchmarks that assess
individual steps of agent trajectories when a base model cannot yet
complete an entire interaction.

\section{Model and Training
Overview}\label{model-and-training-overview}

The model is an MoE Transformer with 80B total parameters and
approximately 3B active parameters per token. It contains 48 Transformer
blocks with hidden size 2048 and one multi-token-prediction layer. Each
MoE layer has 512 routed experts, of which ten are selected per token,
and one shared expert. Routing uses the auxiliary-loss-free balancing
approach of DeepSeek-V3~{[}\hyperref[_heading=h.okq3e2yo5bmy]{22}{]}.

Of the 48 attention layers, 36 use Kimi Delta Attention (KDA) and 12 use
full attention. One full-attention layer follows every three KDA layers.
Attention Residuals (AttnRes) replaces conventional residual summation
with learned weighting of representations across depth.
Section~\hyperref[data-architecture-and-hyperparameter-ablations]{4}
details these choices.

Training proceeds through four stages
(Table~\hyperref[_heading=h.hfd7a7nfvjmt]{5}). General pretraining uses
17.5T tokens and an 8K-token context. The next two stages extend context
first to 32K and then to 256K, with reduced learning rate and batch
size. Hyperparameters for these stages were selected experimentally.
Extending the 32K stage did not substantially improve long-context
benchmarks, so most of the context-extension budget was allocated to the
256K stage.

\Needspace{9\baselineskip}\begingroup\small\setlength{\LTpre}{4pt}\setlength{\LTpost}{5pt}

\protect\phantomsection\label{_heading=h.hfd7a7nfvjmt}{}\captionof{table}{The four sequential training stages. Batch size is measured in tokens.}

\nopagebreak[4]

\small\setlength{\tabcolsep}{3pt}

{\def\LTcaptype{reporttablebody} 
\begin{longtable}[]{@{}
  >{\raggedright\arraybackslash}p{(\linewidth - 8\tabcolsep) * \real{0.2000}}
  >{\raggedright\arraybackslash}p{(\linewidth - 8\tabcolsep) * \real{0.2000}}
  >{\raggedright\arraybackslash}p{(\linewidth - 8\tabcolsep) * \real{0.2000}}
  >{\raggedright\arraybackslash}p{(\linewidth - 8\tabcolsep) * \real{0.2000}}
  >{\raggedright\arraybackslash}p{(\linewidth - 8\tabcolsep) * \real{0.2000}}@{}}
\toprule\noalign{}
\begin{minipage}[b]{\linewidth}\raggedright
\textbf{Parameter}
\end{minipage} & \begin{minipage}[b]{\linewidth}\raggedright
\textbf{General}
\end{minipage} & \begin{minipage}[b]{\linewidth}\raggedright
\textbf{32K extension}
\end{minipage} & \begin{minipage}[b]{\linewidth}\raggedright
\textbf{256K extension}
\end{minipage} & \begin{minipage}[b]{\linewidth}\raggedright
\textbf{Reasoning}
\end{minipage} \\
\midrule\noalign{}
\endhead
\bottomrule\noalign{}
\endlastfoot
Training tokens & 17.5T & 50B & 240B & 280B \\
Learning rate & 8.2e-4 & 4.1e-05 & 4.1e-05 & 4.1e-05 \\
Batch tokens & 33M & 16M & 16M & 16M \\
Context tokens & 8192 & 32768 & 262144 & 131072 \\
\end{longtable}
}

\endgroup

The final stage develops reasoning and agentic interaction
(Section~\hyperref[why-a-separate-reasoning-stage]{7}). We reduce
context from 256K to 128K to save computation. High-quality examples
with longer reasoning traces are relatively rare, while the longest
trajectories more often contain noise and redundant steps. A 128K
context accommodates most useful examples and allows the remaining
budget to cover more training examples.

\section{Architecture and Training
Configuration}\label{architecture-and-training-configuration}

\subsection{Data, Architecture, and Hyperparameter
Ablations}\label{data-architecture-and-hyperparameter-ablations}

We constructed the baseline mixture from the pretraining corpus used for
YandexGPT 5 Base\footnote{\url{https://huggingface.co/yandex/YandexGPT-5-Lite-8B-pretrain}},
incorporating data and preparation methods developed for our previous
Alice AI LLM 235B Base\footnote{\url{https://habr.com/ru/companies/yandex/articles/974594/}}
{[}1{]}.

We developed the data mixture, architecture, and hyperparameter scaling
laws in parallel. To quantify their contributions, we trained four
configurations independently from scratch on 2 trillion tokens each,
using the first-stage training setup. Starting from a baseline with the
Qwen3-Next architecture, we introduced the data mixture, architecture,
and hyperparameters used for AliceAI-Foundation-80B-A3B-Base in that
order, retaining earlier changes at each step. Table 6 summarizes the
configurations.

\Needspace{9\baselineskip}\begingroup\small\setlength{\LTpre}{4pt}\setlength{\LTpost}{5pt}

\captionof{table}{Configurations used in the sequential ablation study.}

\nopagebreak[4]

\small\setlength{\tabcolsep}{3pt}

{\def\LTcaptype{reporttablebody} 
\begin{longtable}[]{@{}
  >{\raggedright\arraybackslash}p{(\linewidth - 6\tabcolsep) * \real{0.2090}}
  >{\raggedright\arraybackslash}p{(\linewidth - 6\tabcolsep) * \real{0.2621}}
  >{\raggedright\arraybackslash}p{(\linewidth - 6\tabcolsep) * \real{0.2605}}
  >{\raggedright\arraybackslash}p{(\linewidth - 6\tabcolsep) * \real{0.2444}}@{}}
\toprule\noalign{}
\begin{minipage}[b]{\linewidth}\raggedright
\textbf{Configuration}
\end{minipage} & \begin{minipage}[b]{\linewidth}\raggedright
\textbf{Architecture}
\end{minipage} & \begin{minipage}[b]{\linewidth}\raggedright
\textbf{Pretraining data}
\end{minipage} & \begin{minipage}[b]{\linewidth}\raggedright
\textbf{Hyperparameters}
\end{minipage} \\
\midrule\noalign{}
\endhead
\bottomrule\noalign{}
\endlastfoot
Baseline & Qwen3-Next & Baseline mixture & Baseline settings \\
+ Data & Qwen3-Next & AliceAI-Foundation-80B-A3B-Base & Baseline
settings \\
+ Architecture & AliceAI-Foundation-80B-A3B-Base &
AliceAI-Foundation-80B-A3B-Base & Baseline settings \\
+ Hyperparameters & AliceAI-Foundation-80B-A3B-Base &
AliceAI-Foundation-80B-A3B-Base & AliceAI-Foundation-80B-A3B-Base \\
\end{longtable}
}

\endgroup

Adjacent comparisons measure the contributions of the updated corpus
(Baseline → + Data), architectural changes (+ Data → + Architecture),
and hyperparameters selected using our scaling laws (+ Architecture → +
Hyperparameters).

\Needspace{9\baselineskip}\begingroup\small\setlength{\LTpre}{4pt}\setlength{\LTpost}{5pt}

\captionof{table}{Ablation results (\%). Configurations are defined in
Table 6. The best result in each row is shown in bold.}

\nopagebreak[4]

\small\setlength{\tabcolsep}{3pt}

{\def\LTcaptype{reporttablebody} 
\begin{longtable}[]{@{}
  >{\raggedright\arraybackslash}p{(\linewidth - 10\tabcolsep) * \real{0.2626}}
  >{\raggedright\arraybackslash}p{(\linewidth - 10\tabcolsep) * \real{0.0757}}
  >{\raggedright\arraybackslash}p{(\linewidth - 10\tabcolsep) * \real{0.1257}}
  >{\raggedright\arraybackslash}p{(\linewidth - 10\tabcolsep) * \real{0.1112}}
  >{\raggedright\arraybackslash}p{(\linewidth - 10\tabcolsep) * \real{0.1772}}
  >{\raggedright\arraybackslash}p{(\linewidth - 10\tabcolsep) * \real{0.2272}}@{}}
\toprule\noalign{}
\begin{minipage}[b]{\linewidth}\raggedright
\textbf{Benchmark}
\end{minipage} & \begin{minipage}[b]{\linewidth}\centering
\textbf{Lang.}
\end{minipage} & \begin{minipage}[b]{\linewidth}\centering
\textbf{Baseline}
\end{minipage} & \begin{minipage}[b]{\linewidth}\centering
\textbf{+ Data}
\end{minipage} & \begin{minipage}[b]{\linewidth}\centering
\textbf{+ Architecture}
\end{minipage} & \begin{minipage}[b]{\linewidth}\centering
\textbf{+ Hyperparameters}
\end{minipage} \\
\midrule\noalign{}
\endhead
\bottomrule\noalign{}
\endlastfoot
HardMultiQA & RU & 51.19 & 56.10 & 58.00 & \textbf{60.00} \\
ExpertFactsQA Medicine & RU & 55.27 & \textbf{59.20} & 57.90 & 58.40 \\
WikiWebFacts, Wiki subset & RU & 76.51 & 80.30 & 79.80 &
\textbf{81.30} \\
ExpertFactsQA Law & RU & 38.04 & 44.70 & \textbf{49.70} & 49.50 \\
\emph{Group mean} & & 55.25 & 60.08 & 61.35 & \textbf{62.30} \\
MATH-500 & EN & 56.39 & 63.10 & 63.90 & \textbf{65.70} \\
EduBench Math & RU & 64.92 & 66.30 & 64.90 & \textbf{69.50} \\
Math Textbooks & RU & 85.19 & 88.90 & 87.40 & \textbf{91.00} \\
\emph{Group mean} & & 68.83 & 72.77 & 72.07 & \textbf{75.40} \\
MBPP 1-shot pass@1 & EN & 72.55 & 75.40 & 78.10 & \textbf{78.40} \\
MBPP 1-shot pass@5 & EN & 83.59 & 85.00 & \textbf{88.80} & 87.40 \\
HumanEval 1-shot pass@1 & EN & 64.66 & 73.00 & \textbf{73.30} & 71.80 \\
HumanEval 1-shot pass@5 & EN & 77.91 & 83.40 & \textbf{87.70} & 85.30 \\
\emph{Group mean} & & 74.68 & 79.20 & \textbf{81.98} & 80.73 \\
MMLU-Pro CoT & EN & 52.70 & 55.90 & \textbf{60.90} & 56.80 \\
MMLU-Pro CoT & RU & 54.00 & 55.60 & 57.20 & \textbf{59.30} \\
\emph{Group mean} & & 53.35 & 55.75 & \textbf{59.05} & 58.05 \\
\emph{Overall macro-average} & & 63.03 & 66.95 & 68.61 &
\textbf{69.12} \\
\end{longtable}
}

\endgroup

Table 7 shows that all three components improve overall performance, as
measured by the macro-average. Updating the data mixture improves scores
across all four task groups and all 13 reported metrics. The
architectural changes provide the largest gains in programming and
reasoning, while also improving factual knowledge. The revised
hyperparameters primarily improve mathematics, with additional gains on
factual tasks. The configuration incorporating all three components
achieves the highest macro-average and the best score on seven of the 13
metrics.

\subsection{Architectural Choices and Training
Stability}\label{architectural-choices-and-training-stability}

\subsubsection{Stabilizing MoE
Activations}\label{stabilizing-moe-activations}

A preliminary run at the learning rate predicted by our scaling law
exhibited a sharp loss spike accompanied by rapid growth in the MoE
down-projection outputs. We did not observe the same instability in
smaller models. However, reducing the number of layers while retaining
the original width and routing top-\(k\) reproduced the failure. We
first investigated whether changes to expert routing could resolve the
instability.

\paragraph{Routing and clipped
SwiGLU.}\label{routing-and-clipped-swiglu.}

We tested the router-logit Z-loss from
ST-MoE~{[}\hyperref[_heading=h.gowrhae9z46n]{23}{]} and replaced the
auxiliary load-balancing loss with the dynamically updated routing
biases used in DeepSeek-V3~{[}\hyperref[_heading=h.okq3e2yo5bmy]{22}{]}.
Neither change eliminated the instability, although DeepSeek routing
improved benchmark performance and was retained. We also tested
hard-clipped SwiGLU as implemented in
GPT-OSS~{[}\hyperref[_heading=h.atq2c0vnb62o]{24}{]}. Values beyond the
clipping threshold have zero derivative, blocking their gradient
contributions to the MLP input projections. This variant reduced
benchmark performance, so we did not adopt it.

\paragraph{Gated RMSNorm.}\label{gated-rmsnorm.}

Following Qiu et al. {[}\hyperref[_heading=h.hs7ep7z6bkcg]{25}{]}, we
investigated explicit feature-wise scaling after normalization. Large
persistent components of the residual stream can dominate the
denominator of RMSNorm, implicitly controlling the scale of other
features. Gated RMSNorm provides an explicit mechanism:

\[y' = \sigma\left( W_{up}\left( swish\left( W_{down}(y) \right) \right) \right) \odot y.\quad\quad(1)\]

We applied it to all normalization operations except query/key
normalization in KDA and full-attention layers. In particular, KDA
updates assume unit-\(L_{2}\)-norm keys, so this normalization must be
preserved. Activation magnitudes decreased, but outliers remained.
Replacing SwiGLU with GLU, as recommended in the original work,
stabilized activations but reduced benchmark performance. Combining
Gated RMSNorm with AttnRes stabilized activations without an observed
performance loss, but the additional projections slowed training by
approximately 5\%. We therefore sought a less expensive solution.

\paragraph{Z-loss on MoE outputs.}\label{z-loss-on-moe-outputs.}

We ultimately applied Z-loss to the MoE output activations. This
regularizer limits activation growth
(Figure~\hyperref[_heading=h.a00uwiya9t7f]{7}) and prevents the
subsequent loss spike (Figure~\hyperref[_heading=h.991z9qs324kj]{8}),
with no observed degradation in benchmark performance and none of the
training overhead introduced by Gated RMSNorm.

\par\addvspace{8pt}\noindent\begin{minipage}{\linewidth}\centering

\protect\phantomsection\label{_heading=h.a00uwiya9t7f}{}

\includegraphics[width=\linewidth,height=.46\textheight,keepaspectratio]{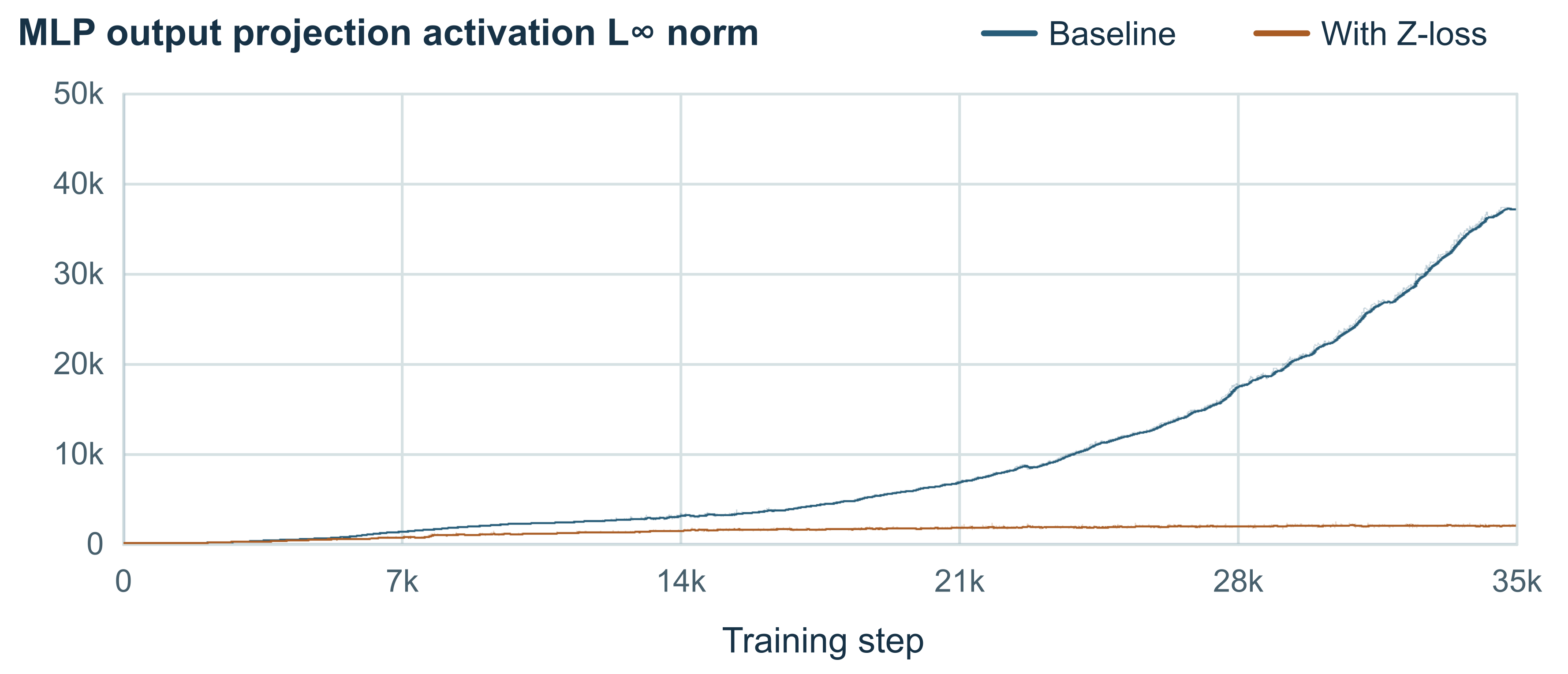}

\par

\captionof{figure}{Norm of MoE down-projection outputs during training,
with and without output Z-loss.}

\end{minipage}\par\addvspace{7pt}

\par\addvspace{8pt}\noindent\begin{minipage}{\linewidth}\centering

\protect\phantomsection\label{_heading=h.991z9qs324kj}{}

\includegraphics[width=\linewidth,height=.46\textheight,keepaspectratio]{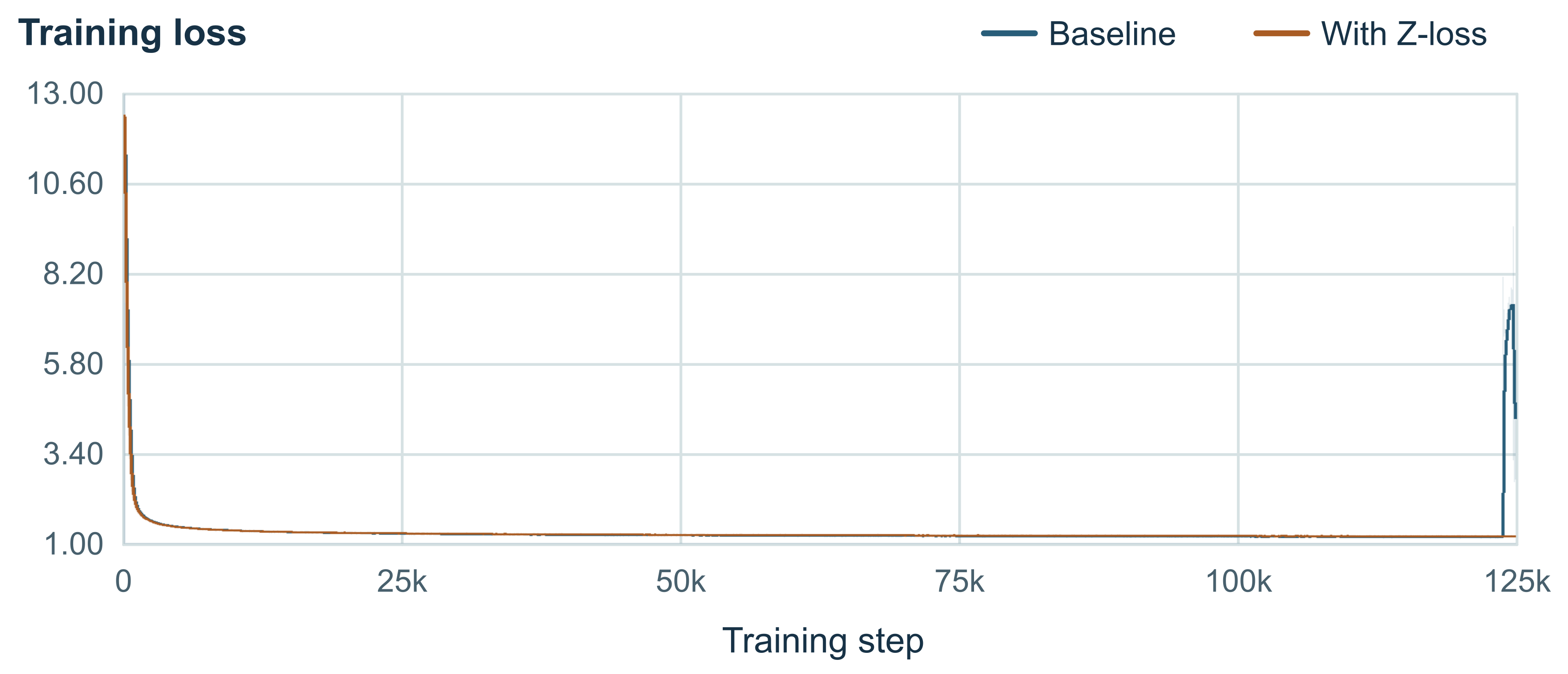}

\par

\captionof{figure}{Training-loss dynamics over a longer horizon. The run
with output Z-loss remains stable.}

\end{minipage}\par\addvspace{7pt}

Let \(x \in \mathbb{R}^{B \times N}\), where \(B\) is the number of tokens and
\(N\) is the activation dimension. For token \(i\), define

\[z_{i} = logsumexp(x_{i}) = log\sum_{j = 1}^{N}exp(x_{ij}).\quad\quad(2)\]

The auxiliary loss is

\[L_{z}(x) = \frac{1}{B}\sum_{i = 1}^{B}z_{i}^{2} = \frac{1}{B}\sum_{i = 1}^{B}\left( \log\sum_{j = 1}^{N}exp(x_{ij}) \right)^{2}.\quad\quad(3)\]

We average over tokens and then over MoE layers, multiply by
\(\lambda_{z}\), and add the result to the main objective:

\[L_{total} = L_{main} + \lambda_{z}L_{z}.\quad\quad(4)\]

Since log-sum-exp is a smooth approximation to the maximum, the loss
responds primarily to large positive activations. Its gradient is

\[\frac{\partial L_{z}}{\partial x_{ij}} = \frac{2}{B}z_{i}softmax(x_{i})_{j}.\quad\quad(5)\]

The softmax factor concentrates the gradient on components with large
values, suppressing the positive outliers that dominate log-sum-exp.
Large negative activations are not directly constrained, since their
contributions are exponentially small. They may nevertheless change
indirectly through shared layer weights and inputs. All operations
associated with this loss are computed in float32 to reduce numerical
error for large activations.

\subsubsection{Muon}\label{muon}

Muon (MomentUm Orthogonalized by Newton--Schulz) is an optimizer for
matrix-valued neural-network parameters. It computes momentum from
gradients and approximately orthogonalizes the momentum matrix using
Newton--Schulz iterations. This matrix-level transformation differs from
AdamW's element-wise updates. Prior work reports comparable performance
at approximately half the training FLOPs of AdamW in a compute-optimal
setting~{[}\hyperref[_heading=h.1zw40gbz19a6]{26}{]}. Orthogonalization,
however, adds computation and communication to each optimizer step.

In a fully sharded data-parallel (FSDP) setting, a parameter matrix may
be distributed across ranks, whereas orthogonalization requires the
complete matrix. Applying Newton--Schulz independently to FSDP shards is
not equivalent to orthogonalizing the full matrix. We first form work
units: MoE tensors are split along the expert axis, while other
two-dimensional weights each form one work unit. We estimate
orthogonalization cost from matrix dimensions and use
longest-processing-time (LPT) scheduling: units are sorted from most to
least expensive and assigned to the least-loaded compute groups. These
groups form stages with balanced work across GPUs, preventing a few
expensive matrices from delaying the entire step. Each matrix is
assigned to an owner GPU. Local shards are packed into communication
buffers and assembled in their original order at the owner. Within a
stage, momentum shards are gathered at their owners, orthogonalized,
split along the original FSDP shard boundaries, and returned to the
original ranks (Figure~\hyperref[_heading=h.ngp4t7apx0bb]{9}).

\par\addvspace{8pt}\noindent\begin{minipage}{\linewidth}\centering

\protect\phantomsection\label{_heading=h.ngp4t7apx0bb}{}

\includegraphics[width=\linewidth,height=.46\textheight,keepaspectratio]{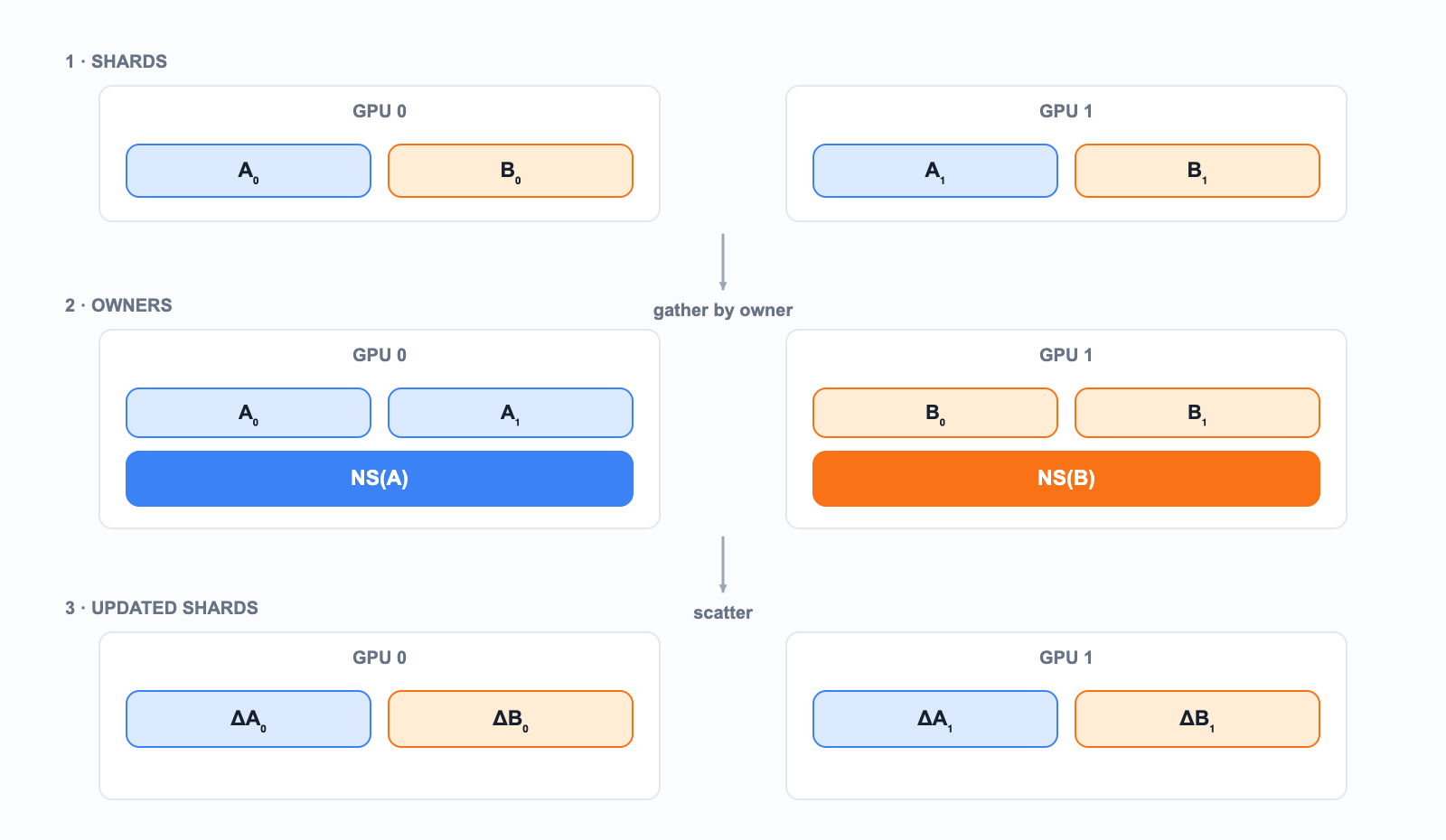}

\par

\captionof{figure}{One optimizer stage: momentum shards are gathered at
owner GPUs, orthogonalized, and returned to their original ranks.}

\end{minipage}\par\addvspace{7pt}

The stages form a pipeline
(Figure~\hyperref[_heading=h.p79skdwg7wdq]{10}). Newton--Schulz
computation for one stage overlaps with gathering the next stage and
returning the previous stage's updates. After pipeline warm-up,
communication and computation largely proceed concurrently. This
implementation approximately halves optimizer-step time in the final
training run.

\par\addvspace{8pt}\noindent\begin{minipage}{\linewidth}\centering

\protect\phantomsection\label{_heading=h.p79skdwg7wdq}{}

\includegraphics[width=\linewidth,height=.46\textheight,keepaspectratio]{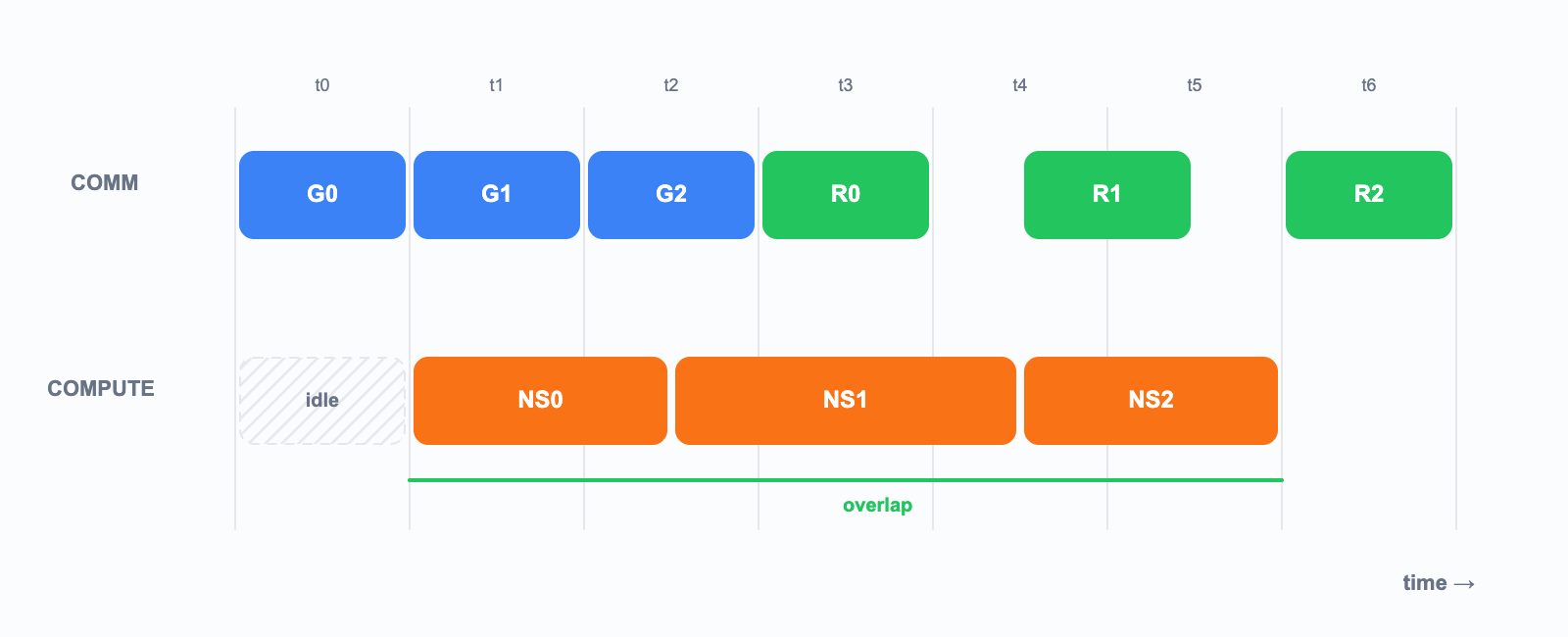}

\par

\captionof{figure}{Communication and orthogonalization overlap in separate
CUDA streams. G denotes momentum gathering, NS Newton--Schulz
orthogonalization, and R returning updates; indices identify stages.}

\end{minipage}\par\addvspace{7pt}

\subsubsection{Kimi Delta Attention}\label{kimi-delta-attention}

Full attention allows each token to access all preceding tokens
directly, but its key--value cache grows with context length. We combine
KDA~{[}\hyperref[_heading=h.dfwgtlqlo5dq]{27}{]} with full attention at
a 3:1 layer ratio. KDA compresses history into a fixed-size state
instead of storing separate keys and values for every token, so its
state memory does not grow with context length.

For each new token, KDA retrieves the value associated with its key,
compares it with the new value, and writes the difference into the
state. It thus incrementally corrects the state rather than replacing
it, while also allowing obsolete information to decay. Unlike the Gated
DeltaNet used in Qwen3-Next, which uses one decay coefficient per
attention head, KDA applies a separate coefficient to each row of the
state matrix. This allows more selective retention and forgetting.

In experiments with 5--15B-parameter models, replacing Gated DeltaNet
with KDA reduced loss by approximately 0.02 and improved mathematics by
about 2--6 percentage points, MMLU-Pro CoT by 5 points, and educational
tasks and information extraction by 2--4 points. We therefore adopted
KDA.

\subsubsection{Attention Residuals}\label{attention-residuals}

Attention Residuals~{[}\hyperref[_heading=h.5fipkt5pphrc]{28}{]}
replaces residual summation with a weighted sum of earlier layer
outputs. Weights are computed per token using a learned pseudo-query and
a softmax over depth. Representations are RMS-normalized before
computing attention weights, preventing their raw norms from determining
those weights.

Our RMSNorm includes a learned feature-wise scale vector \(\gamma\).
This can reintroduce differences in key norms, but additional
parameterization may improve optimization even without increasing
expressivity, as discussed by Wang et al.
{[}\hyperref[_heading=h.k9cvq2epkswd]{29}{]}. Since this scale improved
performance in our experiments, we retained it.

Full AttnRes attends to individual outputs of all earlier layers. We use
Block AttnRes with a block size of eight layers, where an attention
module and an MoE module each count as a layer. Thus one AttnRes block
spans four Transformer blocks. Outputs within a block accumulate into a
partial sum; each subsequent layer attends to this sum and completed
representations from preceding blocks. This reduces the number of
representations read by the kernel and the resulting memory traffic.

We align activation-checkpoint boundaries with AttnRes operations to
avoid redundant saved tensors. Let \(1 \times\) denote the memory
required to retain the input activations of Transformer blocks,
recomputing internal activations during the backward pass. A naive Full
AttnRes implementation retains the input checkpoint plus attention and
MoE outputs, requiring three tensors per Transformer block, or
\(3 \times\) memory (Figure~\hyperref[_heading=h.qt8dj4l4ahl6]{11}).
Moving the checkpoint boundary immediately before the AttnRes operation
following MoE allows the MoE output to serve as the next segment's
checkpoint. The AttnRes result is recomputed, reducing saved activations
to \(2 \times\) (Figure~\hyperref[_heading=h.849cagkfqrzu]{12}).

\par\addvspace{8pt}\noindent\begin{minipage}{\linewidth}\centering

\protect\phantomsection\label{_heading=h.qt8dj4l4ahl6}{}

\includegraphics[width=\linewidth,height=.46\textheight,keepaspectratio]{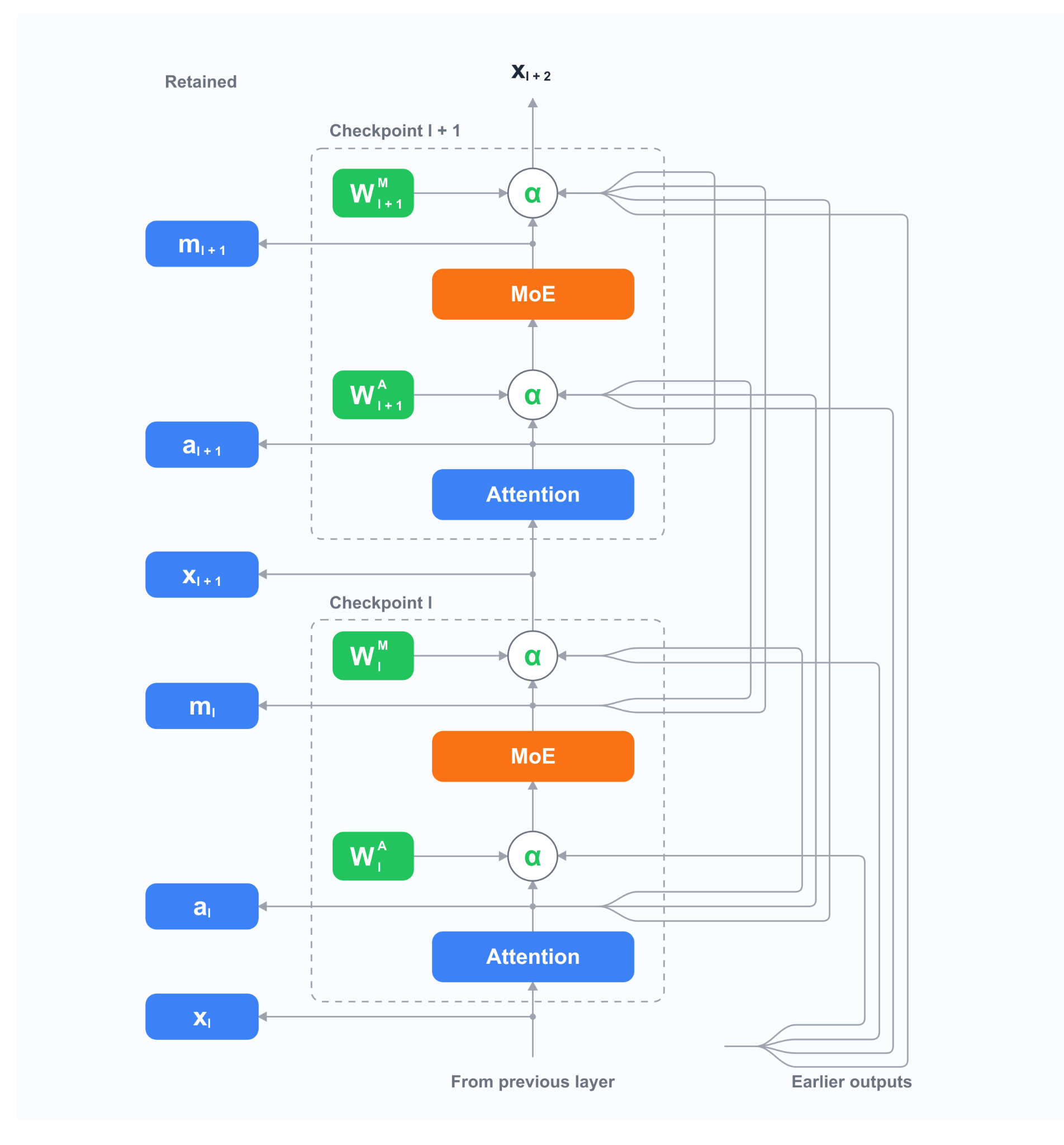}

\par

\captionof{figure}{Naive activation-checkpoint boundaries in Full AttnRes
retain the input checkpoint, attention output, and MoE output.}

\end{minipage}\par\addvspace{7pt}

\par\addvspace{8pt}\noindent\begin{minipage}{\linewidth}\centering

\protect\phantomsection\label{_heading=h.849cagkfqrzu}{}

\includegraphics[width=\linewidth,height=.46\textheight,keepaspectratio]{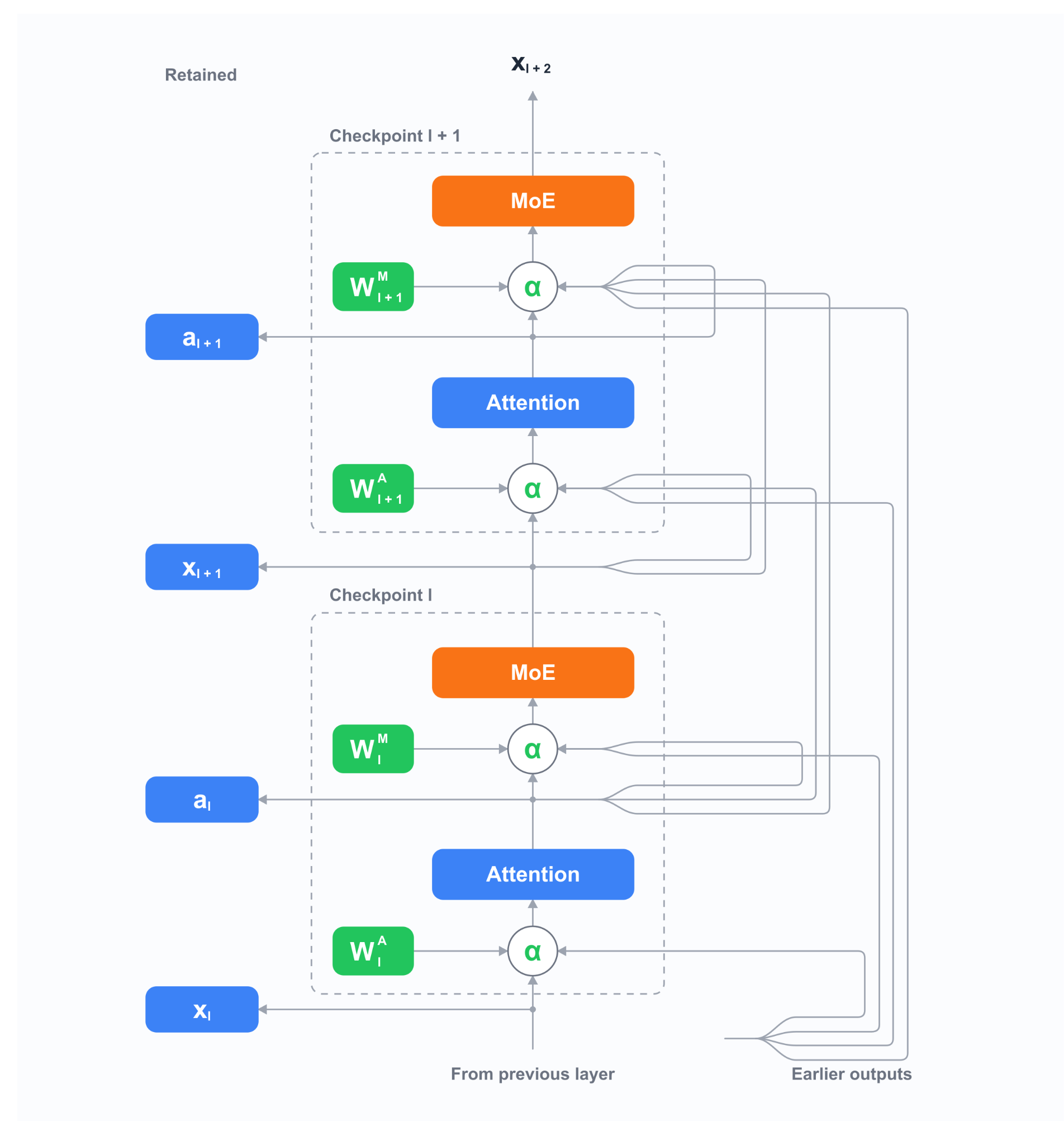}

\par

\captionof{figure}{Shifted activation-checkpoint boundaries let the MoE
output also serve as the next segment's checkpoint.}

\end{minipage}\par\addvspace{7pt}

For Block AttnRes, we similarly save the current partial sum before the
AttnRes operation at Transformer-block boundaries. At the end of an
AttnRes block, that sum also becomes its final representation, so no
additional copy is required. Attention outputs are recomputed, keeping
saved activations at \(1 \times\). Larger AttnRes blocks further reduce
memory traffic but do not further reduce saved activations under this
checkpointing scheme. They also reduce temporary representations during
inference prefill.

On 5--15B models, Block AttnRes reduced loss by approximately 0.017 and
improved Russian and English MMLU-Pro CoT by about 3 points, English
GSM8K CoT by 4.9 points, and several educational benchmarks by 4--5
points. We therefore included it in the final architecture.

\subsection{Scaling Laws for Hyperparameter
Selection}\label{scaling-laws-for-hyperparameter-selection}

\subsubsection{Method}\label{method}

We use small-scale training experiments to estimate the optimal learning
rate \({lr}_{opt}(N,D)\) and batch size \({bs}_{opt}(N,D)\) as functions
of parameter count \(N\) and token budget \(D\). For each fixed pair
\((N,D)\), we fit a loss surface \(loss(lr,bs)\) and find its minimum.
These experiments use models with approximately 5B and 10B total
parameters, trained on 100B--600B tokens.

The fitted scaling law depends strongly on the dataset used to measure
loss, as well as on the chosen functional form. Among the public reports
we reviewed, the Laguna technical
report\footnote{\url{https://poolside.ai/assets/laguna/laguna-m1-xs2-technical-report.pdf}}~{[}\hyperref[_heading=h.i8ubmz4a9qit]{30}{]}
was the only one to provide an explicit formula for the optimal learning
rate. Both Laguna and our model use Muon, but Laguna uses a
warmup--stable--decay (WSD) schedule and we use a linear schedule. For
an approximate comparison, we multiplied Laguna's predicted learning
rate by two. This adjustment should not be interpreted as a general
conversion between the two schedules.

\Needspace{9\baselineskip}\begingroup\small\setlength{\LTpre}{4pt}\setlength{\LTpost}{5pt}

\protect\phantomsection\label{_heading=h.3tqh8wulq307}{}\captionof{table}{Predicted learning rates at a batch size of 33M tokens. The
WSD-to-linear conversion is approximate.}

\nopagebreak[4]

\small\setlength{\tabcolsep}{3pt}

{\def\LTcaptype{reporttablebody} 
\begin{longtable}[]{@{}
  >{\raggedright\arraybackslash}p{(\linewidth - 4\tabcolsep) * \real{0.2700}}
  >{\raggedright\arraybackslash}p{(\linewidth - 4\tabcolsep) * \real{0.2599}}
  >{\raggedright\arraybackslash}p{(\linewidth - 4\tabcolsep) * \real{0.4700}}@{}}
\toprule\noalign{}
\begin{minipage}[b]{\linewidth}\raggedright
\textbf{Model}
\end{minipage} & \begin{minipage}[b]{\linewidth}\raggedright
\textbf{Optimizer and schedule}
\end{minipage} & \begin{minipage}[b]{\linewidth}\raggedright
\textbf{Predicted LR at batch size 33M}
\end{minipage} \\
\midrule\noalign{}
\endhead
\bottomrule\noalign{}
\endlastfoot
Laguna & Muon + WSD & 1.6e-4; approximately 3.2e-4 after conversion to
linear \\
Ours, original validation set & Muon + linear & 5.1e-4 \\
Ours, revised validation sets & Muon + linear & 8.2e-4 \(\pm\) 2.1e-4 \\
\end{longtable}
}

\endgroup

Table~\hyperref[_heading=h.3tqh8wulq307]{8} reports the resulting
predictions. Our predicted optimal learning rate is higher even after
this adjustment. The choice of validation data also substantially
changes the predicted optimum.

\subsubsection{Selecting the Validation
Data}\label{selecting-the-validation-data}

We initially followed the conventional approach of fitting scaling laws
on a held-out subset of the training
corpus~{[}\hyperref[_heading=h.wngo52z3q0z3]{31}{]}. To test the
predicted optimum directly, we ran a 2T-token experiment: if the
predicted batch size is near optimal, halving it and recalculating the
learning rate should not improve downstream benchmarks.

\Needspace{9\baselineskip}\begingroup\small\setlength{\LTpre}{4pt}\setlength{\LTpost}{5pt}

\protect\phantomsection\label{_heading=h.2vq15eq8pmb}{}\captionof{table}{Downstream comparison of the original and halved-batch configurations
after 2T tokens.}

\nopagebreak[4]

\small\setlength{\tabcolsep}{3pt}

{\def\LTcaptype{reporttablebody} 
\begin{longtable}[]{@{}
  >{\raggedright\arraybackslash}p{(\linewidth - 6\tabcolsep) * \real{0.2000}}
  >{\raggedright\arraybackslash}p{(\linewidth - 6\tabcolsep) * \real{0.3000}}
  >{\raggedright\arraybackslash}p{(\linewidth - 6\tabcolsep) * \real{0.3000}}
  >{\raggedright\arraybackslash}p{(\linewidth - 6\tabcolsep) * \real{0.2000}}@{}}
\toprule\noalign{}
\begin{minipage}[b]{\linewidth}\raggedright
\textbf{Metric}
\end{minipage} & \begin{minipage}[b]{\linewidth}\raggedright
\textbf{lr=7.8e-4, bs=16M}
\end{minipage} & \begin{minipage}[b]{\linewidth}\raggedright
\textbf{lr=5.5e-4, bs=8M}
\end{minipage} & \begin{minipage}[b]{\linewidth}\raggedright
\textbf{Difference}
\end{minipage} \\
\midrule\noalign{}
\endhead
\bottomrule\noalign{}
\endlastfoot
Mean score & 63.98\% & 64.44\% & +0.46 pp \\
\end{longtable}
}

\endgroup

The smaller-batch configuration performed better
(Table~\hyperref[_heading=h.2vq15eq8pmb]{9}). At an uncorrected
significance threshold of p \textless{} 0.05, it improved performance on
ten benchmarks, reduced performance on one, and showed no significant
change on the remaining 48 benchmarks. This suggested that the predicted
batch size exceeded the critical value, the learning-rate prediction was
suboptimal, or both.

To determine whether this mismatch arose from the functional form or the
validation data, we compared model rankings by validation loss with
rankings by mean score on a selected benchmark suite. The
ranking-agreement score was only 42\% when loss was measured on held-out
training documents. We then measured loss on benchmark data, relevant
web documents from several domains, and code datasets selected for their
agreement with programming benchmarks.
Table~\hyperref[_heading=h.t2fuvj52nj9e]{10} reports the resulting
ranking-agreement scores.

\Needspace{9\baselineskip}\begingroup\small\setlength{\LTpre}{4pt}\setlength{\LTpost}{5pt}

\protect\phantomsection\label{_heading=h.t2fuvj52nj9e}{}\captionof{table}{Agreement between rankings by loss and by mean benchmark score.}

\nopagebreak[4]

\small\setlength{\tabcolsep}{3pt}

{\def\LTcaptype{reporttablebody} 
\begin{longtable}[]{@{}
  >{\raggedright\arraybackslash}p{(\linewidth - 2\tabcolsep) * \real{0.5000}}
  >{\raggedright\arraybackslash}p{(\linewidth - 2\tabcolsep) * \real{0.5000}}@{}}
\toprule\noalign{}
\begin{minipage}[b]{\linewidth}\raggedright
\textbf{Evaluation dataset}
\end{minipage} & \begin{minipage}[b]{\linewidth}\raggedright
\textbf{Ranking agreement (\%)}
\end{minipage} \\
\midrule\noalign{}
\endhead
\bottomrule\noalign{}
\endlastfoot
Held-out training documents & 42\% \\
Selected code validation set & 94\% \\
Selected benchmark validation set & 93\% \\
Selected web validation set & 87\% \\
\end{longtable}
}

\endgroup

The alternative datasets produced different predictions. Because the
experiments did not unambiguously identify a single best dataset, we
averaged predictions from several of the strongest candidates.
Table~\hyperref[_heading=h.3tqh8wulq307]{8} reports the resulting mean
and standard deviation.

\subsubsection{Stabilizing Training at the Predicted Learning
Rate}\label{stabilizing-training-at-the-predicted-learning-rate}

The baseline run already exhibited rapid growth in MoE activations
followed by a loss spike. We initially suspected that the learning rate
was too high. We reproduced the instability in a model with fewer layers
but the same layer width, which allowed us to test stabilization methods
at substantially lower computational cost. The method described in
Section 4.2.1 enabled stable training of the final model at the learning
rate predicted by our scaling laws.

\section{General Pretraining
Corpus}\label{general-pretraining-corpus-1}

\subsection{Composition}\label{composition-1}

The model is pretrained on a mixture of natural, semi-synthetic, and
fully synthetic data. Natural sources include web pages, publicly
available source code, books, scientific papers, news, multilingual
data, open datasets and domain-specific resources such as legal codes,
clinical guidelines, and medical textbooks. We also include text
extracted from PDF and DjVu documents using optical character
recognition (OCR). Semi-synthetic data include augmented web pages and
generated or corrected solutions to problems drawn from existing
sources. Fully synthetic data consist of tasks and solutions generated
from scratch. The data undergo parsing, deduplication, and evaluation
benchmarks filtering. Figure 13 summarizes the corpus composition.

\par\addvspace{8pt}\noindent\begin{minipage}{\linewidth}\centering

\protect\phantomsection\label{headingh.yxu35u3sq2zg}{}

\includegraphics[width=\linewidth,height=.46\textheight,keepaspectratio]{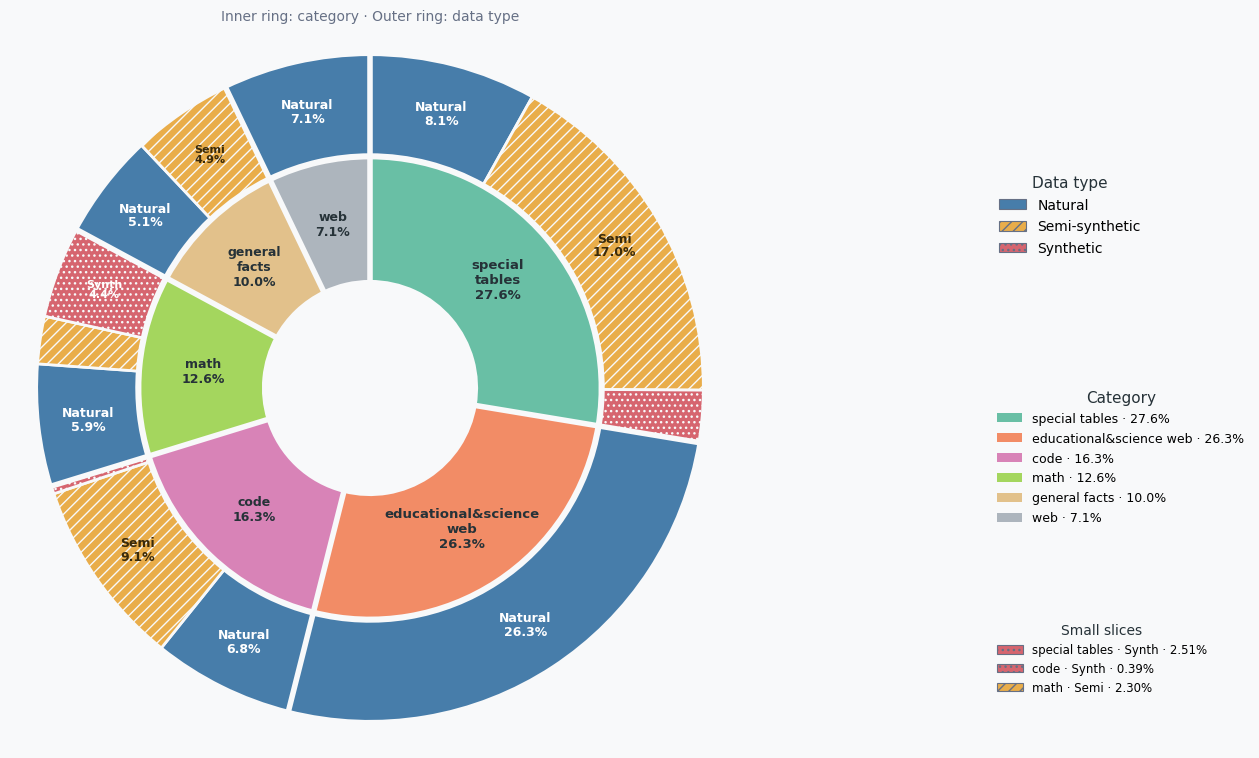}

\par

\captionof{figure}{Natural, semi-synthetic, and fully synthetic data in
the pretraining corpus. The inner ring shows corpus categories and the
outer ring shows data types.}

\end{minipage}\par\addvspace{7pt}

We constructed the baseline mixture from the pretraining corpus used for
YandexGPT 5 Base\footnote{\url{https://huggingface.co/yandex/YandexGPT-5-Lite-8B-pretrain}}.
We expanded and refined it using data collection, generation, and
curation methods developed for our previous Alice AI LLM 235B
Base\footnote{\url{https://habr.com/ru/companies/yandex/articles/974594/}}
{[}1{]}. Alice AI LLM 235B Base was developed through continued
pretraining of Qwen3-235B-A22B, using substantially fewer training
tokens than the from-scratch runs described here. Methods that improved
performance in those shorter runs did not always yield gains when
applied directly to longer runs, as illustrated by factual-data
augmentation in Section 6.4. Adapting data selection, generation, and
augmentation to full-scale pretraining was therefore a central challenge
of this work.

\subsection{Web Data}\label{web-data}

We collect web data using an in-house crawler. From Yandex's crawled
search index, we select Russian- and English-language documents, as well
as the pages returned by Yandex Search in response to user queries. An
internal parser converts page content to Markdown. We then filter
documents by their quality using a cascade of models: a lightweight DSSM
classifier running on CPUs, a 0.5B encoder--decoder model, and, for
selected subsets, an 8B model based on YandexGPT 5 Lite Base.

We construct separate subsets for mathematics, programming, educational
and scientific content, and factual knowledge. A classifier trained on
human-labeled examples selects educational and scientific content,
yielding several trillion tokens. For factual knowledge, we use a
separate cascade of neural classifiers to select documents for
subsequent augmentation. Section 6 describes the data selection and
augmentation pipelines for factual knowledge, mathematics, and
programming.

\subsection{Long-Context Data}\label{long-context-data}

As described in Section 3, we extend the model's context window to 256K
tokens during context extension and use a 128K-token context for the
final reasoning stage. Natural documents approaching these lengths are
scarce, so we rely heavily on synthetic long-context data. Alongside
publicly available books, we include questions that require gathering
and combining information across an entire book, QA over spreadsheets
and other long structured inputs, and QA over collections of medical or
legal documents with each containing closely related terms. We also
include multi-hop search tasks with retrieved documents provided as
context.

Using only document-grounded QA in the synthetic long-context data led
the model to rely too heavily on the supplied documents. This poses a
risk in applications that use search, where retrieved documents often
contain substantial amounts of irrelevant information.

To measure this effect, we constructed a diagnostic benchmark by adding
distracting documents to factual questions. These documents did not
contain the answers. We then compared performance on the same questions
with and without the additional context (Figure 14). Ideally, adding
documents that provide no relevant evidence should not reduce accuracy.

\par\addvspace{8pt}\noindent\begin{minipage}{\linewidth}\centering

\includegraphics[width=\linewidth,height=.46\textheight,keepaspectratio]{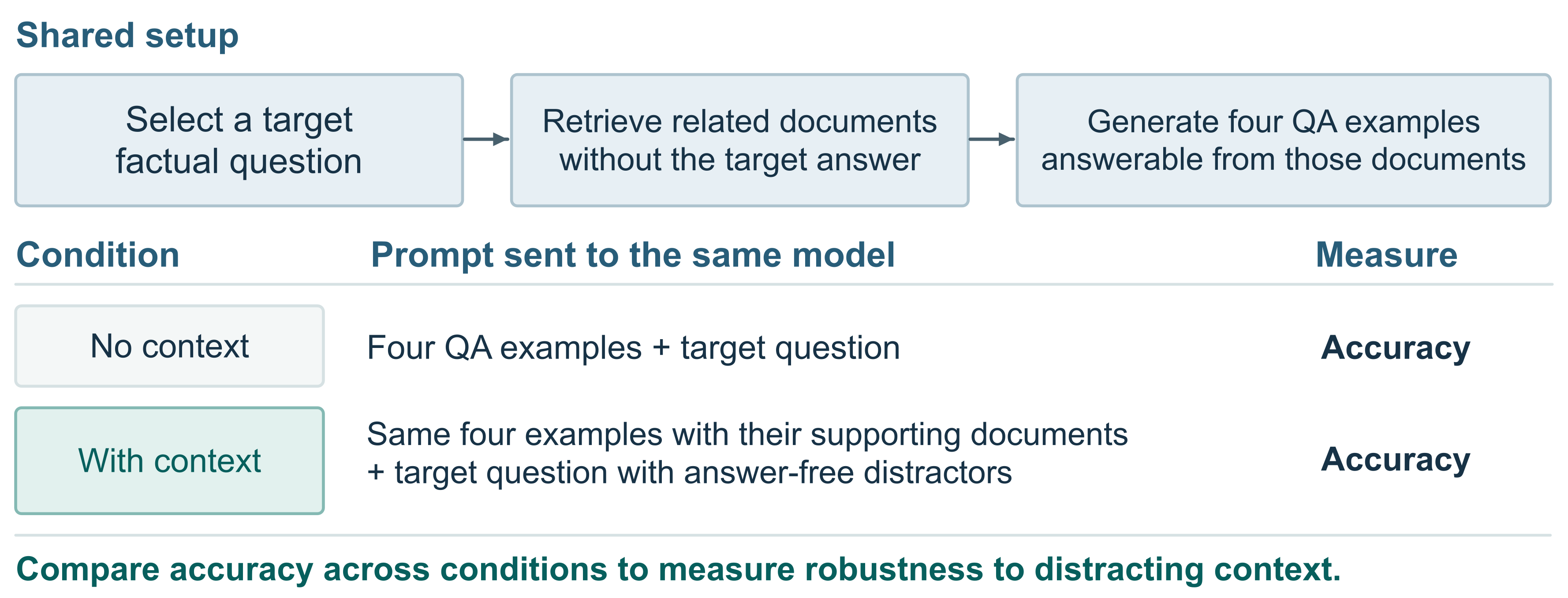}

\par

\captionof{figure}{Effect of distracting context on factual accuracy. Both
conditions use the same four question--answer demonstrations, generated
with access to retrieved documents. Distracting documents supplied for
the test question do not contain its answer. This diagnostic uses four
demonstrations, whereas the main factual evaluations use five.}

\end{minipage}\par\addvspace{7pt}

Adding synthetic long-context data improved scores on standard
long-context benchmarks. However, on our diagnostic benchmark,
introducing irrelevant documents reduced scores by 30--40\% relative to
evaluation without context. Training the model on QA examples with
irrelevant context reduced this relative drop to 10\%.

\subsection{Mathematical Data}\label{mathematical-data}

Our mathematics corpus combines selected mathematical web documents,
augmented versions of these documents, and synthetic problem solutions.
We compared our data with public datasets, including UltraData-Math and
Nemotron-CC-Math-v1 {[}32, 33{]}. Both mathematical web data and their
augmented versions improved scores on mathematics benchmarks in our
pretraining experiments. Synthetic problem solutions also improved
performance. Section 6.8 describes our methods for selecting and
augmenting mathematical web data. We generate synthetic solutions using
an approach similar to that used for YandexGPT 5 Lite Base, described in
our Russian-language technical blog post\footnote{\url{https://habr.com/ru/companies/yandex/articles/885218/}}.

\subsection{Programming Data}\label{programming-data}

In addition to programming web data (Section 6.9), the pretraining
corpus includes code from GitHub repositories selected using an
LLM-based quality filter. These include widely used libraries and other
prominent open-source projects. The corpus also contains solutions to
synthetic programming problems, as described in Section 7.2.2.

\section{Domain-Specific Data
Construction}\label{domain-specific-data-construction}

We initially developed our data selection and augmentation methods for
the shorter continued-pretraining runs of Alice AI LLM 235B Base and
outlined them in a Russian-language blog post\footnote{\url{https://habr.com/ru/companies/yandex/articles/974594/}}
{[}1{]}. Here we describe the full pipeline in detail and explain how we
adapted it for the much longer pretraining runs of
AliceAI-Foundation-80B-A3B-Base.

We first developed our data selection and augmentation methods for
factual data. Section 6.1 explains the motivation for this choice.
Sections 6.2 and 6.3 describe how we identify gaps in the corpus's
factual coverage and select relevant web documents. Section 6.4 presents
the augmentation principles developed for Alice AI LLM 235B Base and
explains how we revised the pipeline when the original approach did not
deliver the expected gains at a larger training scale. Section 6.5
examines the effects of adding large amounts of synthetic QA data to
pretraining and the regularizing effect of data augmentation. We then
describe how we adapt the data selection and augmentation methods
developed for factual knowledge to specialized domains, mathematics,
programming, and STEM.

Our augmentation methods share a central principle with related work,
including the Kimi K2 technical report\footnote{\url{https://arxiv.org/html/2507.20534v1\#S2.SS2}}
{[}34{]}: presenting the same knowledge in diverse forms while
preserving factual content, rather than repeatedly training on identical
text.

\subsection{Factual Knowledge
Acquisition}\label{factual-knowledge-acquisition}

Models acquire much of their factual knowledge during pretraining, and
filling the remaining gaps later can be difficult under a limited
compute budget. Gekhman et al. {[}35{]} find that models learn
unfamiliar facts more slowly than familiar ones during SFT and that
intensive training on new facts can increase the hallucination
frequency. We therefore treat factual knowledge as a core capability of
the base model, needed for both everyday questions and specialized
tasks. With a fixed training token budget, we aim to improve both the
coverage of relevant facts in the corpus and the model's ability to
learn them.

\subsection{Exploring corpus coverage
gaps}\label{exploring-corpus-coverage-gaps}

Our factual benchmarks focus on facts that do not change over time.
Although search tool is available in the deployed system, we evaluate
short factual questions without search to measure the model's parametric
knowledge. We rely on search for rapidly changing information.

Search can supply facts that the model has not learned. More complex
user requests, however, often require combining multiple facts. When
this involves several retrieval steps, latency increases and irrelevant
information may accumulate, making it harder to produce a complete,
coherent answer. Domain knowledge also helps the model formulate
effective search queries. We aim to train a model that draws on broad
factual knowledge and use search to retrieve rare facts and recent
information as needed (Figure 15).

\par\addvspace{8pt}\noindent\begin{minipage}{\linewidth}\centering

\includegraphics[width=\linewidth,height=.46\textheight,keepaspectratio]{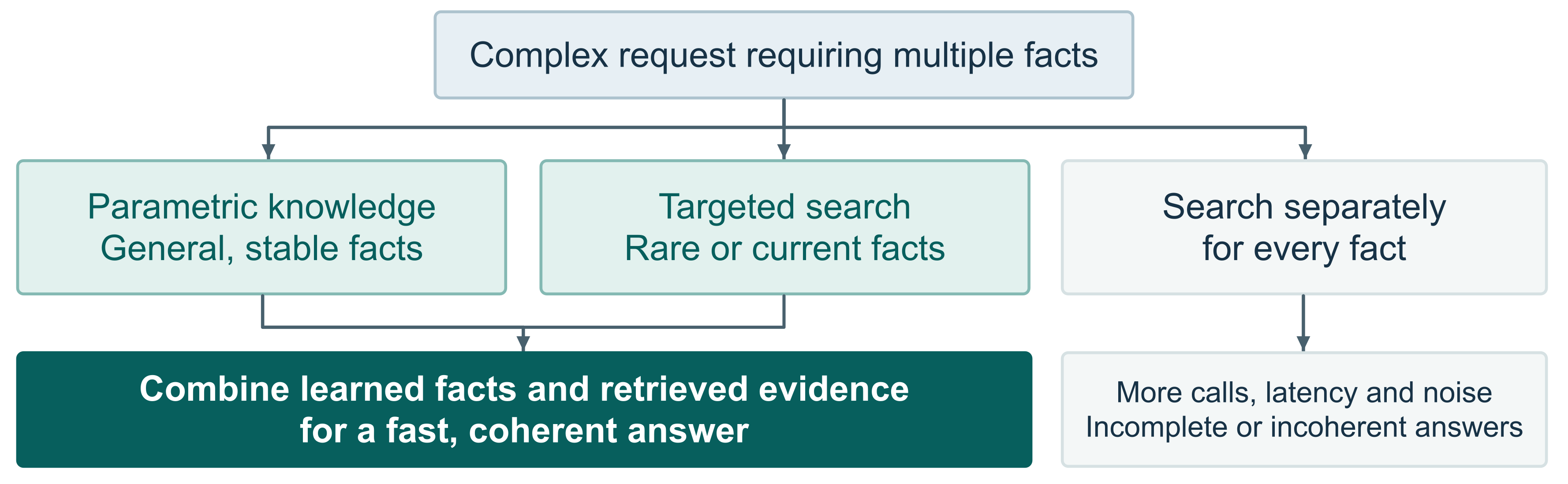}

\par

\captionof{figure}{Complementary roles of parametric knowledge and search
in answering complex questions.}

\end{minipage}\par\addvspace{7pt}

We compared a smaller model with access to search against a larger model
without search on the same factual benchmark (Figure 16). The smaller
model with search substantially outperformed the larger model on short
factual questions. Correct answers supported by retrieved documents
indicated that the required information was available online. We then
investigated whether the relevant documents were missing from the
training corpus or whether the model had failed to learn the facts they
contained.

\par\addvspace{8pt}\noindent\begin{minipage}{\linewidth}\centering

\includegraphics[width=\linewidth,height=.46\textheight,keepaspectratio]{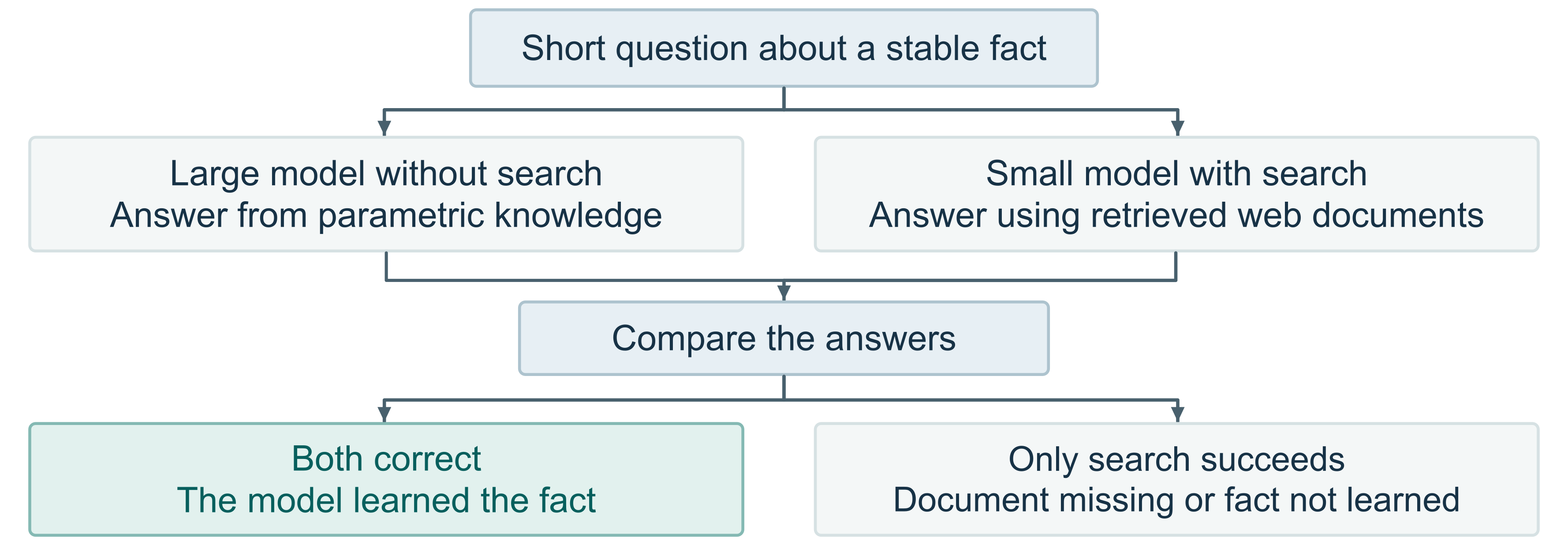}

\par

\captionof{figure}{Diagnosing gaps in corpus coverage and knowledge
acquisition by comparing answers generated with and without search.}

\end{minipage}\par\addvspace{7pt}

We retrieved useful pages for benchmark questions, verified through
training experiments that adding these pages improved factual scores,
and checked whether they were present in the training corpus. These
pages were used only in diagnostic experiments and were not subsequently
reused for corpus construction, to avoid selecting training data
specifically for the benchmark. More than 60\% of these pages were
missing from the final corpus: some had been excluded during initial
source selection, while others had been discarded by subsequent filters
(Figure 17). These findings motivated treating factual web content as a
separate component of the training corpus.

\par\addvspace{8pt}\noindent\begin{minipage}{\linewidth}\centering

\includegraphics[width=\linewidth,height=.46\textheight,keepaspectratio]{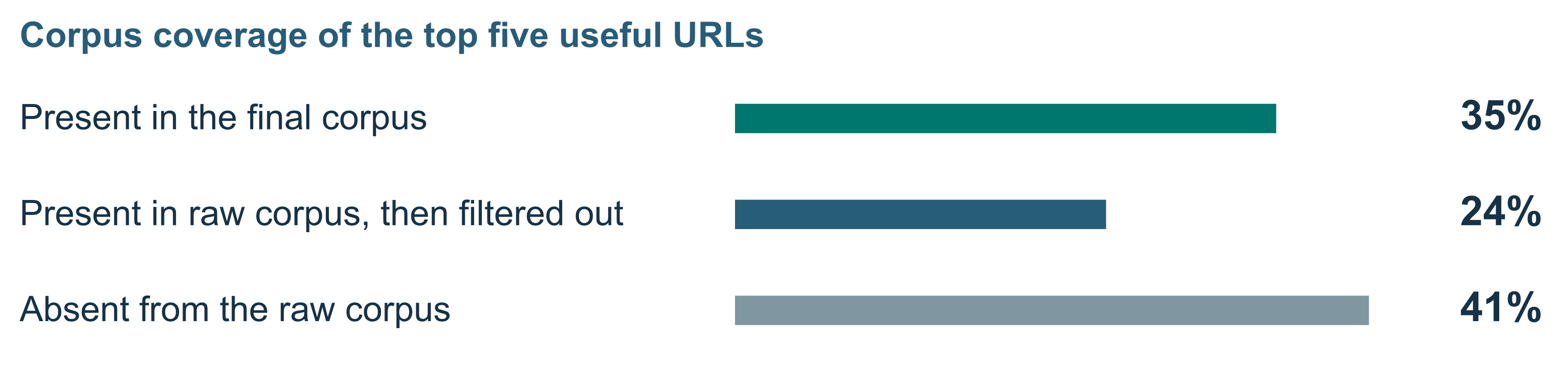}

\par

\captionof{figure}{Coverage of useful factual documents among the top five
retrieved URLs: 35\% survive final selection, 24\% occur in the raw
corpus but are discarded, and 41\% are absent from the raw corpus.}

\end{minipage}\par\addvspace{7pt}

\subsection{Factual-Corpus Selection
Pipeline}\label{factual-corpus-selection-pipeline}

We reviewed and improved every stage of the data pipeline, substantially
increasing the number of URLs processed and doubling the size of the raw
pretraining corpus. This required substantially more CPU capacity and
storage. We then focused on improving the recall of the downstream
filters used to select factual documents.

Earlier pipeline versions relied on lightweight DSSM classifiers. We
used a large LLM to label several hundred thousand documents and trained
0.5B and 8B classifiers on these labels. The two classifiers achieved
similar classification scores but differed substantially in how they
ranked documents, particularly those assigned scores above 0.8 by the 8B
model. Adding these documents to the training corpus produced the most
consistent benchmark gains, indicating that the 0.5B classifier had
rejected valuable training data.

Applying the 8B classifier to the entire corpus would have required more
than 200,000 GPU-hours. We therefore designed a filtering cascade that
prioritizes high recall in its early stages (Figure 18). The cascade
retains approximately 95\% of relevant documents while reducing
computational cost by more than a factor of ten compared with applying
the 8B classifier to the entire corpus.

\par\addvspace{8pt}\noindent\begin{minipage}{\linewidth}\centering

\includegraphics[width=\linewidth,height=.46\textheight,keepaspectratio]{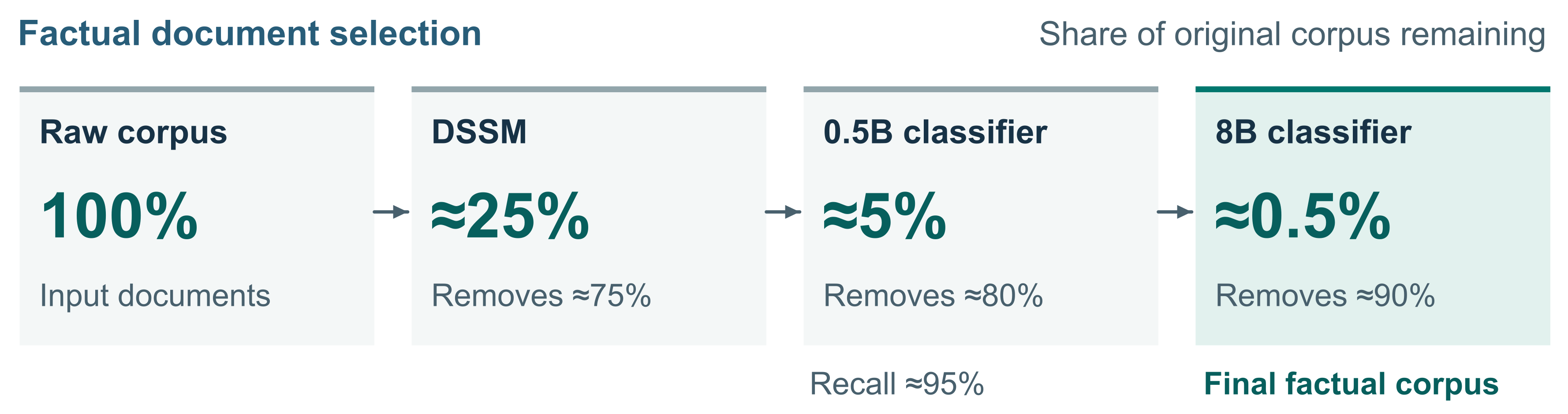}

\par

\captionof{figure}{Filtering cascade for factual web documents.
Percentages indicate the fraction of the original corpus remaining after
each stage. The 8B classifier processes only about 5\% of the corpus.}

\end{minipage}\par\addvspace{7pt}

\subsection{Factual Data
Augmentation}\label{factual-data-augmentation}

\protect\phantomsection\label{_heading=h.3uwidjzbiq14}{}Motivated by the
Physics of Language Models study {[}36{]}, we investigated whether data
augmentation could improve the model's acquisition of factual knowledge.
We selected encyclopedia pages known to contain answers to an internal
benchmark. Using short continued-pretraining runs, we compared training
on augmented versions of these pages with repeated training on the
original documents. Keeping the source documents fixed allowed us to
study knowledge acquisition separately from source selection.

We investigated model-generated and rule-based augmentations, including
the examples in Table 11. Several methods improved benchmark scores in
small-scale experiments, but these gains were no longer statistically
significant when the methods were combined, evaluated in the target
training configuration, or applied at a larger scale. We retained
contextual rewrites, stylistic reformulations, and QA pairs. Table 12
summarizes the small-scale experiments. Rewrites with hard distractors
were introduced later and were not included in that comparison.

\Needspace{9\baselineskip}\begingroup\small\setlength{\LTpre}{4pt}\setlength{\LTpost}{5pt}

\captionof{table}{Illustrative transformations of a factual passage.}

\nopagebreak[4]

\small\setlength{\tabcolsep}{3pt}

{\def\LTcaptype{reporttablebody} 
\begin{longtable}[]{@{}
  >{\raggedright\arraybackslash}p{(\linewidth - 2\tabcolsep) * \real{0.2100}}
  >{\raggedright\arraybackslash}p{(\linewidth - 2\tabcolsep) * \real{0.7900}}@{}}
\toprule\noalign{}
\begin{minipage}[b]{\linewidth}\raggedright
\textbf{Transformation}
\end{minipage} & \begin{minipage}[b]{\linewidth}\raggedright
\textbf{Example}
\end{minipage} \\
\midrule\noalign{}
\endhead
\bottomrule\noalign{}
\endlastfoot
Original & Peter I founded Saint Petersburg in 1703. It remained the
capital of Russia until 1918. \\
Contextual rewrite & As the riverboat passed the Peter and Paul
Fortress, the guide told the tourists: Our city began with this
fortress, founded by Peter I in 1703. Until 1918, it was the heart and
capital of the country. \\
Chronicle-style reformulation & In the year 1703 of Our Lord, Tsar Peter
commanded that a new city be built. Saint Petersburg became the capital
of the Russian land and remained so until the troubled year of 1918. \\
QA pairs & Question: In which century was Saint Petersburg founded?
Answer: In the eighteenth century, in 1703. \\
Structural change & Saint Petersburg was founded by Peter I in 1703. It
remained the capital of Russia until 1918. \\
Entity aliasing & Saint Petersburg is a gray city. The gray city was
founded by Peter I in 1703. It remained the capital of Russia until
1918. \\
Rewrite with difficult distractors & The founding of Saint Petersburg is
sometimes confused with other events of the Petrine era: 1698, Peter I's
return from the Grand Embassy; 1700, the start of the Great Northern
War; 1712, the transfer of the capital to Petersburg; and 1721, the
proclamation of the Russian Empire. The city itself was founded by Peter
I in 1703 and remained the capital until 1918. \\
\end{longtable}
}

\endgroup

\Needspace{9\baselineskip}\begingroup\small\setlength{\LTpre}{4pt}\setlength{\LTpost}{5pt}

\captionof{table}{Factual-augmentation experiments on WIKIBENCH, an early
version of WikiWebFacts based exclusively on Wikipedia. The first row
reports the baseline score; subsequent rows report changes relative to
this baseline.}

\nopagebreak[4]

\small\setlength{\tabcolsep}{3pt}

{\def\LTcaptype{reporttablebody} 
\begin{longtable}[]{@{}
  >{\raggedright\arraybackslash}p{(\linewidth - 2\tabcolsep) * \real{0.5000}}
  >{\raggedright\arraybackslash}p{(\linewidth - 2\tabcolsep) * \real{0.5000}}@{}}
\toprule\noalign{}
\begin{minipage}[b]{\linewidth}\raggedright
\textbf{Training mixture}
\end{minipage} & \begin{minipage}[b]{\linewidth}\raggedright
\textbf{WIKIBENCH score (\%) / change (pp)}
\end{minipage} \\
\midrule\noalign{}
\endhead
\bottomrule\noalign{}
\endlastfoot
Repeated texts & 59.5 \\
QA pairs & +0.1\% \\
Rewrites & +2.3\% \\
Repeated texts + questions & +3.2\% \\
Repeated texts + rewrites & +4.6\% \\
Repeated texts + questions + rewrites & +6.9\% \\
\end{longtable}
}

\endgroup

Initially, augmented data accounted for approximately 5--10\% of
training tokens used for continued pretraining of Alice AI LLM 235B. To
limit costs, we augmented the most useful sources, particularly
Russian-language online encyclopedias and the factual knowledge base
built for search. At this stage, we had not extended augmentation to the
full collection of factual web documents. In a short training
experiment, augmented data improved the HardMultiQA score by
approximately 6 percentage points relative to training on the YandexGPT
5 Lite Base corpus.

We incorporated the factual web data and their augmentations into a
substantially larger training corpus containing many additional
documents. However, the longer pretraining runs did not yield the
expected gains on factual-knowledge benchmarks and sometimes scored
lower than shorter continued-pretraining runs. We attributed this to the
lower relative frequency of individual useful facts in the expanded
training mixture. This is consistent with evidence that whether a model
learns a fact depends not only on its presence in the corpus but also on
how often the model encounters it during training {[}37{]}. Simply
increasing the number of epochs did not resolve the issue in either that
work or our experiments.

We therefore extended augmentation to the broader collection of factual
web documents. Due to cost constraints, we generated augmentations for
approximately 30\% of this corpus. Applying augmentation to this broader
collection improved the HardMultiQA score by approximately 3 percentage
points in a short training run. In a longer training run combining all
corpus improvements, the HardMultiQA score increased by approximately 5
percentage points. Data augmentation and improved recall in corpus
collection and filtering made the largest contributions to the final
model's gains on factual questions.

\subsection{Effects of Synthetic QA on Model
Behavior}\label{effects-of-synthetic-qa-on-model-behavior}

Synthetic question--answer (QA) pairs are widely used in language model
training. Generative search systems provide a natural source of such
data: specialized models produce answers from retrieved context and are
trained to cite supporting sources. Reusing these query--answer pairs
offers a practical way to expand the training corpus. We therefore
investigated the effects of adding search-generated QA data to
pretraining.

Despite the use of specialized models, we observed two recurring issues:
unresolved ambiguity in question--answer pairs and repeated use of
standardized answer formats (Table 13). Irrelevant retrieved context or
ambiguous queries can lead to answers that address an unintended
interpretation of the question. Repeated answer templates can also
introduce patterns that affect subsequent model behavior.

\Needspace{9\baselineskip}\begingroup\small\setlength{\LTpre}{4pt}\setlength{\LTpost}{5pt}

\captionof{table}{Examples of ambiguity and answer-format bias in
synthetic QA. The ambiguous Russian query is shown verbatim; the
remaining queries and answers are translated. Ages reproduce the
recorded experimental outputs.}

\nopagebreak[4]

\small\setlength{\tabcolsep}{3pt}

{\def\LTcaptype{reporttablebody} 
\begin{longtable}[]{@{}
  >{\raggedright\arraybackslash}p{(\linewidth - 6\tabcolsep) * \real{0.1624}}
  >{\raggedright\arraybackslash}p{(\linewidth - 6\tabcolsep) * \real{0.3232}}
  >{\raggedright\arraybackslash}p{(\linewidth - 6\tabcolsep) * \real{0.2605}}
  >{\raggedright\arraybackslash}p{(\linewidth - 6\tabcolsep) * \real{0.2315}}@{}}
\toprule\noalign{}
\begin{minipage}[b]{\linewidth}\raggedright
Issue
\end{minipage} & \begin{minipage}[b]{\linewidth}\raggedright
Training example
\end{minipage} & \begin{minipage}[b]{\linewidth}\raggedright
Problem
\end{minipage} & \begin{minipage}[b]{\linewidth}\raggedright
Potential effect on model behavior
\end{minipage} \\
\midrule\noalign{}
\endhead
\bottomrule\noalign{}
\endlastfoot
Semantic ambiguity & Query: «{\fontencoding{T2A}\selectfont Сколько вороны несут яйца}?» --- an
ambiguous Russian question about crow egg-laying. Answer: A female
hooded crow lays 4--6 eggs between late March and May. & The query can
be interpreted as asking about clutch size, incubation duration, or
breeding season. The answer selects an interpretation without resolving
the ambiguity. & The model may return information about clutch size even
when the question asks about incubation duration. \\
Repeated age-answer format for living people & Query: How old is the
hockey player Ovechkin? Answer: Alexander Ovechkin is 40 (born September
17, 1985). & A birth date is systematically included even though it was
not requested. & The model may treat a birth date as a required part of
an answer about age. \\
Repeated age-answer format for deceased people & Query: How old was
Tarkovsky? Answer: Andrei Tarkovsky was 54 (born April 4, 1932; died
December 29, 1986). & Answers repeatedly include both birth and death
dates, although the question asks only for age. & The model may return a
birth year or lifespan instead of an age. \\
\end{longtable}
}

\endgroup

\subsubsection{Ambiguity in Question--Answer
Pairs}\label{ambiguity-in-questionanswer-pairs}

After adding QA generated from retrieved documents to pretraining, we
evaluated the model with zero, one, and two demonstrations. In the
example shown in Table 14, the model selected unintended interpretations
of a Russian question about crow reproduction.

The test query, «{\fontencoding{T2A}\selectfont Сколько вороны несут яйца по времени}?» (literally,
``How long do crows lay eggs for?''), was intended to ask about
incubation duration, but its wording shared terms with the ambiguous
training example. Without demonstrations, the model returned clutch
size. With one demonstration, it answered in terms of breeding season;
with two, it returned incubation duration.

\Needspace{9\baselineskip}\begingroup\small\setlength{\LTpre}{4pt}\setlength{\LTpost}{5pt}

\captionof{table}{Effect of demonstrations on an ambiguous Russian
question about crow reproduction. The test query is shown in Russian;
demonstration descriptions and model answers are provided in English.}

\nopagebreak[4]

\small\setlength{\tabcolsep}{3pt}

{\def\LTcaptype{reporttablebody} 
\begin{longtable}[]{@{}
  >{\raggedright\arraybackslash}p{(\linewidth - 6\tabcolsep) * \real{0.1029}}
  >{\raggedright\arraybackslash}p{(\linewidth - 6\tabcolsep) * \real{0.3505}}
  >{\raggedright\arraybackslash}p{(\linewidth - 6\tabcolsep) * \real{0.2412}}
  >{\raggedright\arraybackslash}p{(\linewidth - 6\tabcolsep) * \real{0.2846}}@{}}
\toprule\noalign{}
\begin{minipage}[b]{\linewidth}\raggedright
Setting
\end{minipage} & \begin{minipage}[b]{\linewidth}\raggedright
Query and demonstrations
\end{minipage} & \begin{minipage}[b]{\linewidth}\raggedright
Model answer
\end{minipage} & \begin{minipage}[b]{\linewidth}\raggedright
Observation
\end{minipage} \\
\midrule\noalign{}
\endhead
\bottomrule\noalign{}
\endlastfoot
Zero-shot & «{\fontencoding{T2A}\selectfont Сколько вороны несут яйца по времени}?» & A female crow lays
4--6 eggs. & The model returns clutch size instead of incubation
duration. \\
1-shot & The same query, preceded by: In what year was Pushkin born? ---
1799. & From late March to early May. & The model returns the breeding
season rather than incubation duration. \\
2-shot & The same query, preceded by two short QA demonstrations. & The
incubation period lasts 18--19 days. & The model answers according to
the intended interpretation. \\
\end{longtable}
}

\endgroup

These examples show that demonstrations can steer the model toward
different interpretations of the same question. Although demonstrations
helped elicit the intended answer, this sensitivity to demonstration
content is undesirable.

\subsubsection{Bias from Repeated Answer
Formats}\label{bias-from-repeated-answer-formats}

A related effect appeared in questions about people's ages. Synthetic
answers repeatedly included birth dates for living people and both birth
and death dates for deceased people. The demonstrations then influenced
not only the answer format but also which information the model
returned.

In the examples in Table 15, demonstrations containing birth years or
lifespans prompted the model to return dates rather than ages for
deceased people. For living people, the model continued to return ages.
A single demonstration could therefore change the type of information
returned, even though the question still asked for age.

\Needspace{9\baselineskip}\begingroup\small\setlength{\LTpre}{4pt}\setlength{\LTpost}{5pt}

\captionof{table}{Effects of demonstration format on answers to age
questions. All values reproduce recorded model outputs from the
experiment.}

\nopagebreak[4]

\small\setlength{\tabcolsep}{3pt}

{\def\LTcaptype{reporttablebody} 
\begin{longtable}[]{@{}
  >{\raggedright\arraybackslash}p{(\linewidth - 6\tabcolsep) * \real{0.2781}}
  >{\raggedright\arraybackslash}p{(\linewidth - 6\tabcolsep) * \real{0.2363}}
  >{\raggedright\arraybackslash}p{(\linewidth - 6\tabcolsep) * \real{0.2154}}
  >{\raggedright\arraybackslash}p{(\linewidth - 6\tabcolsep) * \real{0.2492}}@{}}
\toprule\noalign{}
\begin{minipage}[b]{\linewidth}\raggedright
Condition
\end{minipage} & \begin{minipage}[b]{\linewidth}\raggedright
Deceased people
\end{minipage} & \begin{minipage}[b]{\linewidth}\raggedright
Living people
\end{minipage} & \begin{minipage}[b]{\linewidth}\raggedright
Interpretation
\end{minipage} \\
\midrule\noalign{}
\endhead
\bottomrule\noalign{}
\endlastfoot
Typical training template & Tarkovsky: 54 years (1932--1986) & Ovechkin:
40 years (born in 1985) & Answers systematically include dates that were
not requested. \\
Zero-shot: How old is/was \{surname\}? & Lermontov: 26 years
(1814--1841) & Pelevin: 63 years (born in 1962) & The model reproduces
the training answer format without demonstrations. \\
Demonstration gives a birth year: Pushkin was born in 1799. & Lermontov:
1814; Turgenev: 1818; Nekrasov: 1821 & Pelevin: 63 years; King: 78
years; Dontsova: 74 years & The model returns birth years for deceased
people and ages for living people. \\
Demonstration gives a lifespan: Pushkin, 1799--1837. & Lermontov:
1814--1841; Turgenev: 1818--1883; Nekrasov: 1821--1878 & Pelevin: 63
years; King: 78 years; Dontsova: 74 years & The model returns lifespans
for deceased people and ages for living people. \\
Control demonstrations without dates & Lermontov: 26 years; Turgenev: 64
years; Nekrasov: 56 years & Pelevin: 63 years; King: 78 years; Dontsova:
74 years & Without date cues, the model returns ages for both groups. \\
\end{longtable}
}

\endgroup

\subsubsection{Regularization through Data
Augmentation}\label{regularization-through-data-augmentation}

We applied stylistic reformulations and contextual rewrites to the QA
training data and repeated the evaluations after training with the
augmented data. On the examples tested, we no longer observed the
semantic and answer-format effects described above.

Greater diversity in wording and context reduced the model's sensitivity
to individual keywords and repeated answer templates. These results
suggest that data augmentation can act as a regularizer, reducing
sensitivity to incidental wording and recurring answer formats.

\subsection{Generalizing the Data Curation
Pipeline}\label{generalizing-the-data-curation-pipeline}

Our approach aims to improve both the coverage of relevant information
in the training corpus and the model's ability to learn that
information. We first assess corpus coverage using a representative
diagnostic benchmark which we keep separate from document-classifier
development. We then improve document selection and evaluate the
usefulness of the selected data through controlled training experiments.

To keep document selection computationally efficient, we use a cascade
with inexpensive, high-recall filters in the early stages and more
expensive models were applied only to the retained documents. To study
knowledge acquisition separately from source selection, we use documents
known to contain the target facts. We compare domain-specific
augmentations and their combinations in small-scale training
experiments, then apply the selected strategy at scale. The following
sections describe how we adapt this approach, originally developed for
factual data, to domain-specific knowledge in law and medicine, as well
as to mathematics, programming, and STEM.

\subsection{Domain-Specific Knowledge: Law and
Medicine}\label{domain-specific-knowledge-law-and-medicine}

\subsubsection{Legal Data}\label{legal-data}

We first categorized legal benchmark questions and training documents by
area of law. Legal content in the web corpus was skewed toward criminal
and international law, while Alice users more often asked practical
legal questions, including questions about administrative law. When
prompted with article numbers, the base model correctly reproduced the
corresponding Russian statutory provisions in only 6\% of cases.

We expanded the corpus using web documents classified by Alice AI 235B
model and 8B classifiers. Documents were classified by legal relevance,
source type, area of law, jurisdiction, position in the hierarchy of
legal acts, usefulness, and text quality. We also used a link-based
pipeline to collect Russian legal codes. Candidate pages were retrieved
by heuristics, and a model identified individual articles. We
deduplicated the extracted articles and also identified provisions that
had been repealed. We then generated rewrites of the statutory text.

We evaluated each data type in controlled training experiments.
Synthetic QA and primary legal sources collected through this link-based
pipeline produced the most consistent improvements. Educational texts
and dialogues improved some evaluation scores, but their effects
depended on the training mixture and sampling weights. Script-generated
rewrites did not yield consistent improvements and were excluded. Table
16 gives translated examples of the retained augmentations.

\Needspace{9\baselineskip}\begingroup\small\setlength{\LTpre}{4pt}\setlength{\LTpost}{5pt}

\captionof{table}{Examples of synthetic QA and textbook-style augmentation
of legal documents.}

\nopagebreak[4]

\small\setlength{\tabcolsep}{3pt}

{\def\LTcaptype{reporttablebody} 
\begin{longtable}[]{@{}
  >{\raggedright\arraybackslash}p{(\linewidth - 4\tabcolsep) * \real{0.1630}}
  >{\raggedright\arraybackslash}p{(\linewidth - 4\tabcolsep) * \real{0.2473}}
  >{\raggedright\arraybackslash}p{(\linewidth - 4\tabcolsep) * \real{0.5678}}@{}}
\toprule\noalign{}
\begin{minipage}[b]{\linewidth}\raggedright
\textbf{Augmentation}
\end{minipage} & \begin{minipage}[b]{\linewidth}\raggedright
\textbf{Generated content}
\end{minipage} & \begin{minipage}[b]{\linewidth}\raggedright
\textbf{Example}
\end{minipage} \\
\midrule\noalign{}
\endhead
\bottomrule\noalign{}
\endlastfoot
Synthetic QA & A practical legal question and a detailed answer grounded
in the legal reasoning of the source document. & Question: An insolvency
administrator violated bankruptcy law without causing actual adverse
consequences. Can the violation be deemed insignificant under Article
2.9 of the Russian Code of Administrative Offenses? Answer: No. The
absence of adverse consequences alone does not establish insignificance.
The offense under Article 14.13(3) does not require proof of adverse
consequences; it is established when the statutory bankruptcy
requirements are violated. {[}\ldots{]} \\
Textbook rewrite & A structured explanation of the legal framework and
hierarchy of legal acts. & Legal framework for establishing commissions.
Municipal commissions for juvenile affairs and protection of their
rights operate under a system of legal acts. Their creation and changes
in membership are regulated legal procedures. At the federal level,
Federal Law No. 131-FZ of October 6, 2003 defines the competence of
municipal representative bodies regarding the creation and membership of
structural units. {[}\ldots{]} \\
\end{longtable}
}

\endgroup

To limit hallucinations in synthetic data, we checked citations to legal
acts, excluded acts no longer in force from the corpus, and removed
incorrect references and repetitive generations. Training on the
resulting legal corpus and augmented data improved the expert law
benchmark score by 5 percentage points in a small-scale training run.

\subsubsection{Medical Data}\label{medical-data}

The medical corpus includes publications from PubMed Central and the
Cochrane Library, WHO documents, Russian Ministry of Health clinical
guidelines, medical textbooks, and content from Russian websites,
encyclopedias, and forums. PDF and DjVu documents were converted to text
using optical character recognition.

To evaluate augmentation methods, we retrieved publications relevant to
challenging medical benchmark questions using keyword search and
embedding-based retrieval, then assessed their relevance using a large
language model. We compared concise summaries, textbook-style rewrites,
and dialogues (Table 17). An example dialogue, translated from Russian,
follows.

\Needspace{9\baselineskip}\begingroup\small\setlength{\LTpre}{4pt}\setlength{\LTpost}{5pt}

\captionof{table}{Augmentation formats for medical documents.}

\nopagebreak[4]

\small\setlength{\tabcolsep}{3pt}

{\def\LTcaptype{reporttablebody} 
\begin{longtable}[]{@{}
  >{\raggedright\arraybackslash}p{(\linewidth - 2\tabcolsep) * \real{0.2102}}
  >{\raggedright\arraybackslash}p{(\linewidth - 2\tabcolsep) * \real{0.7652}}@{}}
\toprule\noalign{}
\begin{minipage}[b]{\linewidth}\raggedright
\textbf{Augmentation}
\end{minipage} & \begin{minipage}[b]{\linewidth}\raggedright
\textbf{Description}
\end{minipage} \\
\midrule\noalign{}
\endhead
\bottomrule\noalign{}
\endlastfoot
Concise summary & A condensed version of a document or section that
preserves its key medical information. \\
Textbook rewrite & A rewrite in the style of a medical textbook chapter;
one variant adds self-assessment questions. \\
Dialogue scenario & Characters discuss symptoms, disease mechanisms,
diagnosis, or treatment. \\
\end{longtable}
}

\endgroup

\emph{\textbf{Example.} Synthetic dialogue generated from a medical
article (translated from Russian).}

\Needspace{9\baselineskip}\begingroup\small\setlength{\LTpre}{4pt}\setlength{\LTpost}{5pt}

\begin{quote}
\emph{Setting: a small hospital staff room with a coffee machine and a
courtyard window. Dr. Petrova and Dr. Ivanova are sitting with mugs.
Alexey joins them with a printed article.}
\end{quote}

\nopagebreak[4]

\small\setlength{\tabcolsep}{3pt}

{\def\LTcaptype{reporttablebody} 
\begin{longtable}[]{@{}
  >{\raggedright\arraybackslash}p{(\linewidth - 2\tabcolsep) * \real{0.1588}}
  >{\raggedright\arraybackslash}p{(\linewidth - 2\tabcolsep) * \real{0.8182}}@{}}
\toprule\noalign{}
\begin{minipage}[b]{\linewidth}\raggedright
\textbf{Speaker}
\end{minipage} & \begin{minipage}[b]{\linewidth}\raggedright
\textbf{Dialogue}
\end{minipage} \\
\midrule\noalign{}
\endhead
\bottomrule\noalign{}
\endlastfoot
\textbf{Alexey} & I was reading about iron in enzymes. What other
proteins contain iron besides hemoglobin and cytochromes? \\
\textbf{Dr. Petrova} & Catalase and peroxidases, for example, are
heme-containing enzymes that break down hydrogen peroxide. Iron in their
active sites helps neutralize reactive oxygen species. \\
\textbf{Dr. Ivanova} & Copper is also involved in iron metabolism. \\
\textbf{Dr. Petrova} & Exactly. Ceruloplasmin oxidizes
Fe\textsuperscript{2+} to Fe\textsuperscript{3+}, allowing iron to bind
to transferrin. With copper deficiency, iron accumulates in cells and
enters the bloodstream less efficiently. \\
\textbf{Alexey} & So copper deficiency can resemble iron deficiency? \\
\textbf{Dr. Petrova} & Clinically, yes: anemia and fatigue may occur.
But the mechanisms differ, so iron should not be prescribed to every
patient with these symptoms. {[}\ldots{]} \\
\end{longtable}
}

\endgroup

In controlled training experiments, models trained on rewritten articles
achieved substantially higher scores than models trained on the original
articles. The results also depended on the model used to generate the
rewrites. Documents retrieved using benchmark questions were used only
to compare augmentation formats and develop the pipeline; they were not
directly included in the final corpus. Training on the expanded medical
corpus and augmented data improved the medical education benchmark score
by 2.7 percentage points in a short training run and the expert medical
benchmark score by 4 percentage points in a longer run.

\subsection{Mathematical Web Data: Selection and
Augmentation}\label{mathematical-web-data-selection-and-augmentation}

\subsubsection{Source Selection}\label{source-selection}

The mathematical web pipeline starts from the same raw web corpus as the
factual-data pipeline in Section 6.3. It uses a high-recall 0.5B filter
followed by an 8B multiclass classifier that assigns quality tiers L1,
L2, and L3. We evaluate the classifiers on mathematical documents
labeled by an LLM and examples from public corpora. An earlier 180M
classifier was fast and achieved high precision, but discarded too many
useful documents and was excluded from the pipeline. Table 18 compares
filtered corpus sizes, measured with our internal tokenizer.

\Needspace{9\baselineskip}\begingroup\small\setlength{\LTpre}{4pt}\setlength{\LTpost}{5pt}

\captionof{table}{Sizes of filtered mathematical web corpora. Token counts
use our internal tokenizer.}

\nopagebreak[4]

\small\setlength{\tabcolsep}{3pt}

{\def\LTcaptype{reporttablebody} 
\begin{longtable}[]{@{}
  >{\raggedright\arraybackslash}p{(\linewidth - 2\tabcolsep) * \real{0.5000}}
  >{\raggedright\arraybackslash}p{(\linewidth - 2\tabcolsep) * \real{0.5000}}@{}}
\toprule\noalign{}
\begin{minipage}[b]{\linewidth}\raggedright
\textbf{Dataset}
\end{minipage} & \begin{minipage}[b]{\linewidth}\raggedright
\textbf{Tokens}
\end{minipage} \\
\midrule\noalign{}
\endhead
\bottomrule\noalign{}
\endlastfoot
UltraData-Math-L2 & 37B \\
Nemotron-CC-Math-v1 (4plus) & 59B \\
Ours & 177B \\
\end{longtable}
}

\endgroup

\subsubsection{Augmentation}\label{augmentation}

For factual data, augmentations present related facts together in varied
contexts. For mathematical data, they must preserve complete
definitions, problem statements, reasoning steps, and solutions while
varying their presentation. We also generate new problems and solutions
based on each source document.

We used 19 augmentation formats inspired by Nemotron-MIND\footnote{\url{https://arxiv.org/pdf/2410.12881}}
and UltraData-Math and adapted for Russian-language text (Table 19).
These formats cover problem generation, dialogues, stylistic rewrites,
and knowledge extraction. For each selected document, we generated
augmented text in all 19 formats. We applied the procedure to
approximately 10\% of the mathematical web corpus. The resulting
augmentation dataset contains 115B tokens, comparable in scale to major
open mathematical augmentation corpora (Table 20).

\Needspace{9\baselineskip}\begingroup\small\setlength{\LTpre}{4pt}\setlength{\LTpost}{5pt}

\captionof{table}{The 19 mathematical augmentation formats.}

\nopagebreak[4]

\small\setlength{\tabcolsep}{3pt}

{\def\LTcaptype{reporttablebody} 
\begin{longtable}[]{@{}
  >{\raggedright\arraybackslash}p{(\linewidth - 4\tabcolsep) * \real{0.1600}}
  >{\raggedright\arraybackslash}p{(\linewidth - 4\tabcolsep) * \real{0.3500}}
  >{\raggedright\arraybackslash}p{(\linewidth - 4\tabcolsep) * \real{0.4901}}@{}}
\toprule\noalign{}
\begin{minipage}[b]{\linewidth}\raggedright
\textbf{Group}
\end{minipage} & \begin{minipage}[b]{\linewidth}\raggedright
\textbf{Augmentation type}
\end{minipage} & \begin{minipage}[b]{\linewidth}\raggedright
\textbf{Description}
\end{minipage} \\
\midrule\noalign{}
\endhead
\bottomrule\noalign{}
\endlastfoot
Problems and answers & qa\_grade\_school & An elementary-school word
problem based on the source document. \\
& qa\_middle\_school & A middle-school problem with a detailed
solution. \\
& qa\_high\_school & A high-school problem combining concepts from at
least two areas of mathematics. \\
& qa\_college & A university-level problem based on the source ideas. \\
Dialogues & conversation\_problem\_solving & A multistep dialogue
working through a mathematical problem. \\
& conversation\_teacher\_student & A teacher guides a student through
the reasoning. \\
& conversation\_two\_students & Two students solve an assignment
collaboratively. \\
& conversation\_two\_professors & Two specialists discuss a mathematical
topic in depth. \\
& conversation\_layman\_expert & An expert explains difficult material
to a nonspecialist. \\
& conversation\_interview & An interview with sequential questions and
answers. \\
& conversation\_debate & A debate comparing positions and arguments. \\
Rewrites & rewrite\_textbook & A structured textbook passage. \\
& rewrite\_lecture\_note & Lecture notes. \\
& rewrite\_learning\_note & Compact study notes. \\
& rewrite\_popular\_science & An accessible popular-science explanation
of mathematical ideas. \\
& rewrite\_blog & An informal blog post on the source topic.. \\
& rewrite\_academic\_paper & A formal presentation in the style of an
academic paper. \\
& rewrite\_wiki & An encyclopedia-style article. \\
Knowledge extraction & knowledge\_extraction & Extraction of
mathematical concepts, statements, and other key knowledge. \\
\end{longtable}
}

\endgroup

\Needspace{9\baselineskip}\begingroup\small\setlength{\LTpre}{4pt}\setlength{\LTpost}{5pt}

\captionof{table}{Mathematical augmentation corpora (tokens).}

\nopagebreak[4]

\small\setlength{\tabcolsep}{3pt}

{\def\LTcaptype{reporttablebody} 
\begin{longtable}[]{@{}
  >{\raggedright\arraybackslash}p{(\linewidth - 2\tabcolsep) * \real{0.5000}}
  >{\raggedright\arraybackslash}p{(\linewidth - 2\tabcolsep) * \real{0.5000}}@{}}
\toprule\noalign{}
\begin{minipage}[b]{\linewidth}\raggedright
\textbf{Dataset}
\end{minipage} & \begin{minipage}[b]{\linewidth}\raggedright
\textbf{Tokens}
\end{minipage} \\
\midrule\noalign{}
\endhead
\bottomrule\noalign{}
\endlastfoot
UltraData-Math-L3 & 89B \\
Nemotron-MIND v1 & 81B \\
Ours (augmentation of 10\% of the source corpus) & 115B \\
\end{longtable}
}

\endgroup

These data produced statistically significant gains in small-scale
training experiments. Improvements were evident in base-model
evaluations and were most consistent after all compared models underwent
the same SFT procedure. Scores increased by 3--10 percentage points on
all evaluated mathematics benchmarks, from school and university
mathematics to difficult reasoning and competition problems.

\subsection{Programming Web Data: Selection and
Augmentation}\label{programming-web-data-selection-and-augmentation}

We collected 245B tokens of raw programming web data and 200B tokens of
augmentations. We used only stylistic rewrites, designed to preserve the
technical content of the source, including program entities, APIs,
commands, dependencies, examples, edge cases, and code semantics. Table
21 lists the seven formats.

\Needspace{9\baselineskip}\begingroup\small\setlength{\LTpre}{4pt}\setlength{\LTpost}{5pt}

\captionof{table}{Stylistic augmentation formats for programming web
data.}

\nopagebreak[4]

\small\setlength{\tabcolsep}{3pt}

{\def\LTcaptype{reporttablebody} 
\begin{longtable}[]{@{}
  >{\raggedright\arraybackslash}p{(\linewidth - 4\tabcolsep) * \real{0.1600}}
  >{\raggedright\arraybackslash}p{(\linewidth - 4\tabcolsep) * \real{0.2600}}
  >{\raggedright\arraybackslash}p{(\linewidth - 4\tabcolsep) * \real{0.5800}}@{}}
\toprule\noalign{}
\begin{minipage}[b]{\linewidth}\raggedright
\textbf{Group}
\end{minipage} & \begin{minipage}[b]{\linewidth}\raggedright
\textbf{Augmentation type}
\end{minipage} & \begin{minipage}[b]{\linewidth}\raggedright
\textbf{Description}
\end{minipage} \\
\midrule\noalign{}
\endhead
\bottomrule\noalign{}
\endlastfoot
Stylistic rewrites & wiki & Encyclopedic exposition with modular
structure, standardized terminology, and a neutral tone. \\
& textbook & Sequential instruction covering concepts, syntax, rules,
procedures, examples, and conclusions. \\
& blog & An accessible technical blog with short sections and practical
emphasis. \\
& popular\_science & Programming concepts explained through analogies
and practical examples, minimizing unexplained jargon. \\
& academic\_paper & Formal, logically structured exposition with precise
technical terminology. \\
& learning\_note & Compact self-study notes with key concepts, short
explanations, and review questions. \\
& lecture\_note & Structured lecture notes with code examples,
implementation details, and common difficulties. \\
\end{longtable}
}

\endgroup

With the SFT procedure held fixed, adding these data during pretraining
improved scores on most programming benchmarks. The largest gains were
on FixEval (+14.7 percentage points) and LiveCodeBench pass@5 (+6.5
percentage points), with gains of 2--3.5 percentage points on MBPP and
EvalPlus.

\subsection{STEM Web Data: Selection and
Augmentation}\label{stem-web-data-selection-and-augmentation}

Initial filtering with a broad STEM classifier produced a corpus
containing several trillion tokens. Adding this corpus improved scores
on engineering benchmarks but reduced performance elsewhere, suggesting
that broad selection retained too much irrelevant material. We applied
additional filters for expert knowledge and selected physics documents,
reducing the corpus to several hundred billion tokens. We applied the
same augmentation formats as for mathematical data. Small-scale training
experiments showed gains on STEM and engineering benchmarks, as well as
on mathematics benchmarks after SFT.

\section{Reasoning and Agentic
Capabilities}\label{reasoning-and-agentic-capabilities}

\subsection{Why a Separate Reasoning
Stage?}\label{why-a-separate-reasoning-stage}

Standard mathematical and programming benchmarks, including MATH-500,
HumanEval, and MBPP, approach saturation for strong base models. We
therefore focus on more challenging benchmarks and report pass@k, the
fraction of problems solved in at least one of k attempts. Subsequent RL
may increase the probability of generating successful trajectories that
the base model can already produce. This motivates introducing reasoning
data before post-training, an approach also explored in OctoThinker and
Front-Loading Reasoning {[}38, 39{]}.

\begingroup\setlength{\emergencystretch}{3em}
Other publicly released models also incorporate reasoning data during
pretraining. Nemotron 3 Super reports AIME pass@32 {[}14{]}. Qwen3.5
also emphasizes STEM and reasoning data {[}40{]}, and its base model
performs well on challenging reasoning benchmarks.\par\endgroup

Adding reasoning examples to the general training mixture improved
pass@k, but concentrating the same data in a separate final stage was
substantially more effective. We hypothesize that the general
capabilities acquired during earlier training allow the model to benefit
more from these challenging examples. Table 22 compares checkpoints
before and after this stage. The largest gains were observed on advanced
mathematics, competitive programming, and STEM competition benchmarks,
reaching tens of percentage points. Gains on conventional CoT benchmarks
were smaller, partly because scores before the reasoning stage were
already high.

\Needspace{9\baselineskip}\begingroup\small\setlength{\LTpre}{4pt}\setlength{\LTpost}{5pt}

\captionof{table}{Performance before and after the reasoning stage (\%).
Starred OlympBench subsets were constructed internally from existing
English-language competition problems.}

\nopagebreak[4]

\small\setlength{\tabcolsep}{3pt}

{\def\LTcaptype{reporttablebody} 
\begin{longtable}[]{@{}
  >{\raggedright\arraybackslash}p{(\linewidth - 8\tabcolsep) * \real{0.3901}}
  >{\raggedright\arraybackslash}p{(\linewidth - 8\tabcolsep) * \real{0.0700}}
  >{\raggedright\arraybackslash}p{(\linewidth - 8\tabcolsep) * \real{0.1800}}
  >{\raggedright\arraybackslash}p{(\linewidth - 8\tabcolsep) * \real{0.1800}}
  >{\raggedright\arraybackslash}p{(\linewidth - 8\tabcolsep) * \real{0.1800}}@{}}
\toprule\noalign{}
\begin{minipage}[b]{\linewidth}\raggedright
\textbf{Benchmark}
\end{minipage} & \begin{minipage}[b]{\linewidth}\raggedright
\textbf{Lang.}
\end{minipage} & \begin{minipage}[b]{\linewidth}\raggedright
\textbf{Before reasoning stage}
\end{minipage} & \begin{minipage}[b]{\linewidth}\raggedright
\textbf{After reasoning stage}
\end{minipage} & \begin{minipage}[b]{\linewidth}\raggedright
\textbf{Change (pp)}
\end{minipage} \\
\midrule\noalign{}
\endhead
\bottomrule\noalign{}
\endlastfoot
\multicolumn{5}{@{}>{\raggedright\arraybackslash}p{(\linewidth - 8\tabcolsep) * \real{1.0000} + 8\tabcolsep}@{}}{%
\emph{Programming}} \\
Codeforces C++ 0-shot pass@8 & EN & 23.3 & 68.9 & +45.6 \\
LiveCodeBench v5--v6 0-shot pass@5 & EN & 37.1 & 80.9 & +43.8 \\
LiveCodeBench v5--v6 CoT 1-shot pass@5 & EN & 63.13 & 69.03 & +5.90 \\
\multicolumn{5}{@{}>{\raggedright\arraybackslash}p{(\linewidth - 8\tabcolsep) * \real{1.0000} + 8\tabcolsep}@{}}{%
\emph{STEM competitions}} \\
OlympBench Phys/Chem pass@8* & EN & 51.81 & 85.54 & +33.73 \\
\multicolumn{5}{@{}>{\raggedright\arraybackslash}p{(\linewidth - 8\tabcolsep) * \real{1.0000} + 8\tabcolsep}@{}}{%
\emph{Conventional mathematics}} \\
MATH-500 & EN & 86.6 & 91.1 & +4.5 \\
EduBench Math & RU & 78.9 & 79.3 & +0.4 \\
EduBench Math University & RU & 67.9 & 70.1 & +2.2 \\
\multicolumn{5}{@{}>{\raggedright\arraybackslash}p{(\linewidth - 8\tabcolsep) * \real{1.0000} + 8\tabcolsep}@{}}{%
\emph{Advanced mathematics}} \\
AIME 2026 pass@32 & EN & 33.3 & 96.7 & +63.4 \\
HMMT Feb 2026 pass@32 & EN & 24.3 & 96.9 & +72.6 \\
IMO-AnswerBench pass@8 & EN & 30.7 & 88.7 & +58.0 \\
OlympBench Math pass@8* & EN & 58.82 & 94.1 & +35.28 \\
\end{longtable}
}

\endgroup

After applying the same cold-start supervised fine-tuning procedure to
both checkpoints, the checkpoint trained with the additional reasoning
stage achieved pass@1 scores 3.8 percentage points higher on HMMT 2026
and 6.6 percentage points higher on Codeforces. These gains suggest that
the reasoning stage does more than teach a response format.

Table 23 reports agentic evaluations of both checkpoints after the same
cold-start training procedure. Scores improve on every listed benchmark,
including under both pass@1 and pass@4 evaluation. These results show
that the benefits of the reasoning stage extend to agentic tasks after
cold-start training. We release only the checkpoint after the reasoning
stage.

\Needspace{9\baselineskip}\begingroup\small\setlength{\LTpre}{4pt}\setlength{\LTpost}{5pt}

\captionof{table}{Agentic evaluation of checkpoints taken before and after
the reasoning stage, with the same cold-start training procedure applied
to both (\%).}

\nopagebreak[4]

\small\setlength{\tabcolsep}{3pt}

{\def\LTcaptype{reporttablebody} 
\begin{longtable}[]{@{}
  >{\raggedright\arraybackslash}p{(\linewidth - 6\tabcolsep) * \real{0.4801}}
  >{\raggedright\arraybackslash}p{(\linewidth - 6\tabcolsep) * \real{0.2000}}
  >{\raggedright\arraybackslash}p{(\linewidth - 6\tabcolsep) * \real{0.2000}}
  >{\raggedright\arraybackslash}p{(\linewidth - 6\tabcolsep) * \real{0.1200}}@{}}
\toprule\noalign{}
\begin{minipage}[b]{\linewidth}\raggedright
\textbf{Benchmark}
\end{minipage} & \begin{minipage}[b]{\linewidth}\raggedright
\textbf{Before reasoning + cold start}
\end{minipage} & \begin{minipage}[b]{\linewidth}\raggedright
\textbf{After reasoning + cold start}
\end{minipage} & \begin{minipage}[b]{\linewidth}\raggedright
\textbf{Change (pp)}
\end{minipage} \\
\midrule\noalign{}
\endhead
\bottomrule\noalign{}
\endlastfoot
BFCL acc pass@1 & 67.8 & 69.2 & +1.4 \\
DeepPlanning travel composite pass@1 & 15.8 & 20.8 & +5.0 \\
DeepPlanning shopping acc pass@1 & 26.7 & 29.2 & +2.5 \\
Tau2 pass@1 & 69.1 & 73.8 & +4.7 \\
Tau2 pass@4 & 93.1 & 94.9 & +1.8 \\
VitaBench pass@1 & 15.7 & 23.6 & +7.9 \\
VitaBench pass@4 & 36.5 & 52.0 & +15.5 \\
SWE-bench Verified, OpenHands pass@1 & 31.0 & 54.2 & +23.2 \\
\end{longtable}
}

\endgroup

\subsection{Reasoning Traces}\label{reasoning-traces}

\subsubsection{Mathematics}\label{mathematics}

Our mathematical reasoning data combine public post-training datasets
with problems extracted from our mathematical web corpus. We retain the
original reasoning traces for some datasets and generate new ones for
others. We also extracted 45,000 Russian-language and 6,000
English-language problems from PDFs. Unlike the broader mathematical web
data in Section 6.8, we focus here on problems requiring advanced
mathematical reasoning and carefully validate both problem statements
and solutions.

We also constructed MaRS (Mathematical Reasoning Set), a 27B-token
corpus derived from our internal collection of web documents. Figure 19
summarizes the pipeline. The initial database contains approximately
330--370B documents; basic parsing and prefiltering leave about 140B.
Subsequent steps are as follows:

\begin{enumerate}
\def\labelenumi{\arabic{enumi}.}
\item
  A mathematical-content classifier uses a deliberately low threshold of
  0.3 to preserve recall, retaining about 1\% of the 140B documents, or
  1.4B documents.
\item
  Chunking yields approximately 1.6B chunks. An LLM extracts problems,
  which are deduplicated and filtered using heuristic rules based on
  regular expressions and a minimum problem-statement length of ten
  words. Approximately 980M problems remain. The retained percentages
  reported below are measured relative to this set of 980M problems.
\item
  An additional mathematics filter removes nonmathematical content,
  underspecified tasks, and purely theoretical passages that do not pose
  a task. Approximately 70\%, or 685M problems, remain.
\item
  We classify problems by difficulty: basic school-level mathematics,
  high-school and introductory university mathematics, advanced
  undergraduate and graduate mathematics, and competition-level
  mathematics. Selecting competition-level problems leaves about 8\%, or
  78M.
\item
  Task-type classification separates computation, proof, construction,
  and existence questions. The main corpus retains computational tasks;
  other types are routed to separate branches. About 4\%, or 39M,
  remain.
\item
  A large language model checks problem statements for parsing errors,
  references to missing context, and missing questions. This leaves
  about 3\%, or 29M.
\item
  We generate eight solutions per problem, apply majority-vote filtering
  at threshold of 0.5, remove programming tasks, filter by language, and
  deduplicate. This final stage removes approximately 97\% of the 29M
  problems entering it, leaving about 1M problems with
  solutions---roughly 0.1\% of the original 980M extracted problems.
\end{enumerate}

\par\addvspace{8pt}\noindent\begin{minipage}{\linewidth}\centering

\includegraphics[width=\linewidth,height=.46\textheight,keepaspectratio]{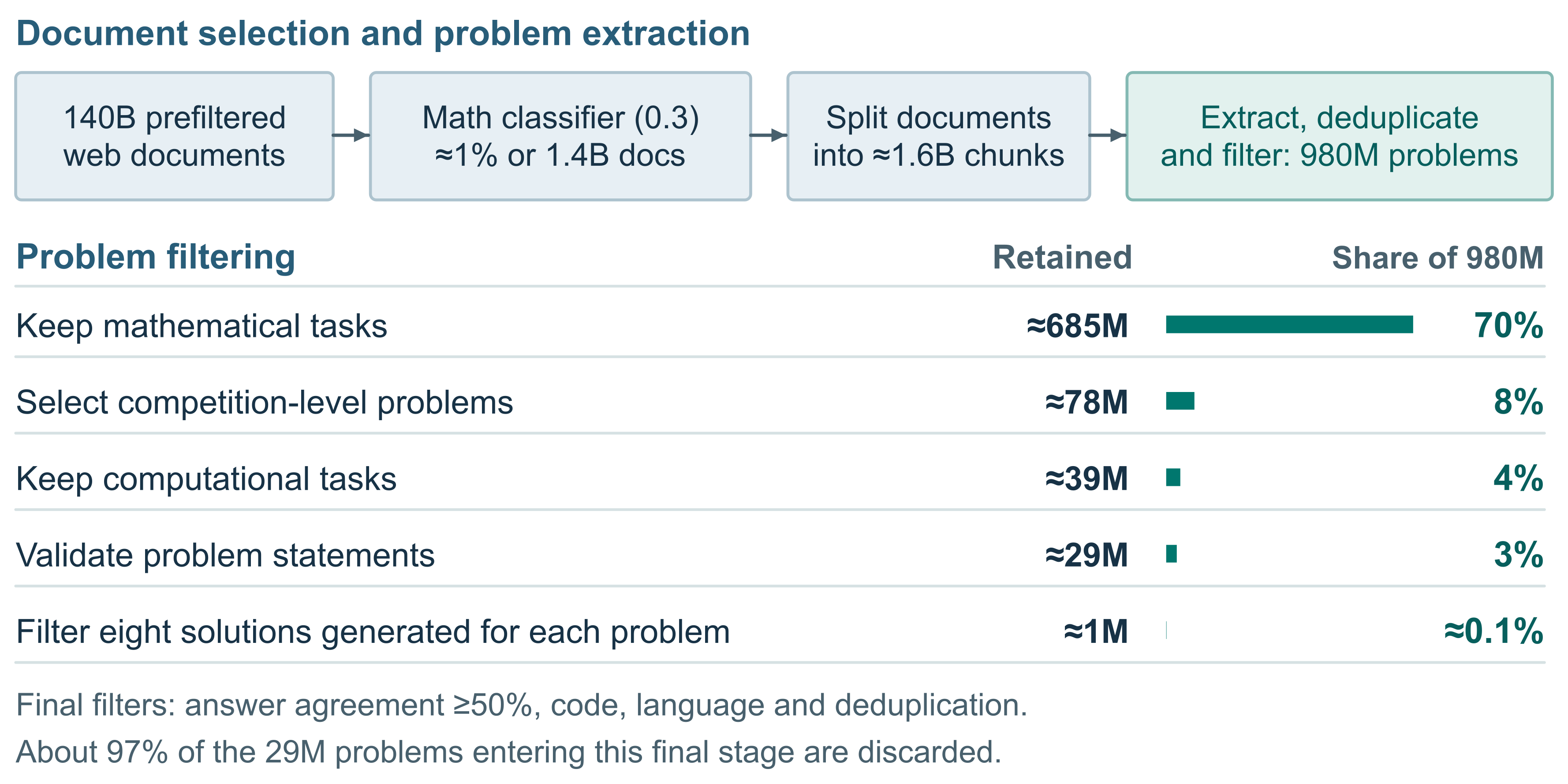}

\par

\captionof{figure}{MaRS extraction and filtering pipeline. The initial
document-retention percentage is relative to the 140B prefiltered
documents. Retained percentages in the subsequent problem-filtering
stages are relative to the 980M extracted problems. Counts and
percentages are approximate.}

\end{minipage}\par\addvspace{7pt}

The pipeline yields approximately one retained problem per 140,000
documents in the prefiltered collection.

Reference answers are unavailable for most problems, so we filter
generated solutions by agreement across independent samples. We generate
eight independent solutions and use a separate LLM to extract each final
answer, since it is not always formatted as \textbackslash boxed\{\}.
For each problem, we identify the most frequent answer and define the
majority score as the fraction of generated solutions that produce it.
We retain solutions matching this answer if the majority score is at
least 0.5; otherwise, we discard the problem. This also removes many
incomplete or ambiguous problems. We can therefore retain multiple
solutions to the same problem.

Long traces are particularly important for challenging math problems.
Limiting training data to solutions shorter than 8,000 tokens
substantially reduced performance, while removing short solutions had no
adverse effect or slightly improved scores. We therefore retain traces
longer than 8,000 tokens in the final training data. We additionally
remove repetitive generations and programming tasks. The latter were
identified using an LLM applied to problem statements and
regular-expression filters applied after generation. The removed
programming tasks accounted for approximately 600M tokens.

In validation experiments, training on these data improved scores on
AIME, HMMT, and IMO benchmarks both before and after cold-start
training. The gains grew as we increased the dataset size, and the same
pattern held in the final training run. Further experiments yielded the
following observations:

\begin{itemize}
\item
  Counting occurrences of reasoning patterns---backtracking,
  verification, subgoal setting, and useful steps---was computationally
  expensive, and the resulting counts were strongly correlated with
  trace length. Selecting traces containing backtracking offered only
  marginal gains beyond length filtering, while selecting for
  verification or subgoal setting yielded no additional gains. These
  filters were omitted from the final pipeline.
\item
  Rejecting solutions that disagreed with answers extracted from the
  source documents reduced performance. Inspection revealed many
  mismatches caused by extraction or parsing errors, including missing
  problem conditions. Correct generated solutions could therefore be
  incorrectly rejected.
\item
  Majority-score filtering itself was beneficial, but its exact
  threshold had limited effect. We tested the thresholds of 0.375, 0.5,
  0.75, and filtering by score ranges.
\item
  Problem difficulty matters as well as trace length. Training on
  solutions generated by strong models for easier problems reduced
  performance on challenging mathematics benchmarks in this experiment.
\end{itemize}

\subsubsection{Programming}\label{programming}

We constructed a synthetic corpus of challenging competitive-programming
problems and used strong models to generate solutions. The first
approach involved the model generating problems based on ideas from
computer science papers, including complete problem statements,
constraints, examples, and code for generating and checking tests.
Another approach was to generate a list of 1,000 algorithms and created
problems based on algorithms from this list. We further increased
difficulty by providing the model with two problems and their respective
solutions, then asking it to generate a new problem combining ideas from
both.

For each problem, we generated detailed reasoning, a Python solution,
and tests. Code execution allows us to filter these data automatically:
we check generated solutions against reference tests and validated
generated tests using known-correct solutions. This removed
nonfunctional code, incorrect tests, and contradictions between
statements and examples.

\subsection{Reasoning Alone Is Insufficient for
Agents}\label{reasoning-alone-is-insufficient-for-agents}

Our initial experiments focused on difficult standalone reasoning tasks.
We compared a checkpoint trained through the reasoning stage without
tool-use data against Qwen3A-35B-Base, applying the same agentic
cold-start training procedure to both. The checkpoint without tool-use
training substantially underperformed Qwen on most agentic benchmarks.
Adding AWM- and GEM-style data during pretraining substantially narrowed
the gap. The resulting model matched or exceeded Qwen on some tasks.

We also compared reinforcement learning with verifiable rewards (RLVR)
for base models pretrained with and without agentic data, after applying
the same cold-start training procedure. During RLVR, the model
pretrained with agentic data trained more stably and reached a higher
level of performance.

\subsection{Agentic Data}\label{agentic-data}

We use several complementary sources. Following the GEM approach of Xu
et al. {[}41{]}, we convert multistep text into tool-use trajectories.
We also include trajectories generated during our own RL runs, an
approach related to continual agent pretraining {[}42{]}. We generate
tool-use trajectories in synthetic environments inspired by AWM and
construct separate datasets for search and software engineering.

\subsubsection{CRUD Scenarios}\label{crud-scenarios}

Many practical agent tasks require working with mutable state: finding
and creating objects, updating fields, deleting records, and checking
the resulting state. Successful behavior requires selecting tools,
planning calls, passing data between them, and preserving consistency.
Inspired by Agent World Model {[}43{]}, the Alice AI LLM post-training
team built a pipeline that generates executable environments, tasks of
varying difficulty, and rubrics for automated evaluation (Figure 20).

\par\addvspace{8pt}\noindent\begin{minipage}{\linewidth}\centering

\includegraphics[width=\linewidth,height=.46\textheight,keepaspectratio]{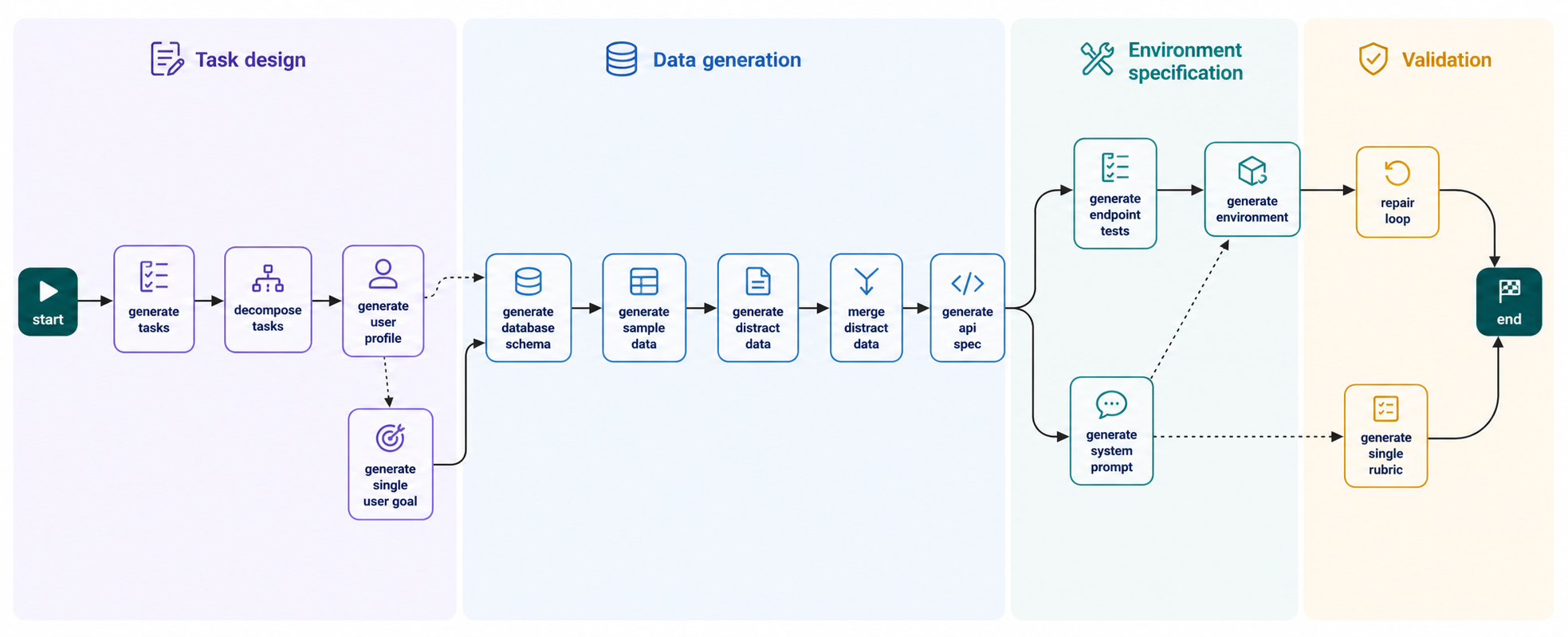}

\par

\captionof{figure}{Synthetic CRUD environment and task generation,
including data generation, executable environments, validation, and
evaluation rubrics.}

\end{minipage}\par\addvspace{7pt}

Generated environments contain many tables and tools, but the tool
schemas contain little nesting and relatively few parameters (Figure
21). Solving these tasks nevertheless requires long sequences of tool
calls. Table 24 reports GLM-5.2 trajectory statistics on easy and hard
tasks. We used trajectory length as a proxy for difficulty and confirmed
manually that the harder tasks require longer, substantive action
sequences.

\par\addvspace{8pt}\noindent\begin{minipage}{\linewidth}\centering

\includegraphics[width=\linewidth,height=.46\textheight,keepaspectratio]{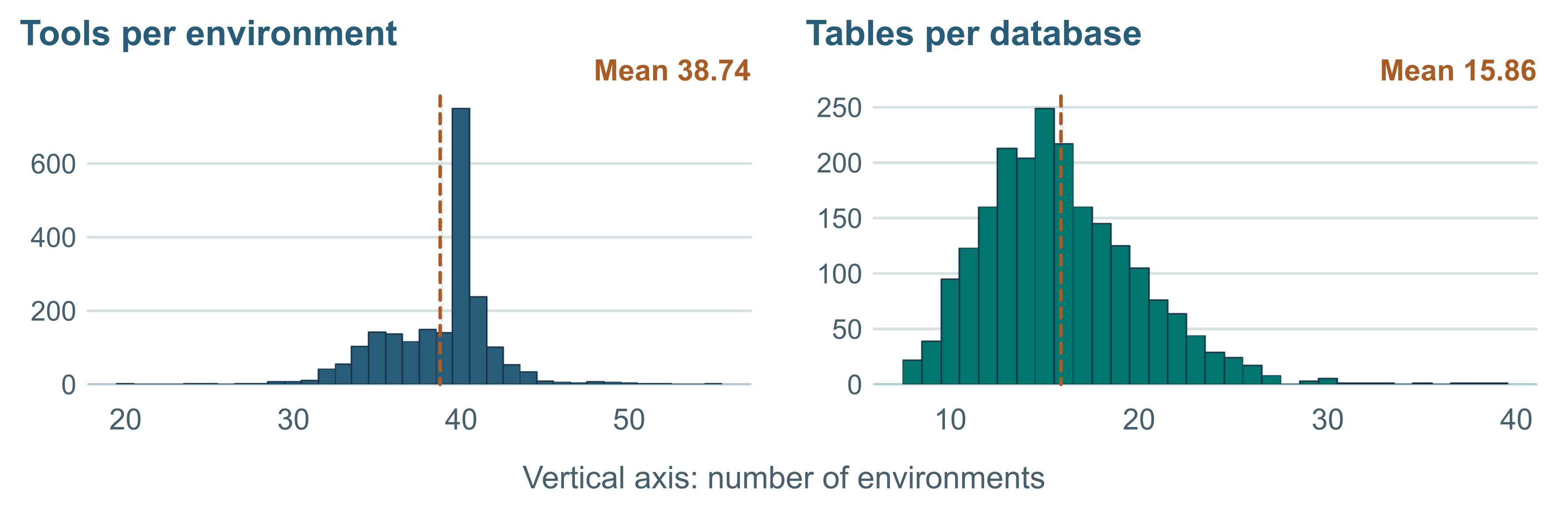}

\par

\captionof{figure}{Distributions of tool and table counts in generated
environments.}

\end{minipage}\par\addvspace{7pt}

\Needspace{9\baselineskip}\begingroup\small\setlength{\LTpre}{4pt}\setlength{\LTpost}{5pt}

\captionof{table}{GLM-5.2 trajectory statistics on generated easy and hard
CRUD tasks. pass@4 is in percent.}

\nopagebreak[4]

\small\setlength{\tabcolsep}{3pt}

{\def\LTcaptype{reporttablebody} 
\begin{longtable}[]{@{}
  >{\raggedright\arraybackslash}p{(\linewidth - 4\tabcolsep) * \real{0.3333}}
  >{\raggedright\arraybackslash}p{(\linewidth - 4\tabcolsep) * \real{0.3333}}
  >{\raggedright\arraybackslash}p{(\linewidth - 4\tabcolsep) * \real{0.3333}}@{}}
\toprule\noalign{}
\begin{minipage}[b]{\linewidth}\raggedright
\end{minipage} & \begin{minipage}[b]{\linewidth}\raggedright
\textbf{easy}
\end{minipage} & \begin{minipage}[b]{\linewidth}\raggedright
\textbf{hard}
\end{minipage} \\
\midrule\noalign{}
\endhead
\bottomrule\noalign{}
\endlastfoot
Mean trajectory length (steps) & 12.63 & 26.75 \\
pass@4 & 58.1 & 32.1 \\
\end{longtable}
}

\endgroup

Original AWM tasks and environments were excluded from training and used
to construct a held-out benchmark. For faster evaluation, we selected
600 tasks with the highest variance in scores across five internal and
external models. Manual inspection found broken environments and noisy
tasks, so absolute scores should be interpreted cautiously. The
benchmark was nevertheless useful for comparing pretraining data
configurations (Table 25).

\Needspace{9\baselineskip}\begingroup\small\setlength{\LTpre}{4pt}\setlength{\LTpost}{5pt}

\captionof{table}{Scores on the internal 600-task AWM-derived evaluation.
Absolute scores are affected by noisy tasks and broken environments.}

\nopagebreak[4]

\small\setlength{\tabcolsep}{3pt}

{\def\LTcaptype{reporttablebody} 
\begin{longtable}[]{@{}
  >{\raggedright\arraybackslash}p{(\linewidth - 2\tabcolsep) * \real{0.5000}}
  >{\raggedright\arraybackslash}p{(\linewidth - 2\tabcolsep) * \real{0.5000}}@{}}
\toprule\noalign{}
\begin{minipage}[b]{\linewidth}\raggedright
\textbf{Model}
\end{minipage} & \begin{minipage}[b]{\linewidth}\raggedright
\textbf{avg reward, \%}
\end{minipage} \\
\midrule\noalign{}
\endhead
\bottomrule\noalign{}
\endlastfoot
Our early checkpoint after cold-start training & 48.3\% \\
GLM-4.5 & 69.8\% \\
GLM-4.7 & 67.3\% \\
GPT-5.2 Thinking & 77.8\% \\
GLM-5 & 80.1\% \\
\end{longtable}
}

\endgroup

Expanding domain coverage improved results more than generating
additional similar tasks within existing domains. We therefore provided
the existing list of domains when generating new ones to reduce
duplication. Four independent interaction trajectories were sampled per
task, producing a 7B-token agentic corpus.

\subsubsection{Search Questions}\label{search-questions}

We generate synthetic difficult multistep search questions from web
documents. Starting from a web page, the pipeline generates a factual
question asking for an entity name, searches for a new document using
that entity, extracts a related entity, and repeats. This forms a chain
of up to ten transitions. We then formulate a single question that
requires tracing the chain in reverse and is designed to take at least
eight search steps to solve. We use automated critique to check clarity,
answerability, and possible shortcuts, then revise the questions
accordingly (Figure 22).

\par\addvspace{8pt}\noindent\begin{minipage}{\linewidth}\centering

\includegraphics[width=\linewidth,height=.46\textheight,keepaspectratio]{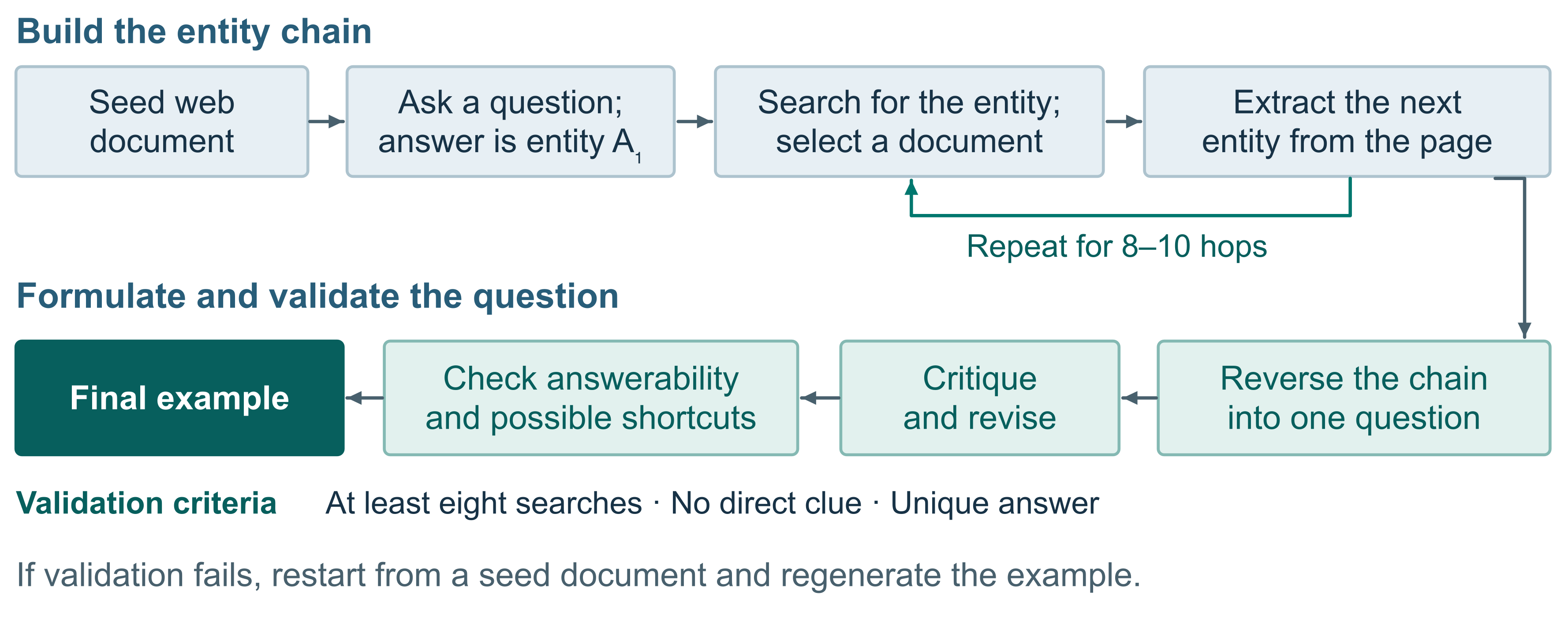}

\par

\captionof{figure}{Generation of multistep search questions by building
entity chains, formulating questions, and applying automated critique
and revision.}

\end{minipage}\par\addvspace{7pt}

The low success rates and long trajectories indicate that the dataset is
challenging for the evaluated models (Table 26). We prepared 10,000
questions and generated approximately 1B tokens of search trajectories.
Adding these trajectories during pretraining improved BrowseCompPlus by
approximately 3 percentage points in our experiments.

\Needspace{9\baselineskip}\begingroup\small\setlength{\LTpre}{4pt}\setlength{\LTpost}{5pt}

\captionof{table}{Open-model performance on generated multistep search
questions. The mean number of agent steps is computed over all attempts,
including unsuccessful ones.}

\nopagebreak[4]

\small\setlength{\tabcolsep}{3pt}

{\def\LTcaptype{reporttablebody} 
\begin{longtable}[]{@{}
  >{\raggedright\arraybackslash}p{(\linewidth - 4\tabcolsep) * \real{0.5001}}
  >{\raggedright\arraybackslash}p{(\linewidth - 4\tabcolsep) * \real{0.2500}}
  >{\raggedright\arraybackslash}p{(\linewidth - 4\tabcolsep) * \real{0.2500}}@{}}
\toprule\noalign{}
\begin{minipage}[b]{\linewidth}\raggedright
\end{minipage} & \begin{minipage}[b]{\linewidth}\raggedright
\textbf{Qwen3.5-397B-A17B}
\end{minipage} & \begin{minipage}[b]{\linewidth}\raggedright
\textbf{Kimi-K2.6}
\end{minipage} \\
\midrule\noalign{}
\endhead
\bottomrule\noalign{}
\endlastfoot
Fraction correct & 0.39 & 0.43 \\
Mean number of agent steps & 9.51 & 12.0 \\
\end{longtable}
}

\endgroup

\subsubsection{Software Engineering}\label{software-engineering}

Following daVinci-Dev {[}44{]}, we combine generated agent trajectories
with examples produced without an interactive agent loop (agentless
data). Adding these data improved scores on the internal SweProxy
benchmark by 7 percentage points and the SWE evaluation by 3 percentage
points in our experiments.

\section{Chat Template and Data
Format}\label{chat-template-and-data-format}

Agentic trajectories are stored as sequences of messages in a format
similar to the OpenAI chat format, extended with a model-specific meta
role: a trajectory is a sequence of messages with system, user,
assistant, tool, and meta roles; available tool definitions are supplied
separately in a tools field. Before tokenization, we serialize each
trajectory using our Jinja chat template. It specifies role prefixes and
the serialization of reasoning, tool descriptions, function calls, and
execution results.

For subsequent cold-start training or RL, we recommend using this
message format and the released chat template\footnote{\url{https://huggingface.co/yandex/AliceAI-Foundation-80B-A3B-Base/blob/main/finetune/chat_template.jinja}}
to match the format used during pretraining.

\section*{Conclusion}\label{conclusion}

We have presented AliceAI-Foundation-80B-A3B-Base, an open
mixture-of-experts model trained from scratch. It surpasses our previous
proprietary Alice AI LLM in Russian-language factual knowledge while
using substantially fewer total and active parameters. It also
outperforms larger open base models on WikiWebFacts and HardMultiQA, and
comparable open base models on IMO-AnswerBench and LiveCodeBench under
zero-shot pass@k evaluation.

Controlled ablations with a fixed training budget demonstrate that our
architecture improves overall performance relative to Qwen3-Next. We
also found that hyperparameter scaling-law predictions depend strongly
on the validation data used to fit them. Revising these datasets
produced learning-rate and batch-size predictions that improved
downstream performance over our baseline settings. An auxiliary loss on
MoE output activations enabled stable training at the predicted learning
rate.

We provide a detailed description of how we analyzed and improved the
pretraining data pipeline, beginning with source collection and
filtering. These improvements increased the number of useful documents
retained for training. We introduced factual web data as a separate
component of the corpus, developed its retrieval and curation pipeline,
and adapted these methods to mathematics, programming, and STEM. We also
documented the augmentation strategies used across these domains.

Finally, we describe the construction of the reasoning-stage dataset in
detail, from collecting and generating mathematical and programming
problems to producing and filtering reasoning traces and tool-use
trajectories. Training on reasoning traces improved performance on
reasoning benchmarks. Adding tool-use trajectories during pretraining
improved agentic benchmark performance when the same cold-start training
procedure was applied to the compared models.

The pretrained checkpoint is openly available under the Apache 2.0
license. We also release WikiWebFacts and HardMultiQA, two benchmarks
designed to assess factual knowledge relevant to Russian-speaking users,
together with their evaluation protocols. These releases support
independent evaluation and further research on pretraining data,
knowledge acquisition, reasoning, and agentic capabilities.

\section*{Acknowledgements}\label{acknowledgements}

We thank Alexander Boymel, Petr Ermakov, and Timur Gaskarov for
proofreading and corrections.

We are grateful to Ekaterina Enikeeva for helping translate this report
into English.

\Needspace{.43\textheight}
\section*{Appendix A. Authors and
Contributions}\label{appendix-a.-authors-and-contributions}

\textbf{Evaluation and Analysis}

Irina Lialikova, Tatiana Taekina, Anastasia Volotova, Assol Kubaeva,
Kirill Alekseev, Pavel Otlivanchik, Mukhammadfirdavs Kosimov, and Vladislav
Negodin.

\textbf{Data Curation and Model Development}

Nikolay Skachkov, Ekaterina Redina, Maksim Konoplev, Evgenii Uskov,
Veronika Zykova, Ivan Tsarikov, Artemii Zakharov, Mariia Godunova, Dmitrii
Lunin, Maria Nikiforova, Valerii Levanov, Anton Kalsin, Daniil Panteleev, Alexey Gerasev,
Ilia Kopytin, Yermek Kapushev, and Egor Liakhov.

\textbf{Model Architecture}

Ivan Vinogradov, Anton Andriushchenko, Maksim Nikitin, Daniil Sukhoi,
Alexander Mazitov, Maksim Abraham, Philip Zmushko, and Maksim Ignatov.

\textbf{Inference Infrastructure}

Ivan Sapozhkov, Timofey Byzov, and Edgar Shmavonian.

\section*{References}\label{references}

\small\raggedright\interlinepenalty=10000\setlength{\parskip}{3pt}

\protect\phantomsection\label{_heading=h.78rggyhkzwah}{}{[}1{]} Alice AI
LLM Technical Report.
\url{https://habr.com/ru/companies/yandex/articles/974594/}, 2025.

\protect\phantomsection\label{_heading=h.22q7kzru1l9p}{}{[}2{]} Mandar
Joshi, Eunsol Choi, Daniel~S. Weld, and Luke Zettlemoyer. TriviaQA: A
Large Scale Distantly Supervised Challenge Dataset for Reading
Comprehension. \url{https://arxiv.org/abs/1705.03551}, 2017.

\protect\phantomsection\label{_heading=h.nbdmi4tu2oiu}{}{[}3{]} Tom
Kwiatkowski, Jennimaria Palomaki, Olivia Redfield, et~al. Natural
Questions: A Benchmark for Question Answering Research.
\url{https://aclanthology.org/Q19-1026/}, 2019.

\protect\phantomsection\label{_heading=h.3rz67r67kuuq}{}{[}4{]} Jonathan
Berant, Andrew Chou, Roy Frostig, and Percy Liang. Semantic Parsing on
Freebase from Question-Answer Pairs.
\url{https://aclanthology.org/D13-1160/}, 2013.

\protect\phantomsection\label{_heading=h.2p67eldlh7ec}{}{[}5{]} Jason
Wei, Nguyen Karina, Hyung~Won Chung, et~al. Measuring short-form
factuality in large language models.
\url{https://arxiv.org/abs/2411.04368}, 2024.

\protect\phantomsection\label{_heading=h.gturprr44st}{}{[}6{]} Yubo
Wang, Xueguang Ma, Ge~Zhang, et~al. MMLU-Pro: A More Robust and
Challenging Multi-Task Language Understanding Benchmark.
\url{https://arxiv.org/abs/2406.01574}, 2024a.

\protect\phantomsection\label{_heading=h.xhnjvb9w2dn}{}{[}7{]} M-A-P
Team, Xinrun Du, Yifan Yao, et~al. SuperGPQA: Scaling LLM Evaluation
across 285 Graduate Disciplines. \url{https://arxiv.org/abs/2502.14739},
2025.

\protect\phantomsection\label{_heading=h.5r81yw8g9oeh}{}{[}8{]} Dan
Hendrycks, Collin Burns, Saurav Kadavath, et~al. Measuring Mathematical
Problem Solving With the MATH Dataset.
\url{https://arxiv.org/abs/2103.03874}, 2021.

\protect\phantomsection\label{_heading=h.bklrgfrfrlut}{}{[}9{]}
Terry~Yue Zhuo, Minh~Chien Vu, Jenny Chim, et~al. BigCodeBench:
Benchmarking Code Generation with Diverse Function Calls and Complex
Instructions. \url{https://arxiv.org/abs/2406.15877}, 2024.

\protect\phantomsection\label{_heading=h.hih2dyk9dnms}{}{[}10{]} Naman
Jain, King Han, Alex Gu, et~al. LiveCodeBench: Holistic and
Contamination Free Evaluation of Large Language Models for Code.
\url{https://arxiv.org/abs/2403.07974}, 2024.

\protect\phantomsection\label{_heading=h.8g5tcyk6mw3j}{}{[}11{]} Mark
Chen, Jerry Tworek, Heewoo Jun, et~al. Evaluating Large Language Models
Trained on Code. \url{https://arxiv.org/abs/2107.03374}, 2021a.

\protect\phantomsection\label{_heading=h.l5sclznlh0k3}{}{[}12{]} Jacob
Austin, Augustus Odena, Maxwell Nye, et~al. Program Synthesis with Large
Language Models. \url{https://arxiv.org/abs/2108.07732}, 2021.

\protect\phantomsection\label{_heading=h.xpemf2syv2y2}{}{[}13{]} Yang
Yue, Zhiqi Chen, Rui Lu, et~al. Does Reinforcement Learning Really
Incentivize Reasoning Capacity in LLMs Beyond the Base Model?
\url{https://arxiv.org/abs/2504.13837}, 2025.

\protect\phantomsection\label{_heading=h.hzsdvcm30v1u}{}{[}14{]} NVIDIA,
Aakshita Chandiramani, Aaron Blakeman, et~al. Nemotron 3 Super: Open,
Efficient Mixture-of-Experts Hybrid Mamba-Transformer Model for Agentic
Reasoning. \url{https://arxiv.org/abs/2604.12374}, 2026.

\protect\phantomsection\label{_heading=h.tagj5qdjldqz}{}{[}15{]} Di~Wu,
Hongwei Wang, Wenhao Yu, Yuwei Zhang, Kai-Wei Chang, and Dong Yu.
LongMemEval: Benchmarking Chat Assistants on Long-Term Interactive
Memory. \url{https://arxiv.org/abs/2410.10813}, 2024.

\protect\phantomsection\label{_heading=h.m9vokux8vld1}{}{[}16{]} Zhiyu
Chen, Wenhu Chen, Charese Smiley, et~al. FinQA: A Dataset of Numerical
Reasoning over Financial Data. \url{https://arxiv.org/abs/2109.00122},
2021b.

\protect\phantomsection\label{_heading=h.sdrn8c2szi6m}{}{[}17{]} Victor
Barres, Honghua Dong, Soham Ray, Xujie Si, and Karthik Narasimhan.
\(\tau^{2}\)-Bench: Evaluating Conversational Agents in a Dual-Control
Environment. \url{https://arxiv.org/abs/2506.07982}, 2025.

\protect\phantomsection\label{_heading=h.82csxetqjvhn}{}{[}18{]} Wei He,
Yueqing Sun, Hongyan Hao, et~al. VitaBench: Benchmarking LLM Agents with
Versatile Interactive Tasks in Real-world Applications.
\url{https://arxiv.org/abs/2509.26490}, 2025.

\protect\phantomsection\label{_heading=h.hhxjg5l63n7d}{}{[}19{]} Yinger
Zhang, Shutong Jiang, Renhao Li, et~al. DeepPlanning: Benchmarking
Long-Horizon Agentic Planning with Verifiable Constraints.
\url{https://arxiv.org/abs/2601.18137}, 2026.

\protect\phantomsection\label{_heading=h.ox5e9jqvoq4r}{}{[}20{]}
Carlos~E. Jimenez, John Yang, Alexander Wettig, et~al. SWE-bench: Can
Language Models Resolve Real-World GitHub Issues?
\url{https://arxiv.org/abs/2310.06770}, 2023.

\protect\phantomsection\label{_heading=h.r0prcmdlmmzh}{}{[}21{]} Xingyao
Wang, Boxuan Li, Yufan Song, et~al. OpenHands: An Open Platform for AI
Software Developers as Generalist Agents.
\url{https://arxiv.org/abs/2407.16741}, 2024b.

\protect\phantomsection\label{_heading=h.okq3e2yo5bmy}{}{[}22{]}
DeepSeek-AI, Aixin Liu, Bei Feng, et~al. DeepSeek-V3 Technical Report.
\url{https://arxiv.org/abs/2412.19437}, 2024.

\protect\phantomsection\label{_heading=h.gowrhae9z46n}{}{[}23{]} Barret
Zoph, Irwan Bello, Sameer Kumar, et~al. ST-MoE: Designing Stable and
Transferable Sparse Expert Models.
\url{https://arxiv.org/abs/2202.08906}, 2022.

\protect\phantomsection\label{_heading=h.atq2c0vnb62o}{}{[}24{]} GPT-OSS
reference implementation.
\url{https://github.com/openai/gpt-oss/blob/main/gpt_oss/torch/model.py},
2025.

\protect\phantomsection\label{_heading=h.hs7ep7z6bkcg}{}{[}25{]} Zihan
Qiu, Zeyu Huang, Kaiyue Wen, et~al. A Unified View of Attention and
Residual Sinks: Outlier-Driven Rescaling is Essential for Transformer
Training. \url{https://arxiv.org/abs/2601.22966}, 2026.

\protect\phantomsection\label{_heading=h.1zw40gbz19a6}{}{[}26{]}
Jingyuan Liu, Jianlin Su, Xingcheng Yao, et~al. Muon is Scalable for LLM
Training. \url{https://arxiv.org/abs/2502.16982}, 2025.

\protect\phantomsection\label{_heading=h.dfwgtlqlo5dq}{}{[}27{]} Kimi
Team, Yu~Zhang, Zongyu Lin, et~al. Kimi Linear: An Expressive, Efficient
Attention Architecture. \url{https://arxiv.org/abs/2510.26692}, 2025a.

\protect\phantomsection\label{_heading=h.5fipkt5pphrc}{}{[}28{]} Kimi
Team, Guangyu Chen, Yu~Zhang, et~al. Attention Residuals.
\url{https://arxiv.org/abs/2603.15031}, 2026.

\protect\phantomsection\label{_heading=h.k9cvq2epkswd}{}{[}29{]} Mingze
Wang, Shuchen Zhu, Yuxin Fang, Binghui Li, Kai Shen, and Shu Zhong.
Negligible in Size, Significant in Effect: On Scale Vectors in Large
Language Models. \url{https://arxiv.org/abs/2605.26895}, 2026.

\protect\phantomsection\label{_heading=h.i8ubmz4a9qit}{}{[}30{]} Laguna
M1 XS2 Technical Report.
\url{https://poolside.ai/assets/laguna/laguna-m1-xs2-technical-report.pdf},
2026.

\protect\phantomsection\label{_heading=h.wngo52z3q0z3}{}{[}31{]} Jared
Kaplan, Sam McCandlish, Tom Henighan, et~al. Scaling Laws for Neural
Language Models. \url{https://arxiv.org/abs/2001.08361}, 2020.

{[}32{]} UltraData-Math.
\url{https://huggingface.co/datasets/openbmb/UltraData-Math}, 2026.

{[}33{]} Nemotron-CC-Math-v1.
\url{https://huggingface.co/datasets/nvidia/Nemotron-CC-Math-v1}, 2025.

{[}34{]} Kimi Team, Yifan Bai, Yiping Bao, et~al. Kimi K2: Open Agentic
Intelligence. \url{https://arxiv.org/abs/2507.20534}, 2025b.

{[}35{]} Zorik Gekhman, Gal Yona, Roee Aharoni, et~al. Does Fine-Tuning
LLMs on New Knowledge Encourage Hallucinations?
\url{https://aclanthology.org/2024.emnlp-main.444/}, 2024.

{[}36{]} Zeyuan Allen-Zhu and Yuanzhi Li. Physics of Language Models:
Part 3.1, Knowledge Storage and Extraction.
\url{https://arxiv.org/abs/2309.14316}, 2023.

{[}37{]} Hoyeon Chang, Jinho Park, Seonghyeon Ye, et~al. How Do Large
Language Models Acquire Factual Knowledge During Pretraining?
\url{https://arxiv.org/abs/2406.11813}, 2024.

{[}38{]} Zengzhi Wang, Fan Zhou, Xuefeng Li, and Pengfei Liu.
OctoThinker: Mid-training Incentivizes Reinforcement Learning Scaling.
\url{https://arxiv.org/abs/2506.20512}, 2025.

{[}39{]} Syeda~Nahida Akter, Shrimai Prabhumoye, Eric Nyberg, et~al.
Front-Loading Reasoning: The Synergy between Pretraining and
Post-Training Data. \url{https://arxiv.org/abs/2510.03264}, 2025.

{[}40{]} Qwen3.5. \url{https://qwen.ai/blog?id=qwen3.5}, 2026.

{[}41{]} Zhihao Xu, Rumei Li, Jiahuan Li, et~al. Unlocking Implicit
Experience: Synthesizing Tool-Use Trajectories from Text.
\url{https://arxiv.org/abs/2601.10355}, 2026.

{[}42{]} Liangcai Su, Zhen Zhang, Guangyu Li, et~al. Scaling Agents via
Continual Pre-training. \url{https://arxiv.org/abs/2509.13310}, 2025.

{[}43{]} Agent World Model.
\url{https://github.com/Snowflake-Labs/agent-world-model}, 2026.

{[}44{]} Ji~Zeng, Dayuan Fu, Tiantian Mi, et~al. daVinci-Dev:
Agent-native Mid-training for Software Engineering.
\url{https://arxiv.org/abs/2601.18418}, 2026.

{[}45{]} Andrey But. Introducing YandexGPT 5 in Alice, the cloud, and
open source (in Russian).
\url{https://habr.com/ru/companies/yandex/articles/885218/}, 2025.

{[}46{]} Syeda Nahida Akter, Shrimai Prabhumoye, John Kamalu, et al.
MIND: Math Informed syNthetic Dialogues for Pretraining LLMs.
\url{https://arxiv.org/abs/2410.12881}, 2024.
\end{document}